%% file: main_ICLR_SSL.tex
\documentclass{article} 
\usepackage{iclr2019_conference,times}

\input{math_commands.tex}

\usepackage{hyperref}
\usepackage{url}
\usepackage{bbm}

\PassOptionsToPackage{numbers}{natbib}

\usepackage[verbose=true,letterpaper]{geometry}
  \newgeometry{
    textheight=9in,
    textwidth=5.5in,
    top=1in,
    headheight=12pt,
    headsep=25pt,
    footskip=30pt
  }
  
\usepackage{parskip}

\usepackage[utf8]{inputenc} 
\usepackage[T1]{fontenc}    
\usepackage{hyperref}       
\usepackage{url}            
\usepackage{booktabs}       
\usepackage{amsfonts}       
\usepackage{nicefrac}       
\usepackage{microtype}      
\usepackage{algorithmic}
\usepackage{algorithm}
\usepackage{natbib}
\usepackage{tikz}

\usepackage{xcolor}
\definecolor{dark-red}{rgb}{0.4,0.15,0.15}
\definecolor{dark-blue}{rgb}{0.15,0.15,0.4}
\definecolor{medium-blue}{rgb}{0,0,0.5}
\hypersetup{
   colorlinks, linkcolor={dark-blue},
   citecolor={black}, urlcolor={medium-blue}
}

\usepackage{amsfonts}
\usepackage{color}

\usepackage{url}
\usepackage{amsmath}
\usepackage{verbatim}

\newcommand{\tr}{\text{tr}}

\usepackage{subcaption}

\usepackage[flushleft]{threeparttable}
\usepackage{makecell}

\newlength{\mzerolen}

\newcommand{\tbf}[1]{\textbf{{#1}}}
\newcommand{\D}{\mathcal{D}}
\newcommand{\N}{\mathcal{N}}

\usepackage{soul}

\usepackage{xspace}

\newcommand{\pii}{$\Pi$\xspace}
\renewcommand{\pi}{\pii}
\newcommand{\tpm}{$\pm$ }
\newcommand{\fastswa}{fast-SWA\xspace}

\newcommand{\swa}{SWA\xspace} 

\newcommand{\tb}[1]{\textbf{#1}}

\usepackage{upgreek}
\usepackage[export]{adjustbox}
\usepackage{enumitem}

\makeatletter
\renewenvironment{abstract}{%
    \if@twocolumn
      \section*{\abstractname}%
    \else 
      \begin{center}%
        {\bfseries \large\abstractname\vspace{\z@}}
      \end{center}%
      \quotation
    \fi}
    {\if@twocolumn\else\endquotation\fi}
\makeatother

\date{}

\title{There are Many Consistent Explanations of Unlabeled Data: Why You Should Average}

\author{
Ben Athiwaratkun, Marc Finzi, Pavel Izmailov, Andrew Gordon Wilson \\
\texttt{\{pa338, maf388, pi49, andrew\}@cornell.edu} \\
Cornell University
}

\iclrfinalcopy
\begin{document}

\maketitle

\begin{abstract}

\noindent
Presently the most successful approaches to semi-supervised
learning are based on \emph{consistency regularization}, whereby a model is
trained to be robust to small perturbations of its inputs and parameters. To 
understand consistency regularization, we conceptually explore how loss geometry
interacts with training procedures.   
The consistency loss dramatically improves generalization performance over
supervised-only training;
however, we show that SGD struggles to converge on the consistency loss and continues to make large steps that lead to 
 changes in predictions on the test data.
Motivated by these observations, we propose to train consistency-based methods 
with Stochastic Weight Averaging (SWA), a recent approach which averages weights
along the trajectory of SGD with a modified learning rate schedule.
We also propose \emph{fast-SWA},
which further accelerates convergence by averaging multiple points within each
cycle of a cyclical learning rate schedule.
With weight averaging, we achieve the 
best known semi-supervised results on CIFAR-10 and CIFAR-100, over many different quantities of
labeled training data. For example, we achieve 5.0\% error on CIFAR-10 with only 4000 labels,
compared to the previous best result in the literature of 6.3\%.

\end{abstract}

\section{Introduction} \label{sec:intro}

Recent advances in deep unsupervised learning, such as generative adversarial networks (GANs) \citep{gans}, 
have led to an explosion of interest in semi-supervised learning.  Semi-supervised methods make 
use of both unlabeled and labeled training data to improve performance over purely supervised 
methods.  Semi-supervised learning is particularly valuable in applications such as medical imaging, 
where labeled data may be scarce and expensive \citep{ssl-eval}.

Currently the best semi-supervised results are obtained by \emph{consistency-enforcing}
approaches \citep{bachman2014learning, pi, mt, vat2, vad}.
These methods use unlabeled data to stabilize their predictions under input or weight 
perturbations. Consistency-enforcing methods can be used at scale with state-of-the-art 
architectures. For example, the recent Mean Teacher \citep{mt} model has been used with 
the Shake-Shake \citep{shake} architecture and has achieved the best  
semi-supervised performance on the consequential CIFAR benchmarks.

This paper is about conceptually 
understanding and improving consistency-based 
semi-supervised learning methods. Our approach 
can be used as a guide for exploring how 
loss geometry interacts with training procedures in general.
We provide several novel observations about  the training 
objective and optimization trajectories of the popular 
\pi \citep{pi} and Mean Teacher \citep{mt} consistency-based models.
Inspired by these
findings, we propose to improve SGD solutions via stochastic weight averaging (SWA) 
\citep{swa}, a recent method that averages weights 
of the networks corresponding to different training epochs to obtain a 
single model with improved generalization.
On a thorough empirical study we show that this procedure achieves the best known 
semi-supervised results on consequential benchmarks. In particular:

\begin{itemize}[leftmargin=*]
\item We show in Section \ref{sec:flatness_theory} that a simplified \pi model 
    implicitly regularizes the norm of the Jacobian of the network outputs with 
    respect to both its inputs and its weights, which in turn encourages flatter 
    solutions. 
    Both the reduced Jacobian norm and flatness of solutions have been related 
    to generalization in the literature
    \citep{sokolic2017robust,sensitivity_gen_nn,entropy-sgd, schmidhuber, keskar2017,swa}.
    Interpolating 
    between the weights corresponding to different epochs of training 
    we demonstrate that the solutions of \pi and Mean Teacher models are indeed flatter
    along these directions (Figure \ref{fig:sgd_sgd_rays}).

\item In Section \ref{sec:trajectories}, we compare the training trajectories of 
    the \pi, Mean Teacher, and supervised models and find that the distances 
    between the weights corresponding to different epochs are much larger for 
    the consistency based models. The error curves of consistency models are 
    also wider (Figure \ref{fig:sgd_sgd_rays}), which can be explained by the
    flatness of the solutions discussed in section \ref{sec:flatness_theory}. 
    Further we observe that the predictions of the SGD iterates can differ significantly
    between different iterations of SGD.

\item We observe that for consistency-based methods, SGD does not converge to a single point but continues to explore many solutions with high distances apart.
    Inspired by this observation, we propose to 
 average the weights corresponding to SGD iterates, or ensemble the predictions
 of the models corresponding to these weights. Averaging 
    weights of SGD iterates compensates for larger steps,
    stabilizes SGD trajectories and obtains a solution that is centered in a
    flat region of the loss (as a function of weights).   
    Further, we show that the SGD iterates correspond to models with diverse predictions -- using weight averaging or ensembling allows us to make use of the improved diversity and obtain a better solution compared to the SGD iterates.
    In Section \ref{sec:empirical_convexity} we demonstrate that both
    ensembling predictions and averaging weights of the networks corresponding to 
    different training epochs significantly improve generalization performance
    and find that the improvement is much larger for the \pi and Mean Teacher
    models compared to supervised training. We find that averaging 
    weights provides similar or improved accuracy compared to ensembling,
    while offering the computational benefits and convenience of working with 
    a single model. Thus, we focus on weight averaging for the  
    remainder of the paper.

\item Motivated by our observations in Section \ref{sec:analysis} 
    we propose to apply Stochastic Weight Averaging (SWA) \citep{swa} to the
    \pi and Mean Teacher models. 
    Based on our results in Section \ref{sec:empirical_convexity} we propose
    several modifications to SWA in Section \ref{sec:better_solutions_averaging}.
    In particular, we propose \fastswa, which (1) uses a learning rate schedule with longer
    cycles to increase the distance between the weights that are averaged and the
    diversity of the corresponding predictions; and (2)
    averages weights of multiple networks within each cycle 
    (while SWA only averages weights corresponding to the lowest values of
    the learning rate within each cycle). In Section \ref{sec:exps}, we show that \fastswa 
    converges to a good solution much faster than \swa. 

\item Applying weight averaging to the \pi and Mean Teacher models we improve the best reported results
    on CIFAR-10 for $1k$, $2k$, $4k$ and $10k$ labeled examples, as
    well as on CIFAR-100 with $10k$ labeled examples. For example, we obtain
    $5.0\%$ error on CIFAR-10 with only $4k$ labels, improving the best
    result reported in the literature \citep{mt} by $1.3\%$. 
    We also apply weight averaging to a state-of-the-art domain adaptation technique \citep{da_mt}
    closely related to the Mean Teacher model and improve the best reported results on 
    domain adaptation from CIFAR-10 to STL from $19.9\%$ to $16.8\%$ error. 

\item  We release our code at 
\url{https://github.com/benathi/fastswa-semi-sup}

\end{itemize}

\section{Background}
\label{sec:background}

\subsection{Consistency Based Models}
We briefly review semi-supervised learning with consistency-based models.
This class of models encourages predictions to stay similar under small perturbations of inputs or 
network parameters. 
For instance, two different translations of the same image should result in similar predicted probabilities. 
The consistency of a model (student) can be measured against its own predictions (e.g. \pi model) or
predictions of a different teacher network (e.g. Mean Teacher model). 
In both cases we will say a \emph{student} network measures consistency against a \emph{teacher} network.

\paragraph{\textbf{Consistency Loss}}
In the semi-supervised setting, we have access to labeled data 
$\D_L = \{(x^L_i, y^L_i)\}_{i=1}^{N_L}$, and unlabeled data 
$\D_U = \{x^U_i\}_{i=1}^{N_U}$.

Given two perturbed inputs $x',x''$ of $x$ and the perturbed weights $w_f'$ and $w_g'$, the consistency loss penalizes the difference between the \emph{student}'s predicted probablities $f(x'; w_f')$ and the \emph{teacher}'s $g(x''; w_g')$. This loss is typically the Mean Squared Error or KL divergence:
\begin{equation}
\label{eq:consistency}
\ell_{\text{cons}}^{\text{MSE}}(w_f,x) = \| f(x'; w_f') - g(x'', w_g') \|^2 \text{ or } \ell_{\text{cons}}^{\text{KL}}(w_f,x) = \text{KL}( f(x'; w_f') || g(x'', w_g') ) \,.
\end{equation}

The total loss used to train the model can be written as 
\begin{equation}
  \label{eq:semisup_loss}
  L(w_f) = 
  \underbrace{\sum_{(x, y) \in \D_L} \ell_{\text{CE}} (w_f, x, y)}_{L_{\text{CE}}} + 
  \lambda \underbrace{ \sum_{x \in \D_L \cup \D_U} \ell_{\text{cons}} (w_f,  x)}_{L_{\text{cons}}},
\end{equation}
where for classification $L_{\text{CE}}$ is the cross entropy between the model predictions
and supervised training labels. The parameter $\lambda > 0$ controls the relative importance 
of the consistency term in the overall loss.

\paragraph{\textbf{\pi Model}}
The \pi model, introduced in \citet{pi} and \citet{sajjadi_regularization_2016}, uses the student model $f$ as its own teacher.
  The data (input) perturbations include random translations,
 crops, flips and additive Gaussian noise. Binary dropout \citep{dropout} is used for weight perturbation.

\paragraph{\textbf{Mean Teacher Model}}
The Mean Teacher model (MT) proposed in \citet{mt} uses the same data 
and weight perturbations as the \pi model; 
however, the teacher weights $w^g$ are the exponential moving average (EMA) 
of the student weights $w^f$: 
$ w_g^{k} = \alpha\cdot w_g^{k-1} + (1 - \alpha) \cdot w_f^k$.
The decay rate $\alpha$ is usually set between $0.9$ and $0.999$. 
The Mean Teacher model has the best known
results on the CIFAR-10 semi-supervised learning benchmark \citep{mt}.

\paragraph{\textbf{Other Consistency-Based Models}}

Temporal Ensembling (TE)  \citep{pi} uses an exponential moving average of the 
student outputs as the teacher outputs in the consistency term for training. 
Another approach, Virtual Adversarial Training (VAT) \citep{vat2}, enforces the 
consistency between predictions on the original data inputs and the data perturbed in an 
adversarial direction $x' = x+ \epsilon r_{\text{adv}}$, where 
$r_{\text{adv}} =  \arg \max_{r: \|r\| = 1} \textrm{KL}[f(x,w)\|f(x+\xi r,w)]$. 

\section{Understanding Consistency-Enforcing Models}
\label{sec:analysis}

In Section \ref{sec:flatness_theory}, we study a
simplified version of the \pi model theoretically and show that it penalizes 
the norm of the Jacobian of the outputs with respect to inputs, as well as 
the eigenvalues of the Hessian, both of which have been related to generalization 
\citep{sokolic2017robust, sensitivity_gen_nn, dinh_sharp_2017, entropy-sgd}.
In Section \ref{sec:trajectories} we empirically study the training trajectories
of the \pi and MT models and compare them to the training
trajectories in supervised learning. We show that even late in training
consistency-based methods make large training steps, leading to significant
changes in predictions on test. 
In Section 
\ref{sec:empirical_convexity} we show that averaging weights or ensembling
predictions of the models proposed by SGD at different training epochs can
lead to substantial gains in accuracy and that these gains are much larger 
for \pi and MT than for supervised training.

\subsection{Simplified \pi Model Penalizes Local Sharpness} \label{sec:flatness_theory}

\paragraph{Penalization of the input-output Jacobian norm.}
Consider a simple version of the \pi model, where we only apply small additive perturbations
to the student inputs: $x' = x + \epsilon z,~ z \sim \N(0, I)$ with $\epsilon \ll 1$, 
and the teacher input is unchanged: $x'' = x.$\footnote{This assumption can be 
relaxed to $x'' = x+\epsilon z_2 ~~ z_2 \sim \N(0, I)$ without changing the results of the analysis,
since $\epsilon (z_1 - z_2) = 2\epsilon \bar{z}~$ with $~\bar{z} \sim \N(0, I)$.} 
Then the consistency loss $\ell_{cons}$ (Eq.\ \ref{eq:consistency}) becomes 
$  \ell_{cons}(w, x,\epsilon) = 
  \|f(w, x + \epsilon z) - f(w, x)\|^2 $.
Consider the estimator 
$\hat{Q} = \lim_{\epsilon \to 0}\tfrac{1}{\epsilon^2} 
\tfrac{1}{m}\sum_{i=1}^m\ell_{cons}(w, x_i,\epsilon)$.
We show in Section \ref{sec:pi_model} that
\[
    \mathbb{E} [ \hat{Q}] =  \mathbb{E}_x[\|J_x\|^2_F] 
    \quad \textrm{and} \quad  
    \textrm{Var}[\hat{Q}] =  \frac{1}{m}\bigg(\textrm{Var}[\|J_x\|^2_F] + 2 \mathbb{E}[\|J_x^TJ_x\|^2_F]\bigg),
\] 
where $J_x$ is the Jacobian of the network's outputs with respect to its inputs
evaluated at $x$, $\|\cdot\|_F$ represents Frobenius norm, and the expectation 
$\E_x$ is taken over the distribution of labeled and unlabeled data.
That is, $\hat{Q}$ is an unbiased estimator of $\E_x [\|J_x\|^2_F]$ with
variance controlled by the minibatch size $m$. Therefore, the 
consistency loss implicitly penalizes $\E_x [\|J_x\|^2_F]$. 

The quantity $||J_x||_F$ 
has been related to generalization both theoretically \citep{sokolic2017robust}
and empirically \citep{sensitivity_gen_nn}.
For linear models $f(x) = Wx$, penalizing $||J_x||_F$ exactly corresponds to weight decay, also known as 
$L_2$ regularization, since for linear models $J_x = W$, and $\|W\|_F^2 = \|\textrm{vec}(W)\|_2^2$.
Penalizing $\mathbb{E}_x [\|J_x\|^2_F]$
is also closely related to the graph based (manifold) regularization in 
\citet{zhu_semi-supervised_nodate} which uses the graph Laplacian to approximate $\mathbb{E}_x[\|\nabla_\mathcal{M} f\|^2]$ for
nonlinear models, making use of the manifold structure of unlabeled data.

Isotropic perturbations investigated in this simplified \pi model will not in general lie along the data manifold, and it would be more
pertinent to enforce consistency to perturbations sampled from the space of natural images. In fact, we can interpret consistency with respect to 
standard data augmentations (which are used in practice) as penalizing the manifold Jacobian norm in the same manner as above. See Section \ref{sec:pi_model} for more details.

\paragraph{Penalization of the Hessian's eigenvalues.}
Now, instead of the input perturbation, consider the weight perturbation
$w' = w + \epsilon z$. Similarly, the consistency loss is an unbiased estimator for 
$\mathbb{E}_x [\|J_w\|^2_F]$, where $J_w$ is the Jacobian of the network outputs 
with respect to the weights $w$.
In Section \ref{sec:broad} we show that for the $\textrm{MSE}$ loss,
the expected trace of the Hessian of the loss
$\mathbb{E}_x [\tr(H)]$ can be decomposed into two terms, one of which is 
$\mathbb{E}_x [\|J_w\|^2_F]$. 
As minimizing the consistency loss of a simplified \pi model penalizes 
$\mathbb{E}_x [\|J_w\|^2_F]$,
it also penalizes 
$\mathbb{E}_x [\tr(H)]$. 
As pointed out in \citet{dinh_sharp_2017} and \citet{entropy-sgd}, the 
eigenvalues of $H$ encode the 
local information about sharpness of the loss for a given solution $w$.
Consequently, the quantity $\tr(H)$ which is the sum of the Hessian 
eigenvalues is related to the notion of sharp and flat optima, which has 
recently gained attention as a proxy for generalization performance \citep[see e.g.][]{schmidhuber, keskar2017,swa}.
Thus, based on our analysis, the consistency loss in the simplified \pi model 
encourages flatter solutions.

\subsection{Analysis of Solutions along SGD Trajectories}
\label{sec:trajectories}

In the previous section we have seen that in a simplified \pi model, the 
consistency loss encourages lower input-output Jacobian norm and Hessian's eigenvalues, which are related to better generalization.
In this section we analyze the properties of minimizing the
consistency loss in a practical setting.
Specifically, we explore the trajectories followed by SGD for the consistency-based models 
and compare them to the trajectories in supervised training.

\begin{figure}[h]
  \centering
  \begin{tabular}[c]{c c c c}
  \hspace*{-0.4cm}
\minipage{0.25\textwidth}
       \centering
      \hspace*{-0.3cm}
      \includegraphics[width=0.96\textwidth,trim={0 0 0 0}, clip]{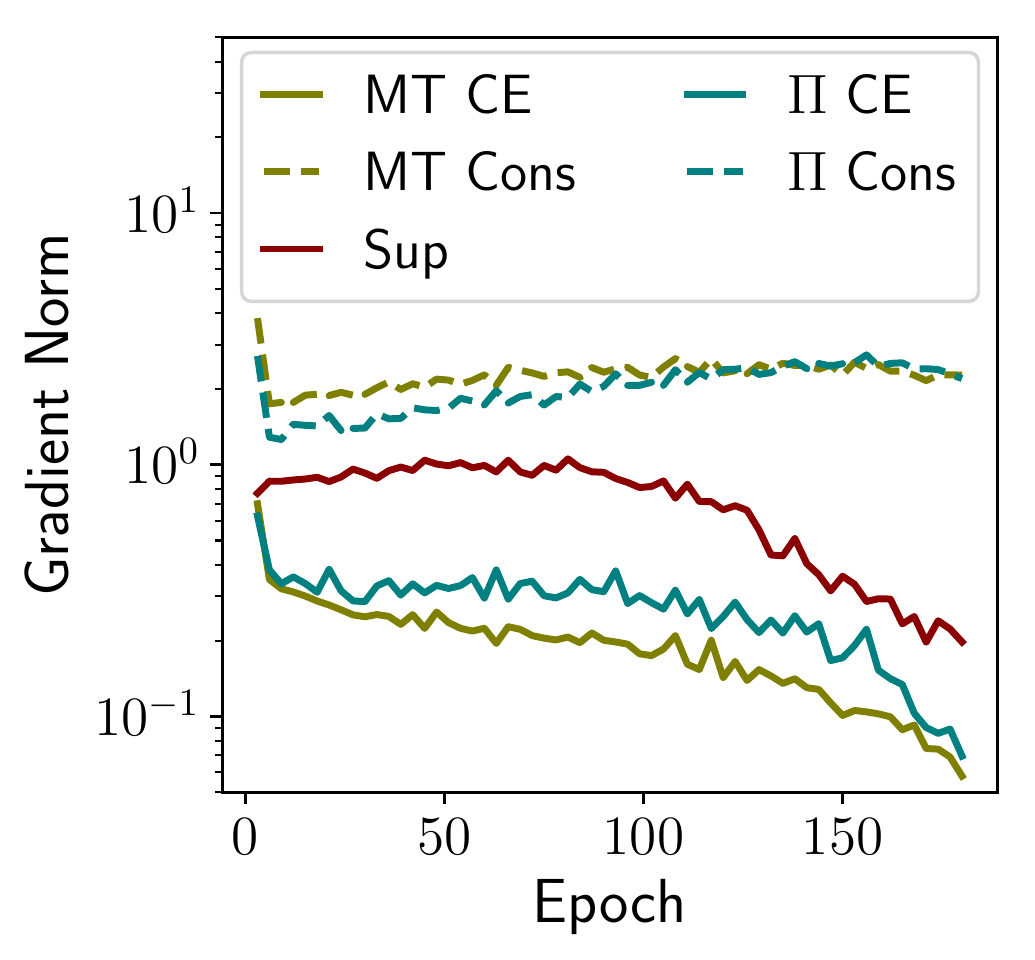}
      \subcaption{Gradient Norm} \label{fig:gradnorm}
\endminipage
& 
\hspace*{-0.4cm}
\minipage{0.25\textwidth}
       \centering
      \hspace*{-0.3cm}
      \includegraphics[width=0.96\textwidth,trim={0 0 0 0}, clip]{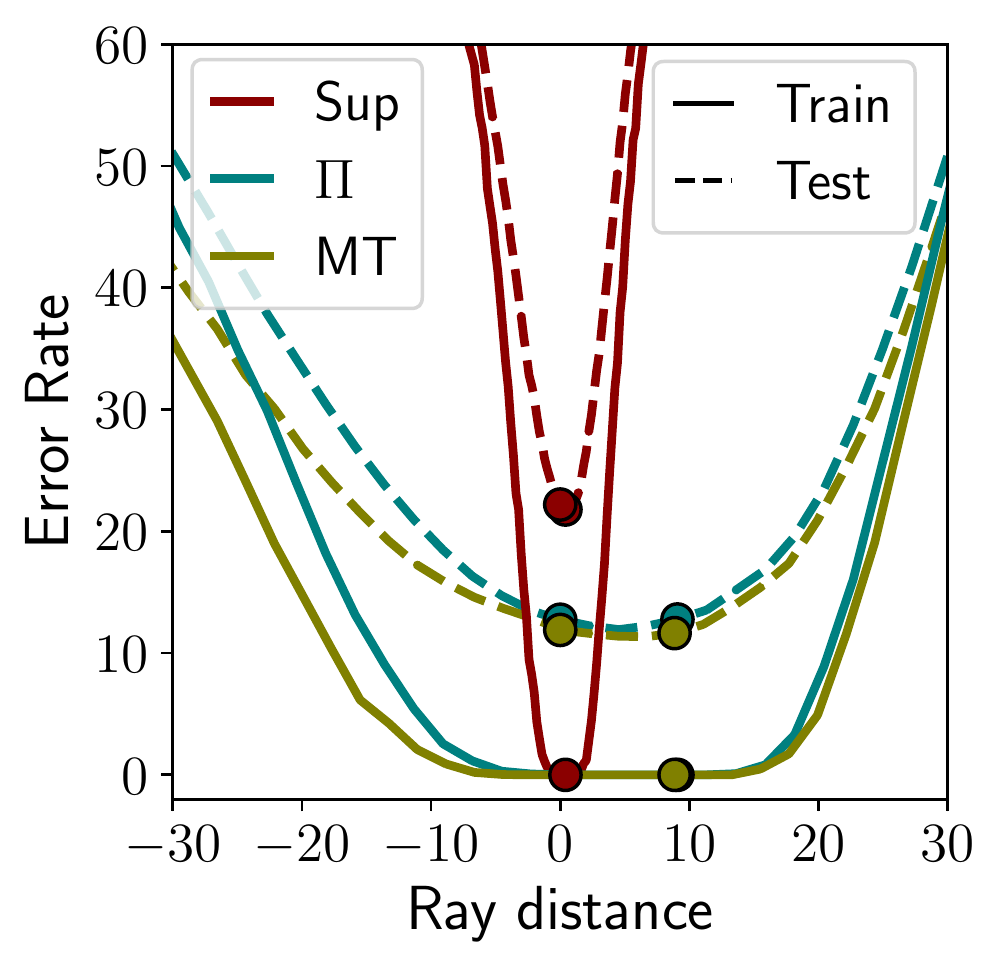}
      \subcaption{SGD-SGD Rays} \label{fig:sgd_sgd_rays}
\endminipage 
&
\hspace*{-0.4cm}
\minipage{0.25\textwidth}
       \centering
      \hspace*{-0.3cm}
      \includegraphics[width=0.96\textwidth,trim={0 0 0 0}, clip]{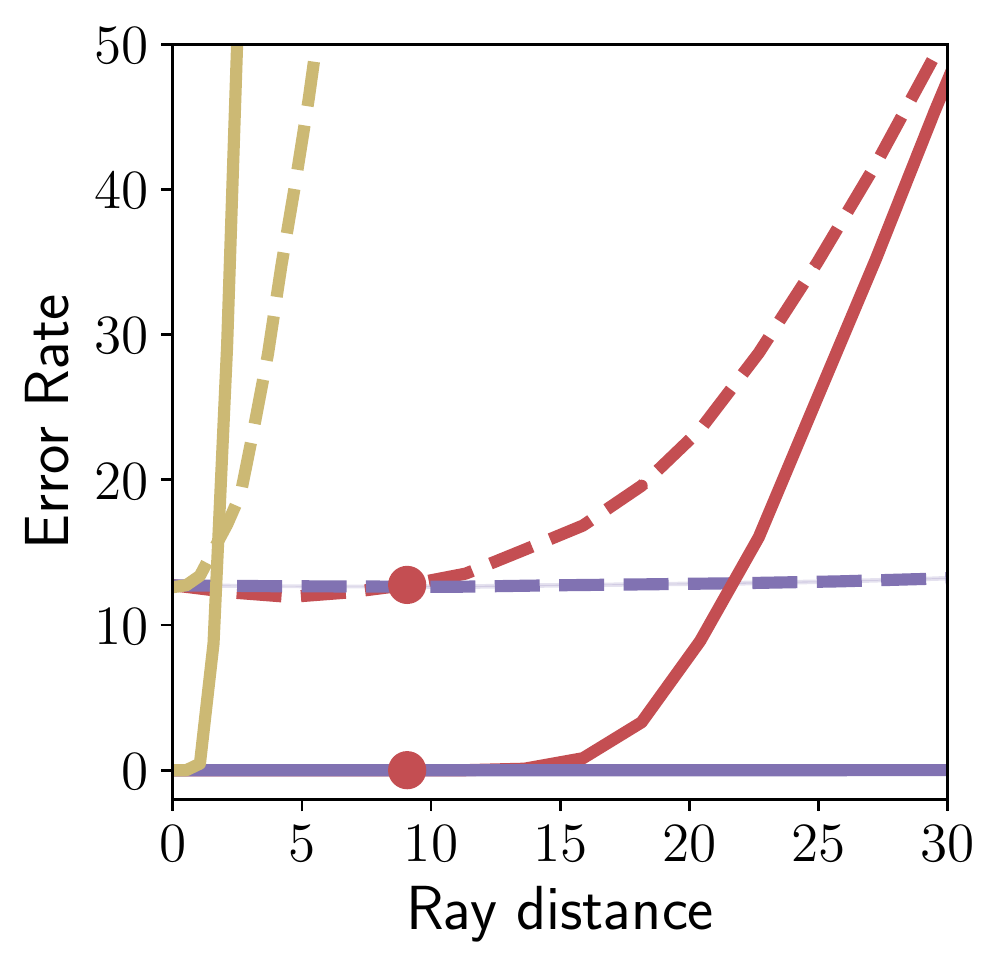}
      \subcaption{\pi model} \label{fig:pi_allrays}
\endminipage
&
\hspace*{-0.4cm}
\minipage{0.25\textwidth}
       \centering
      \hspace*{-0.3cm}
      \includegraphics[width=0.96\textwidth,trim={0 0 0 0}, clip]{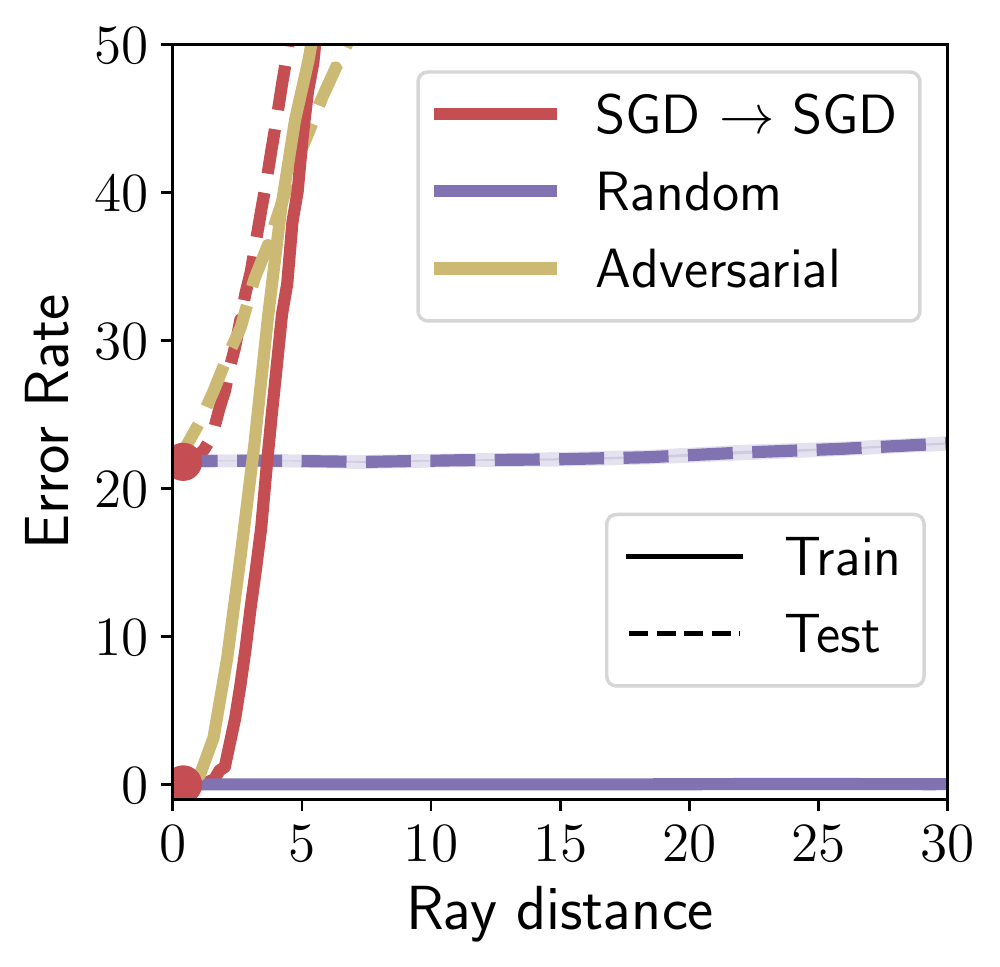}
      \subcaption{Supervised model} \label{fig:sup_allrays}
\endminipage
\end{tabular}
  \caption{
    \tb{(a)}: The evolution of the  gradient norm for the consistency regularization term (Cons) and
    the cross-entropy term (CE) in the \pi, MT, and standard supervised (CE only) models during training.
    \tb{(b)}: Train and test errors along rays connecting two SGD solutions for each respective model. 
    \tb{(c)} and \tb{(d)}: Comparison of errors along rays connecting two SGD solutions, random rays, and adversarial rays for the \pi and supervised models. 
    See Section \ref{sec:sup_rayplots} for the analogous Mean Teacher model's
    plot.
  }
  \label{fig:trajectories}
\end{figure}

We train our models on CIFAR-10 using $4k$ labeled data for $180$ epochs. 
The \pi and Mean Teacher models use $46k$ data points as unlabeled data
(see Sections \ref{sec:sup_architectures} and \ref{sec:sup_hypers} for details). 
First, in Figure \ref{fig:gradnorm} we visualize the evolution of
norms of the gradients of the cross-entropy term $\|\nabla L_{\text{CE}}\|$ and consistency term
$\|\nabla L_{\text{cons}}\|$ along the trajectories of the \pi, MT, and standard supervised models (using CE loss only). 
We observe that $\|\nabla L_{\text{Cons}}\|$ remains high until the end
of training and dominates the gradient $\|\nabla L_{\text{CE}}\|$ of the 
cross-entropy term for the \pi and MT models. Further, for both
the \pi and MT models, $\|\nabla L_{\text{Cons}}\|$ is much larger
than in supervised training 
implying that the \pi and MT models are making 
substantially larger steps until the end of training. 
These larger steps suggest that rather than converging to a single 
minimizer, SGD continues to actively explore a large set of solutions 
when applied to consistency-based methods. 

For further understand this observation, we analyze the behavior of 
train and test errors in the region of weight space around the solutions of 
the \pi and Mean Teacher models.
First, we consider the one-dimensional rays
$\phi(t) = t \cdot w_{180} + (1-t) w_{170},~t \ge 0,$ 
connecting the weight vectors 
$w_{170}$ and $w_{180}$ corresponding to epochs $170$ and
$180$ of training. 
We visualize the train and test errors (measured on the labeled data)
 as functions of the distance from
the weights 
$w_{170}$ 
in Figure \ref{fig:sgd_sgd_rays}. 
We observe that the distance between the weight vectors 
$w_{170}$ and $w_{180}$ 
is much larger for
the semi-supervised methods compared to supervised training, which is consistent with
our observation that the gradient norms are larger which implies larger steps during optimization in the \pi and MT models. 
Further, we observe that the train and test error surfaces are much wider along the directions connecting 
$w_{170}$ and $w_{180}$ 
for the consistency-based methods compared to supervised training.
One possible explanation for the increased width is the effect of the consistency
loss on the Jacobian of the network and the eigenvalues of the Hessian of the
loss discussed in Section \ref{sec:flatness_theory}. 
We also observe that the test errors of 
interpolated weights can be lower than 
errors of the two SGD solutions between which we interpolate.
This error 
reduction is larger in the consistency models (Figure \ref{fig:sgd_sgd_rays}). 

We also analyze the error surfaces along \emph{random} and \emph{adversarial} 
rays starting at the SGD solution 
$w_{180}$ 
for each model. 
For the random rays we sample $5$ random vectors $d$ from the unit sphere and calculate the average 
train and test errors of the network with
weights $w_{t_1} + s d$ for $s \in [0,30]$. With adversarial rays we evaluate 
the error along the directions of the fastest 
ascent of test or train loss $d_{adv} = \frac{\nabla L_{CE}}{|| \nabla L_{CE} ||}$. 
We observe that while the solutions of the \pi and MT models are much wider 
than supervised training solutions along the SGD-SGD directions 
(Figure \ref{fig:sgd_sgd_rays}), their widths along random and adversarial rays 
are comparable (Figure \ref{fig:pi_allrays}, \ref{fig:sup_allrays}) 

We analyze the error along SGD-SGD rays for two reasons. Firstly, in fast-SWA we are averaging solutions traversed by SGD, so the rays connecting SGD iterates serve as a proxy for the space we average over. Secondly, we are interested in evaluating the width of the solutions that we explore during training which we expect will be improved by the consistency training, as discussed in Section \ref{sec:flatness_theory} and \ref{sec:broad}. We expect width along random rays to be less meaningful because there are many directions in the parameter space that do not change the network outputs \citep{sharp_minima_generalize, sgd_tiny_subspace, empirical_hessian_nn}. However, by evaluating SGD-SGD rays, we can expect that these directions corresponds to meaningful changes to our model because individual SGD updates correspond to directions that change the predictions on the training set. Furthermore, we observe that different SGD iterates produce significantly different predictions on the test data.

Neural networks in general are known to be resilient to noise, explaining why both MT, \pi and supervised models are flat along random directions \citep{gen_bounds_compression}. At the same time neural networks are susceptible to targeted perturbations (such as adversarial attacks). We hypothesize that we do not observe improved flatness for semi-supervised methods along adversarial rays because we do not choose our input or weight perturbations adversarially, but rather they are sampled from a predefined set of transformations.

Additionally, we analyze whether the larger optimization steps for the \pi and MT
models translate into higher \emph{diversity} in predictions. 
We define diversity of a pair of models $w_1, w_2$ as 
$\text{Diversity}(w_1, w_2) = \frac{1}{N}\sum_{i=1}^N \mathbbm{1}{[y_i^{(w_1)} \ne y_i^{(w_2)}]}$,
the fraction of test samples where the predicted labels between the two models differ.
We found that for the \pi and MT models, 
the
Diversity($w_{170}$, $w_{180}$) 
is $7.1\%$ and $6.1\%$ of the test data points respectively, which is much higher than $3.9\%$ in supervised learning. 
The increased diversity in the predictions of the networks traversed by SGD
supports our conjecture that for the \pi and MT models SGD struggles
to converge to a single solution and continues to actively explore the set of
plausible solutions until the end of training.

\subsection{Ensembling and Weight Averaging}
\label{sec:empirical_convexity}

In Section \ref{sec:trajectories}, we observed that the \pi and MT models 
continue taking large steps in the weight space at the end of training.
Not only are the distances between weights larger, we observe these models to have higher diversity.
In this setting, using the last SGD iterate
to perform prediction is not ideal since many solutions explored by
SGD are equally accurate but produce different predictions. 

\begin{figure}[h]
  \centering
  \begin{tabular}[c]{c c c c}
  \hspace*{-0.4cm}
\minipage{0.25\textwidth}
       \centering
      \hspace*{-0.1cm}
      \includegraphics[height=3.5cm,trim={44 0 20 -15}, clip=True]{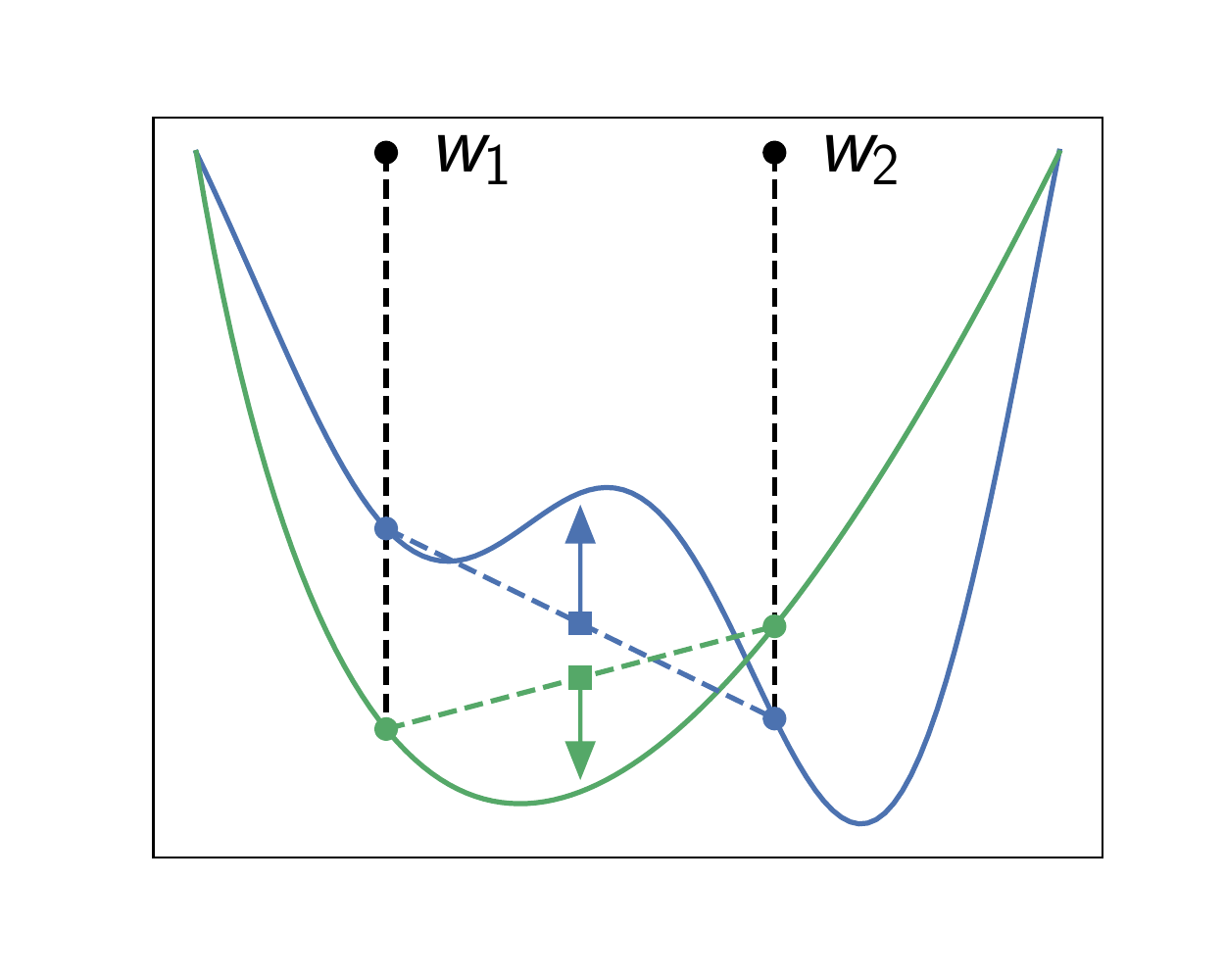}
      \subcaption{Convex vs. non-convex} \label{fig:convex_nonconvex}
\endminipage
&   \hspace*{-0.6cm}
\minipage{0.25\textwidth}
       \centering
      \hspace*{-0.1cm}
      \includegraphics[height=3.5cm,trim={0 0 0 0}, clip]{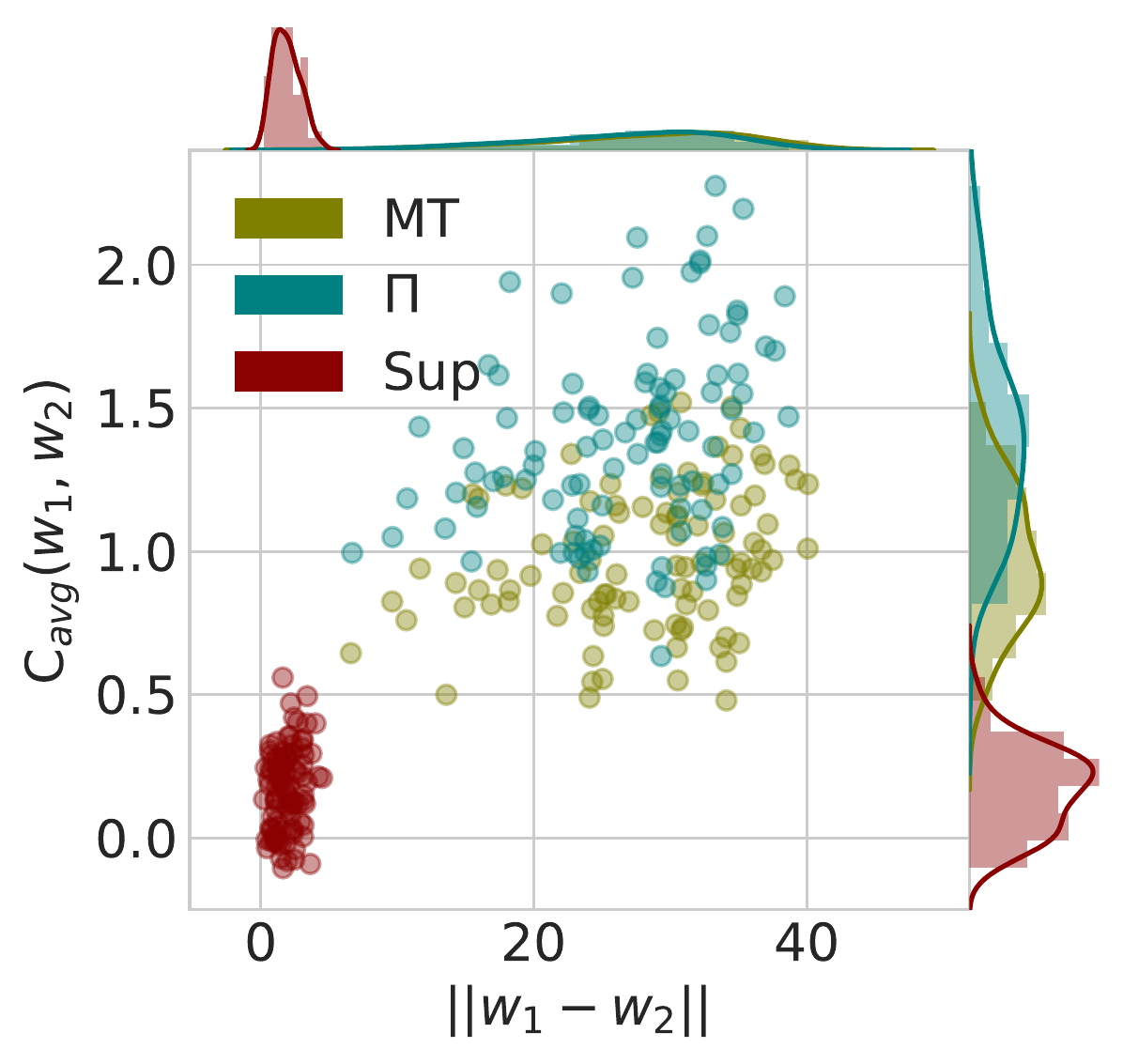}
      \vspace*{-0.4cm}\subcaption{$C_{\text{avg}}$ vs. $||w_1-w_2||$} \label{fig:weightaverage_dist}
\endminipage
& 
\hspace*{-0.2cm}
\minipage{0.25\textwidth}
       \centering
      \hspace*{-0.3cm}
      \includegraphics[height=3.5cm,trim={0 0 0 0}, clip]{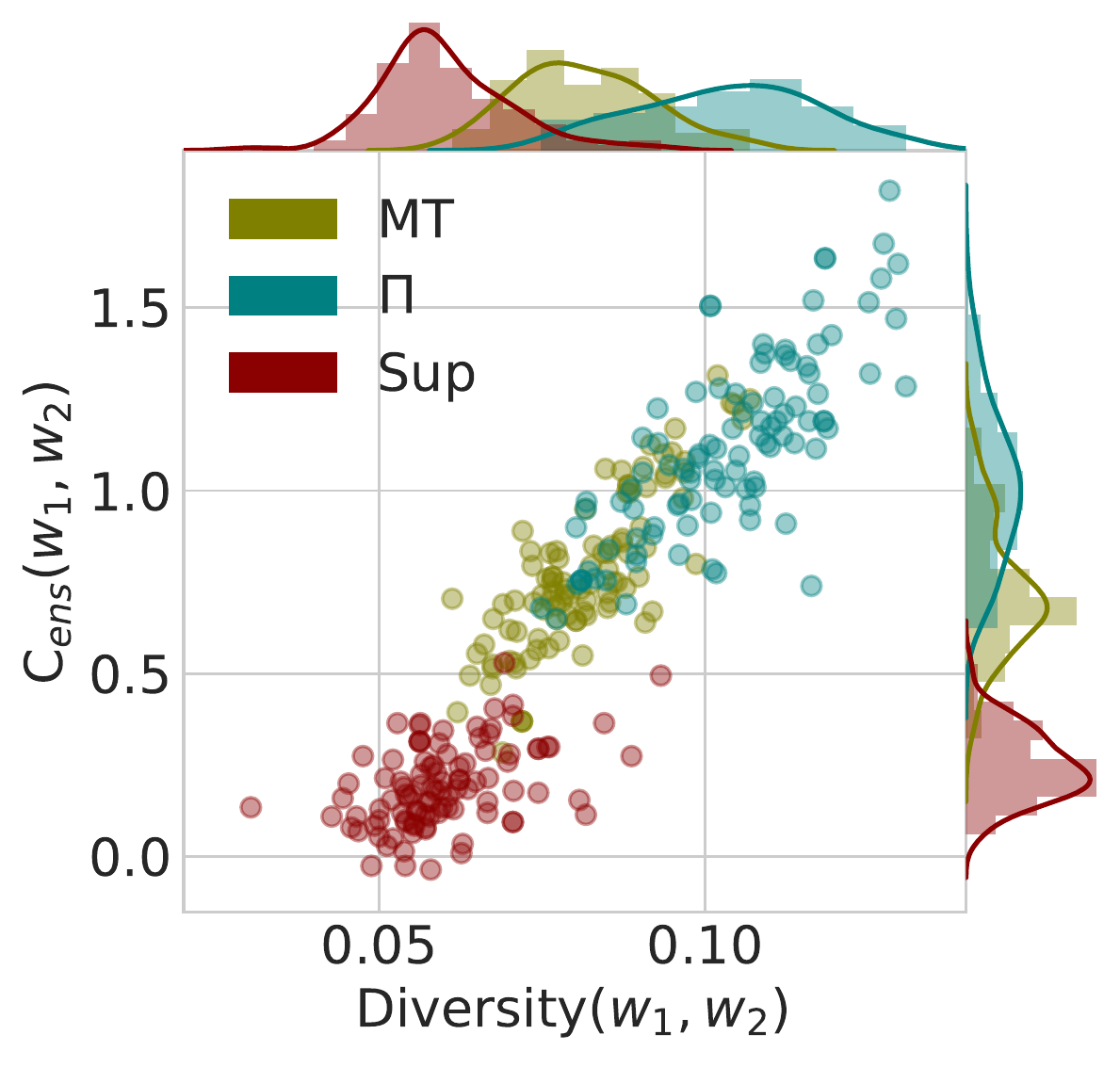}
     \vspace*{-0.4cm} \subcaption{$C_{\text{ens}}$ vs. diversity} \label{fig:ensemble_diversity}
\endminipage
&
\hspace*{-0.3cm}
\minipage{0.25\textwidth}
       \centering
      \hspace*{-0.1cm}
       \includegraphics[height=3.5cm,trim={150 250 190 240}, clip]{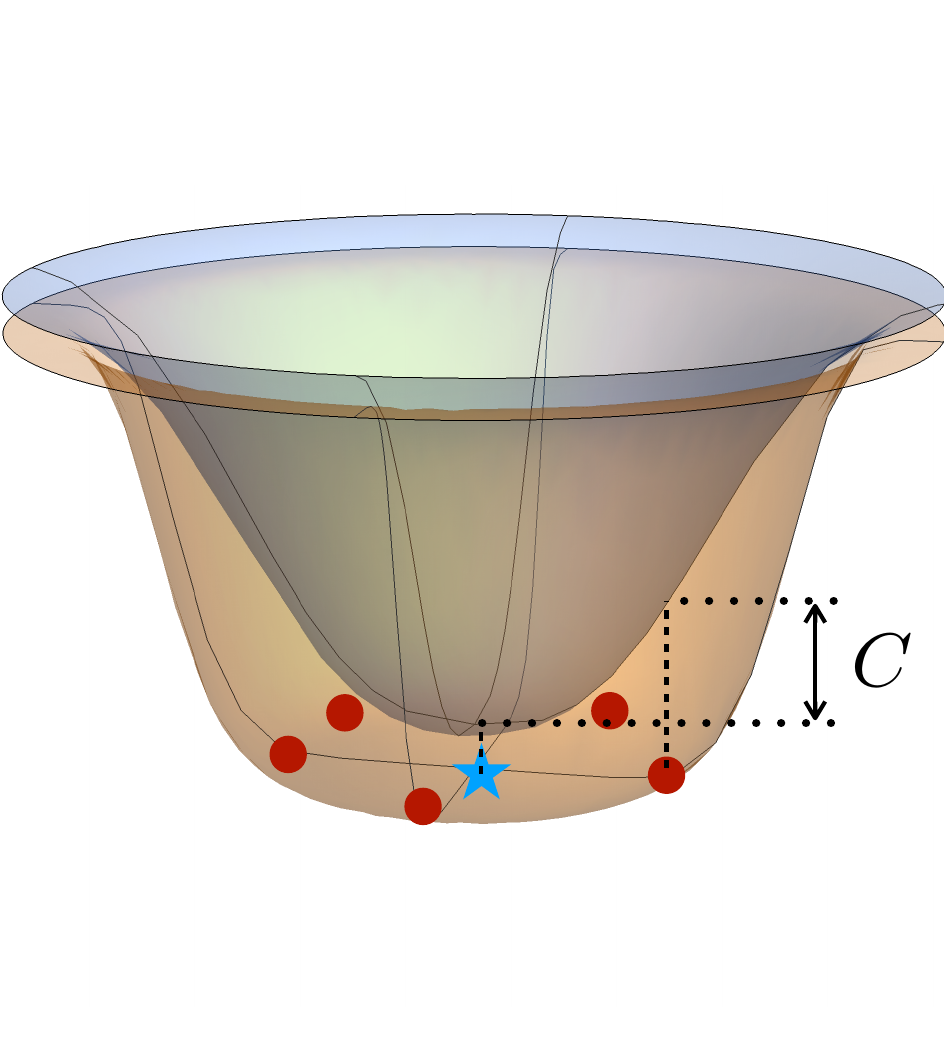}
      \vspace*{-0.0cm}
      \subcaption{Gain from flatness} \label{fig:err_surface}
\endminipage 
\end{tabular}
  \caption{
    \tb{(a)}: Illustration of a convex and non-convex function and Jensen's inequality.
    \tb{(b)}: Scatter plot of the decrease in error $C_{\text{avg}}$ for weight averaging versus distance. 
    \tb{(c)}: Scatter plot of the decrease in error $C_{\text{ens}}$ for prediction ensembling versus diversity.
    \tb{(d)}: Train error surface (orange) and Test error surface (blue). The SGD solutions (red dots) 
 around a locally flat minimum are far apart due to the flatness of the train surface (see Figure \ref{fig:sgd_sgd_rays}) which leads to large error reduction of the SWA solution (blue dot). 
  }
  \label{fig:convexity}
\end{figure}

\paragraph{Ensembling.} 
In Section \ref{sec:trajectories} we showed that 
the diversity in predictions is significantly larger for the \pi and Mean Teacher
models compared to purely supervised learning. 
The diversity of these iterates suggests that we can achieve greater benefits from ensembling. 
We use the same CNN architecture and hyper-parameters as in Section 
\ref{sec:trajectories} but extend the training time by doing $5$ learning rate 
cycles of $30$ epochs after the normal training ends at epoch $180$
(see \ref{sec:sup_architectures} and \ref{sec:sup_hypers} for details).
We sample random pairs of weights $w_1$, $w_2$ from epochs $180, 183, \hdots, 330$
and measure the error reduction from ensembling these pairs of models,
$
C_{\text{ens}} \equiv 
\frac 1 2 \textrm{Err}(w_1) + \frac 1 2 \textrm{Err}(w_2) - 
\textrm{Err}\left(\text{Ensemble}(w_1, w_2)\right)
$.
In Figure \ref{fig:ensemble_diversity} we visualize $C_{\text{ens}}$, 
against the diversity of the corresponding pair of models. 
We observe a strong correlation between the diversity in predictions of the 
constituent models and ensemble performance, and therefore $C_{\text{ens}}$ is substantially larger for \pi and Mean Teacher models.
As shown in \citet{swa}, ensembling can be well approximated by weight averaging if the weights are close by.

\paragraph{Weight Averaging.}
First, we experiment on averaging random pairs of weights at the end of training and analyze the performance with respect to the weight distances.
Using the the same pairs from above, we evaluate the performance of the model formed by averaging the pairs of weights,
  $C_{avg}(w_1,w_2) \equiv \frac 1 2 \textrm{Err}(w_1) + \frac 1 2 \textrm{Err}(w_2) - 
  \textrm{Err}\left(\frac 1 2 w_1 + \frac 1 2 w_2\right)$. 
Note that 
$C_{avg}$ is a proxy for convexity: if $C_{avg}(w_1, w_2) \ge 0$ for any pair
of points $w_1$, $w_2$, then by Jensen's inequality the error function
is convex (see the left panel of Figure \ref{fig:convexity}).
While the error surfaces for neural networks are known to be 
highly non-convex, they may be approximately convex
in the region traversed by SGD late into training \citep{goodfellow_loss}. In fact, in Figure \ref{fig:weightaverage_dist}, we find that the error surface of the SGD trajectory is approximately convex due to $C_{avg}(w_1, w_2)$ being mostly positive. 
Here we also observe that the distances between pairs of weights are much larger for the \pi and MT
models than for the supervised training; and as a result, weight averaging achieves a larger gain for these models. 

In Section \ref{sec:trajectories} we observed that for the \pi and Mean Teacher
models SGD traverses a large flat region of the weight space late in training.
Being very high-dimensional, this set has most of its volume concentrated
near its boundary. 
Thus, we find SGD iterates at the periphery of this flat
region (see Figure \ref{fig:err_surface}). 
We can also explain this behavior via the argument of \citep{mandt2017}.
Under certain assumptions SGD iterates can be thought of as samples 
from a Gaussian distribution centered at the minimum of the loss,
and samples from high-dimensional Gaussians are known to be concentrated on
the surface of an ellipse and never be close to the mean.
Averaging the SGD iterates
(shown in red in Figure \ref{fig:err_surface}) we can
move towards the center (shown in blue) of the flat region, stabilizing the SGD trajectory 
and improving the width of the resulting solution, and consequently improving generalization.

We observe that the improvement $C_{\text{avg}}$ from weight averaging ($1.2\pm 0.2\%$ over MT and \pi pairs) is on par or 
larger than the benefit $C_{\text{ens}}$ of prediction ensembling ($0.9\pm 0.2\%$) 
The smaller gain from ensembling might be due to the dependency of the ensembled solutions, since they are from the same SGD run as opposed to independent restarts as in typical ensembling settings.
For the rest of the paper, we focus attention on weight averaging because of its lower costs at test time and slightly higher performance compared to ensembling. 

\section{SWA and fast-SWA}
\label{sec:better_solutions_averaging}
\label{sec:swa}

In Section \ref{sec:analysis} we analyzed the training trajectories 
of the \pi, MT, and supervised models. We observed that the \pi and MT models
continue to actively explore the set of plausible solutions, producing diverse
predictions on the test set even in the late stages of training. Further, in section
\ref{sec:empirical_convexity} we have seen that 
averaging weights 
leads to significant gains in performance for the \pi and 
MT models. In particular these gains are much larger than in supervised
setting. 

Stochastic Weight Averaging (\swa) \citep{swa} is a recent approach that is 
based on averaging weights traversed by SGD with a modified learning rate schedule.
In Section \ref{sec:analysis} we analyzed averaging pairs of weights corresponding
to different epochs of training and showed that it improves the test accuracy.
Averaging multiple weights reinforces this effect, and
SWA was shown to significantly improve generalization performance in supervised learning.
Based on our results in section \ref{sec:empirical_convexity}, we can expect 
even larger improvements in generalization when applying SWA to the \pi and MT models.

\begin{figure}[t]
\centering
\centering
\hspace*{-1.0cm}
\begin{subfigure}[t]{0.4\linewidth}
	\includegraphics[trim={0 200 0 280},width=\linewidth,valign=t, clip=True]{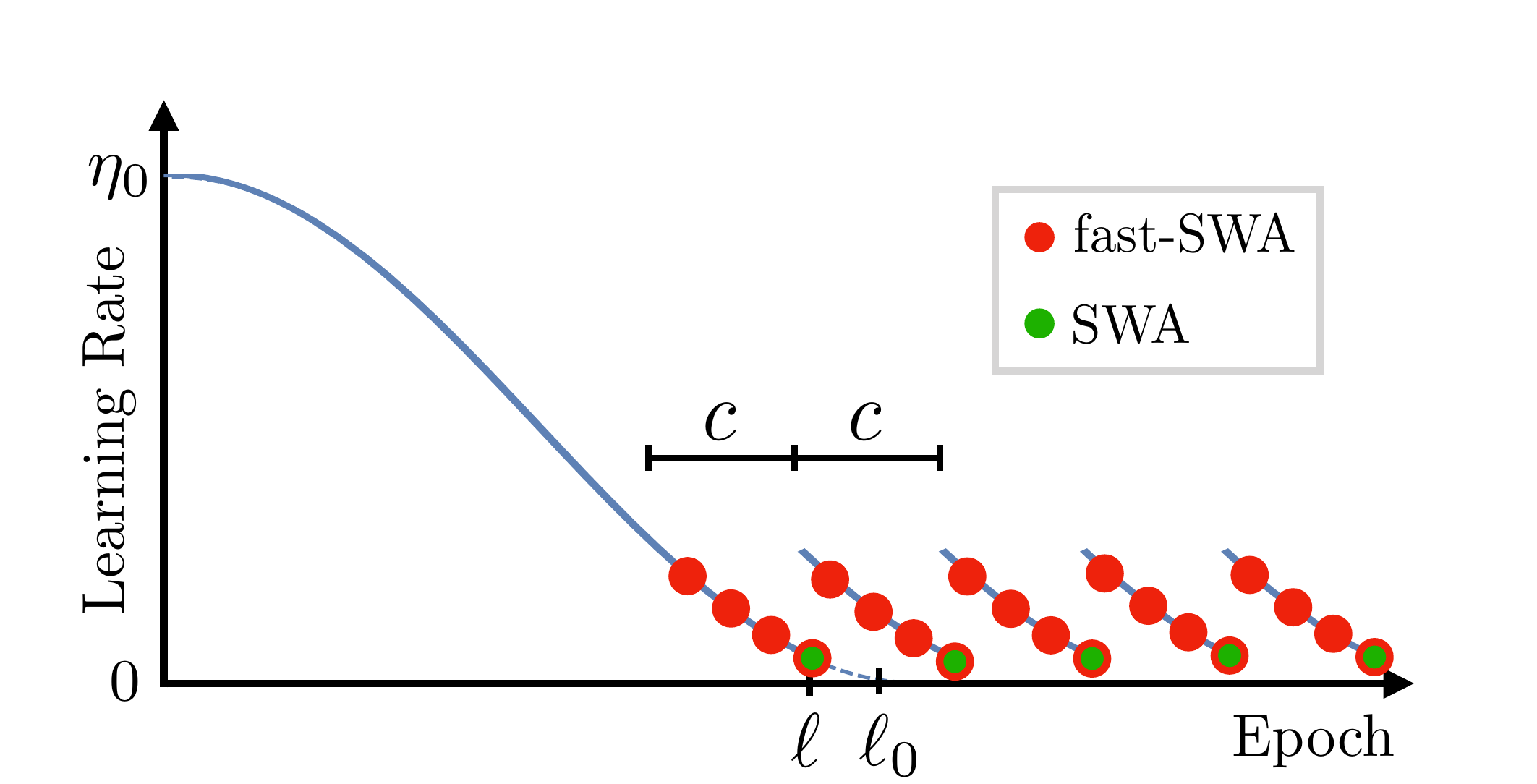}
\end{subfigure}
\begin{subfigure}[t]{0.28\linewidth}
	\includegraphics[trim={0 200 0 200},width=\linewidth,valign=t, clip=True]{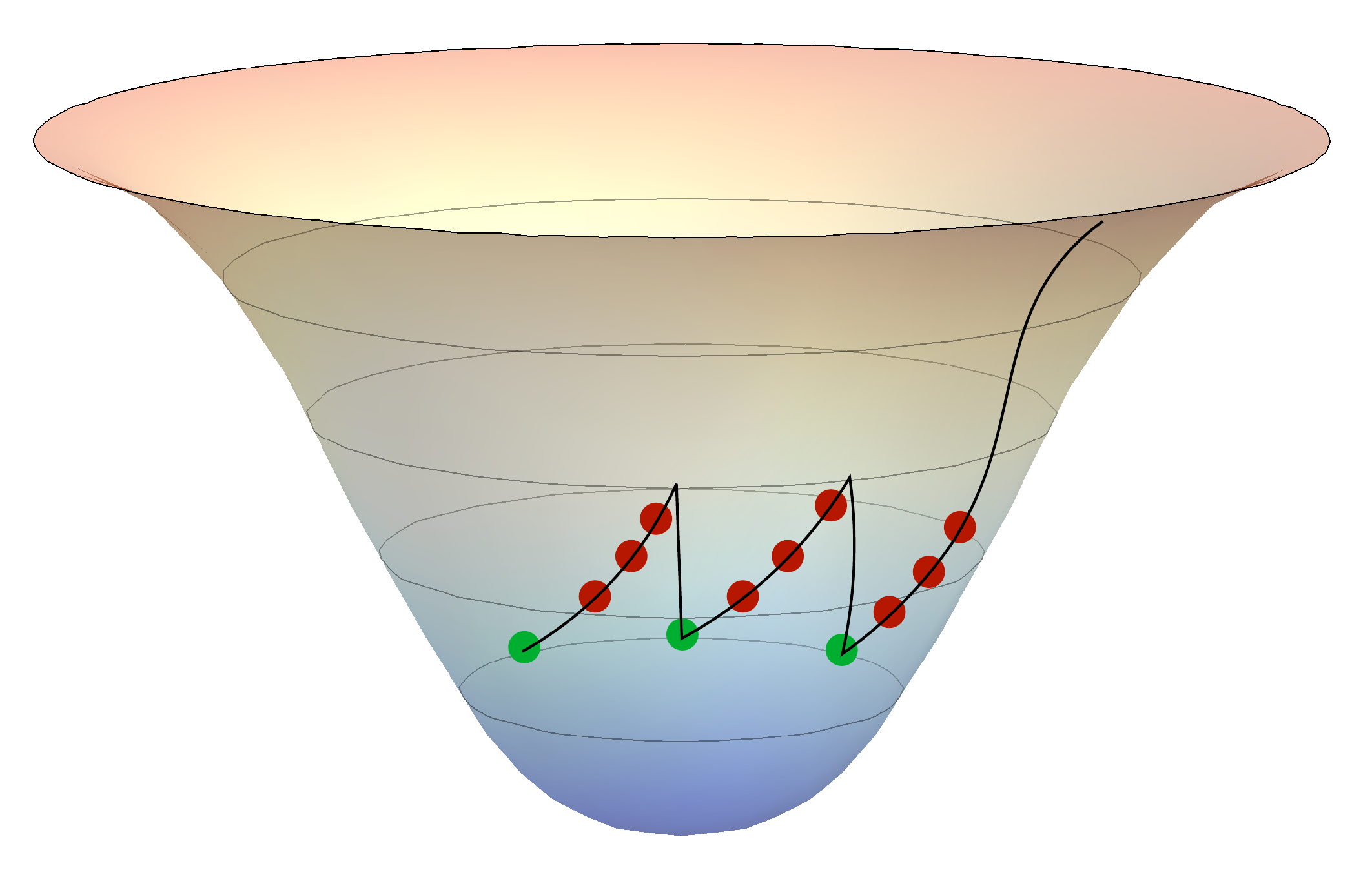}
\end{subfigure}
\hspace*{1.0cm}
\begin{subfigure}[t]{0.20\linewidth}
	\includegraphics[trim={150 250 150 250},width=\linewidth,valign=t, clip=True]{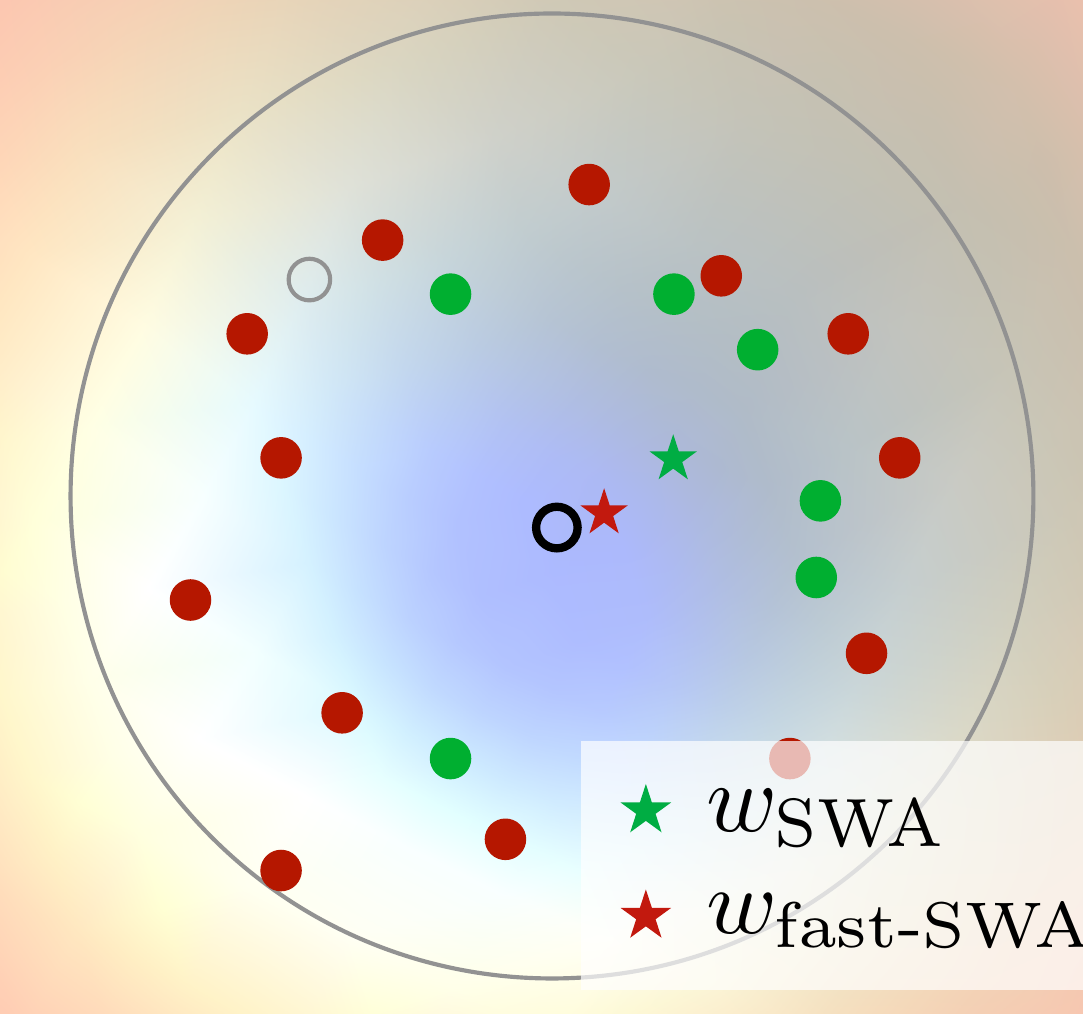}
\end{subfigure}
  \caption{ {\bf Left}: Cyclical cosine learning rate schedule and 
  \swa and \fastswa averaging strategies.
  {\bf Middle}: Illustration of the solutions explored by the cyclical cosine annealing schedule on an error surface.
  {\bf Right}: Illustration of \swa and \fastswa averaging strategies. \fastswa
	averages more points but the errors of the averaged points, as indicated by the heat color, are higher.
  }
  \label{fig:swa_model}
\end{figure}

\swa typically starts from a pre-trained model, and then averages 
points in weight space traversed by SGD with a constant or cyclical learning rate.
We illustrate the cyclical cosine learning rate schedule in Figure \ref{fig:swa_model} (left) and the SGD solutions explored in Figure \ref{fig:swa_model} (middle). 
For the first $\ell \le \ell_0$ epochs the network is pre-trained using the cosine
annealing schedule where the learning rate at epoch $i$ is set equal to 
$\eta(i) = 0.5 \cdot \eta_0 (1 +  \text{cos} \left( \uppi \cdot i / \ell_0 \right) )$. After $\ell$ epochs, we use a cyclical 
schedule, repeating the learning rates from epochs $[\ell - c, \ell]$, where
$c$ is the cycle length.
SWA collects the 
networks corresponding to the minimum values of the learning 
rate (shown in green in Figure \ref{fig:swa_model}, left) and averages their 
weights. The model with the averaged weights $w_{\text{SWA}}$ is then used
to make predictions. 
We propose to apply \swa to the student network both for the \pi and Mean
Teacher models. 
Note that the \swa weights do not interfere with training. 

Originally, \citet{swa} proposed using cyclical learning rates with small cycle length for \swa. However, as we have seen in Section \ref{sec:empirical_convexity}
(Figure \ref{fig:convexity}, left) the benefits of averaging are the most
prominent when the distance between the averaged points is large. 
Motivated by this observation, we instead use longer learning rate cycles $c$.  
Moreover, \swa updates the average weights only once per cycle, which 
means that many additional training epochs are needed in order to collect enough 
weights for averaging.
To overcome this limitation, we propose \tb{\fastswa},
a modification of \swa that averages networks
corresponding to every $k < c$ epochs starting from epoch $\ell - c$. 
We can also average multiple weights within a single epoch setting $k<1$. 

Notice that most of the models included in the \fastswa average (shown in red in Figure 
\ref{fig:swa_model}, left) have higher errors than those included in the \swa average
(shown in green in Figure \ref{fig:swa_model}, right)
since they are obtained when the learning rate is high. It is our contention that 
including more models in the \fastswa weight average can more than compensate for the larger
errors of the individual models.
Indeed, our experiments in Section \ref{sec:exps} show that \fastswa  converges 
substantially faster than \swa and has lower performance variance. We analyze this 
result theoretically in Section \ref{sec:theory_fastswa}).

\section{Experiments} \label{sec:exps}

We evaluate the \pi  and MT models 
(Section \ref{sec:swa}) on CIFAR-10 and CIFAR-100 with varying numbers of 
labeled examples.
We show that  \fastswa and \swa improve the 
performance of the \pi and MT models, 
as we expect from our observations in Section \ref{sec:analysis}.
In fact, in many cases \fastswa improves on the best results reported in the 
semi-supervised literature. 
We also demonstrate that the preposed \fastswa obtains high performance much faster than \swa. 
We also evaluate \swa applied to a consistency-based 
domain adaptation model \citep{da_mt}, closely related to the MT model,
for adapting CIFAR-10 to STL.  We improve the best reported 
test error rate for this task from $19.9\%$ to $16.8\%$. 

We discuss the experimental setup in Section \ref{sec:exp_setup}.
We provide the results for CIFAR-10 and CIFAR-100 datasets in Section 
\ref{sec:cifar10} and \ref{sec:cifar100}. 
We summarize our results in comparison to the best previous 
results in Section \ref{sec:soa}. We show several additional results and 
detailed comparisons in Appendix \ref{sec:add_results}. 
We provide analysis of train and test error surfaces of \fastswa solutions along the directions connecting \fastswa and SGD in Section \ref{sec:sup_rayplots}.

\subsection{Setup} \label{sec:exp_setup}

We evaluate the weight averaging methods \swa and \fastswa on different network 
architectures and learning rate schedules. We are able to improve on the base models in all settings. 
In particular, we consider a $13$-layer CNN and 
a 12-block (26-layer) Residual Network \citep{resnet} with Shake-Shake regularization \citep{shake}, which we refer to simply as \emph{CNN} and \emph{Shake-Shake} respectively (see Section \ref{sec:sup_architectures} for details on the architectures). 
For training all methods we use 
the stochastic gradient descent (SGD) optimizer with
the cosine annealing learning rate described in Section \ref{sec:swa}. 
We use two learning rate schedules, the \emph{short schedule} with $\ell=180, \ell_0=210, c=30$, similar to the experiments in \citet{mt}, and the \emph{long schedule} with $\ell=1500, \ell_0=1800, c=200$, similar to the experiments in \citet{shake}. We note that the long schedule improves the performance of the base models compared to the short schedule; however, \swa can still further improve the results. 
See Section \ref{sec:sup_hypers} of the Appendix for more details on other hyperparameters.
We repeat each CNN experiment $3$ times with different random seeds to estimate
the standard deviations for the results in the Appendix.

\subsection{CIFAR-10} \label{sec:cifar10_cnn} \label{sec:cifar10}

\begin{figure}[h]
  \centering
\minipage{0.25\textwidth}
       \centering
       \centering
       \includegraphics[clip, trim={0 0 0 0},width=1.0\linewidth]{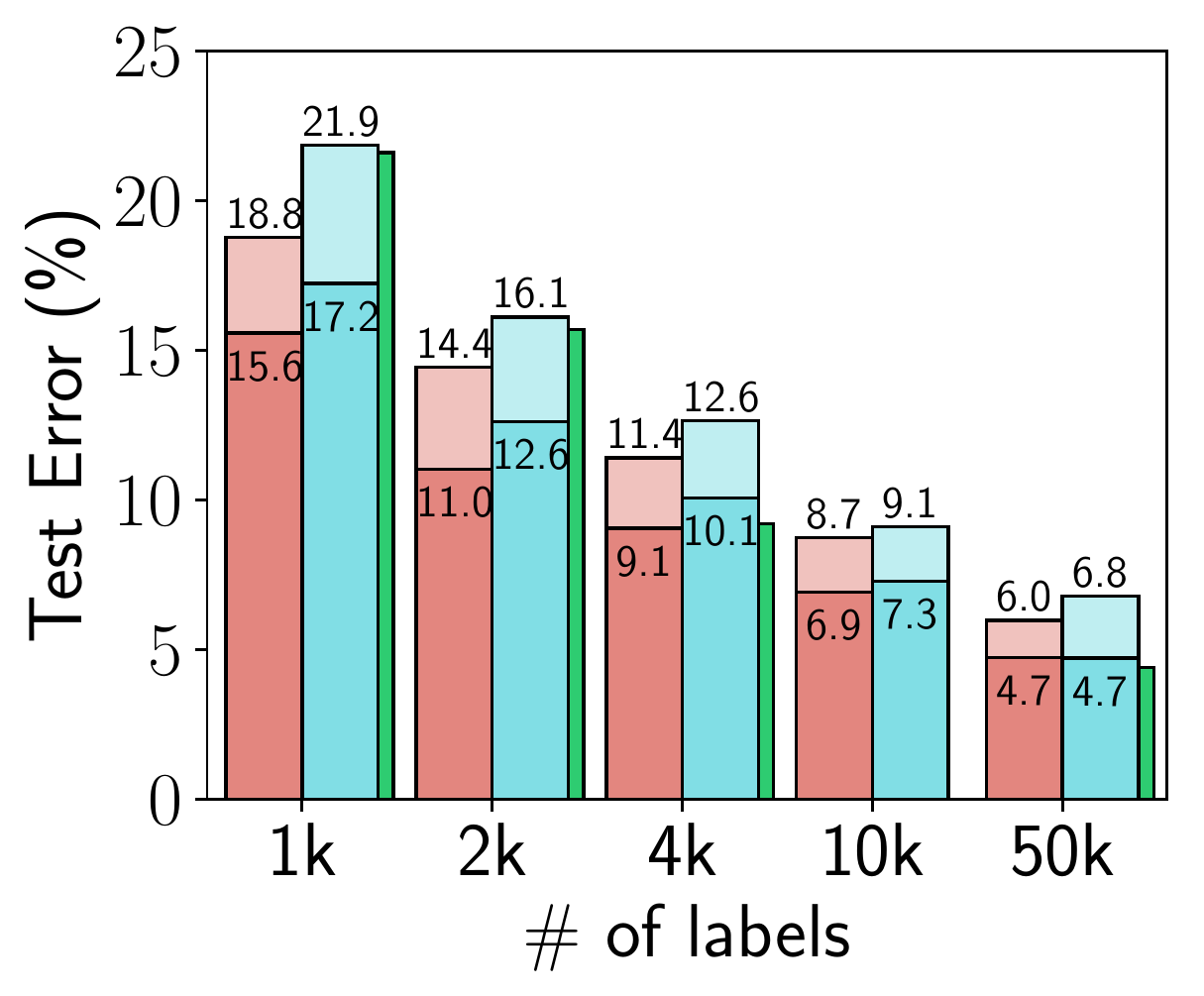}
   \vspace{-1.5\baselineskip}
  \subcaption{\label{fig:c10_cnn}}
\endminipage\hfill
\minipage{0.25\textwidth}
       \centering
       \centering
       \includegraphics[clip, trim={0 0 0 0},width=1.02\linewidth]{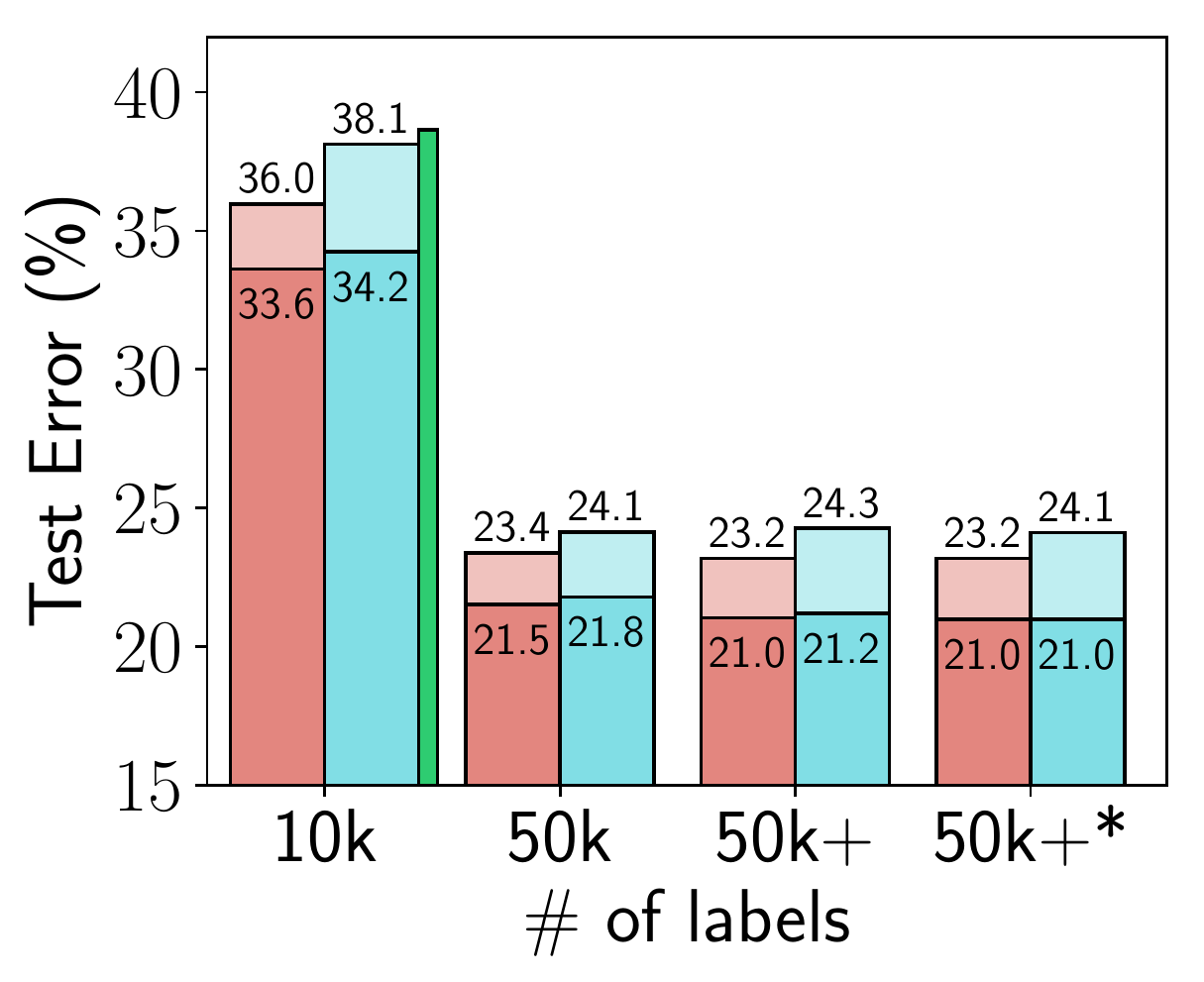}
   \vspace{-1.5\baselineskip}
  \subcaption{\label{fig:c100_cnn}}
  \endminipage\hfill
  \minipage{0.25\textwidth}
\centering
    \includegraphics[clip, trim={0 0 0 0}, width=1.02\linewidth]{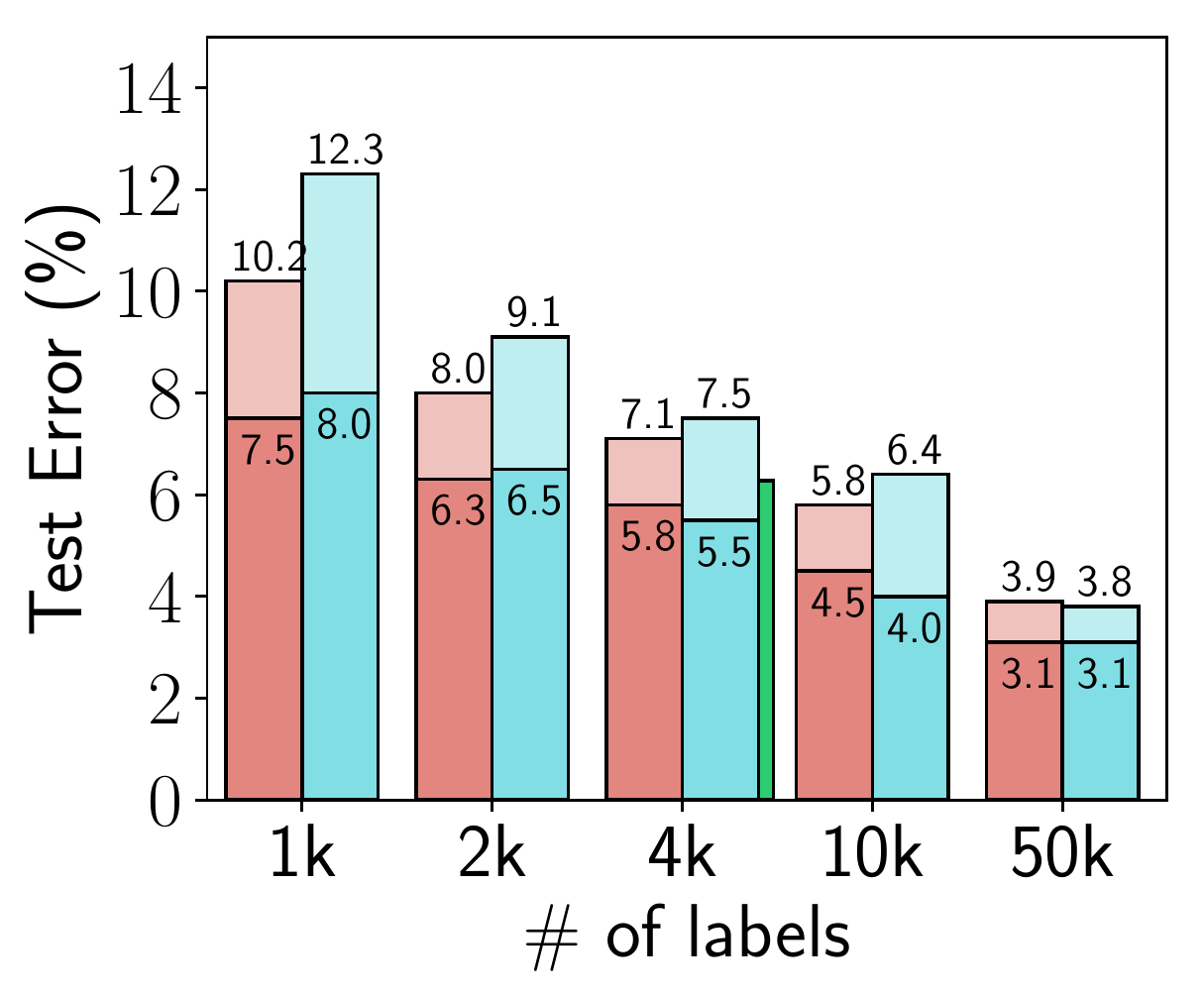}
    \vspace{-1.5\baselineskip}
 \subcaption{\label{fig:c10_shake_short}}
\endminipage\hfill
       \minipage{0.25\textwidth}
       \centering
       \centering
       \includegraphics[clip, trim={0 0 0 0},width=1.02\linewidth]{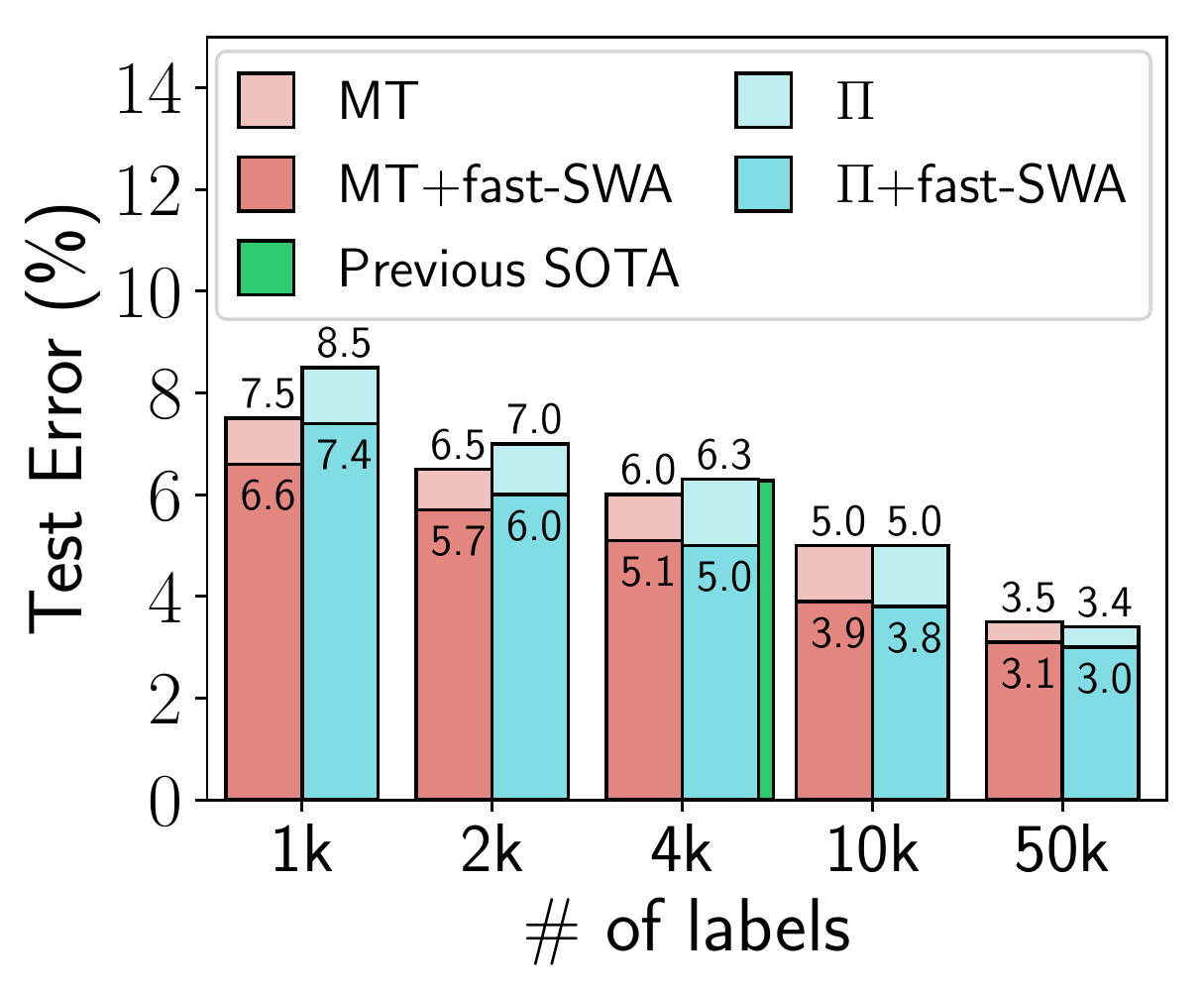}
  \vspace{-1.5\baselineskip}
  \subcaption{\label{fig:c10_shake_long}}
\endminipage\hfill
  \caption{ Prediction errors of \pi and MT models with and without \fastswa. \tb{(a)} CIFAR-10 with CNN 
    \tb{(b)} CIFAR-100 with CNN. $50k+$ and $50k+^*$ correspond to $50k$+$500k$ and $50k$+$237k^*$ settings
  \tb{(c)} CIFAR-10 with ResNet + Shake-Shake using the short schedule
  \tb{(d)} CIFAR-10 with ResNet + Shake-Shake using the long schedule.
  }
  \label{fig:barplots}
\end{figure}

We evaluate the proposed \fastswa method using the \pi and MT
models on the CIFAR-10 dataset \citep{krizhevsky_learning_nodate}. 
We use $50k$ images for training with  $1k$, $2k$, $4k$, $10k$ and $50k$ labels and report the top-1 errors on the test set ($10k$ images). 
We visualize the results for the CNN and Shake-Shake architectures in Figures
\ref{fig:c10_cnn}, \ref{fig:c10_shake_short}, and \ref{fig:c10_shake_long}.
For all quantities of labeled data,
\fastswa substantially improves test accuracy in both architectures. 
Additionally, in Tables \ref{table:cifar10_cnn13_additional}, 
\ref{table:cifar10_shakeshake} of the Appendix 
we provide a thorough comparison of different averaging strategies
as well as results for VAT \citep{vat2}, TE \citep{te}, and other baselines.

Note that we applied \fastswa for VAT as well which is another popular approach for semi-supervised learning. We found that the improvement on VAT is not drastic -- our base implementation obtains $11.26\%$ error where \fastswa reduces it to $10.97\%$ (see Table \ref{table:cifar10_cnn13_additional} in Section \ref{sec:add_results}). 
It is possible that the solutions explored by VAT are not as diverse as in \pi and MT models due to VAT loss function.
Throughout the experiments, we focus on the \pi and MT models as they have been shown to scale to powerful networks such as Shake-Shake and obtained previous state-of-the-art performance. 

\begin{figure}[h!]
\centering
\vspace{-0.4\baselineskip}
\minipage{0.32\textwidth}
       \centering
       \includegraphics[clip, trim={5 0 50 60},width=1.0\linewidth]{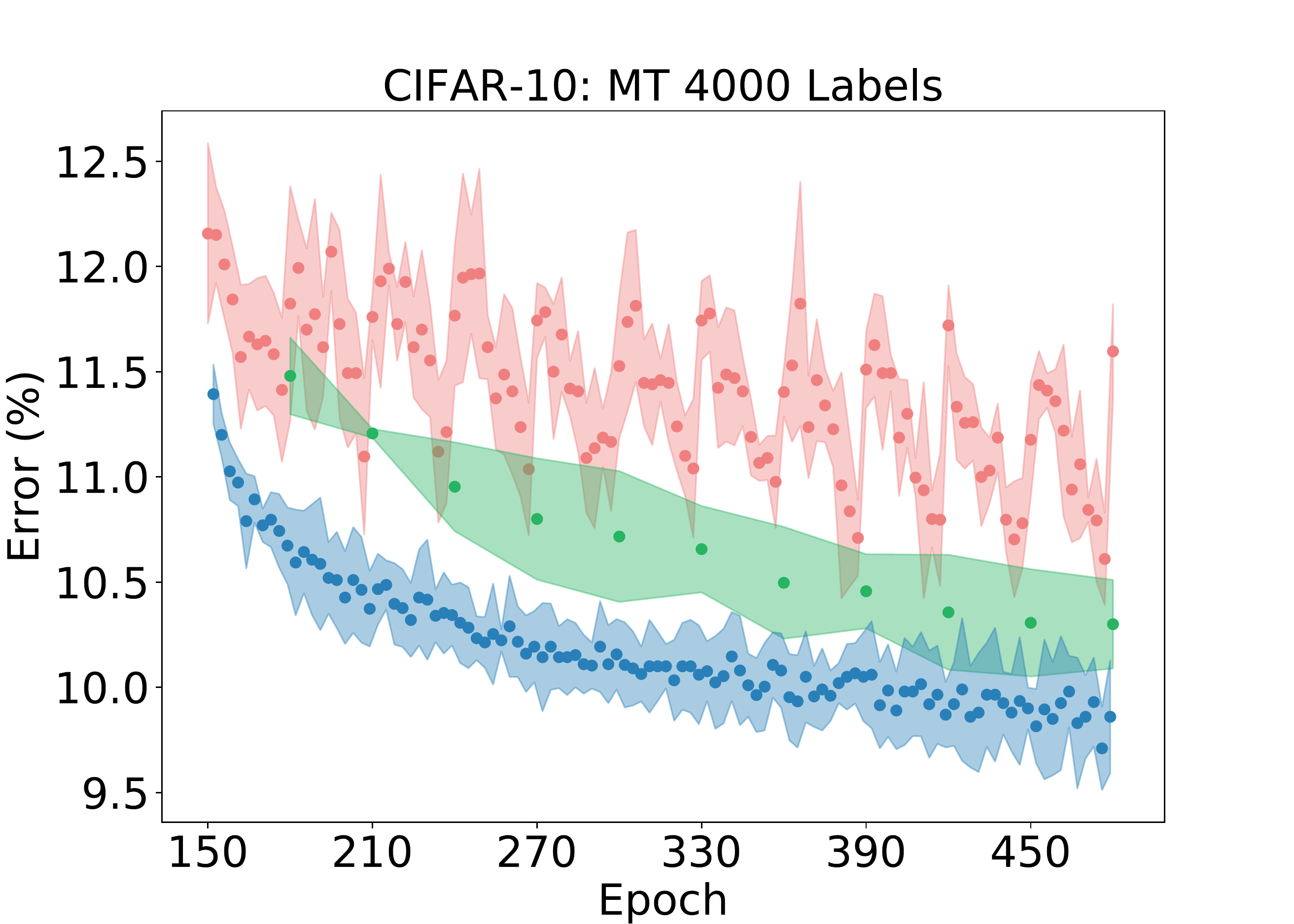}
\endminipage \hfill
\minipage{0.32\textwidth}
       \centering
        \includegraphics[clip, trim={15 0 50 60},width=1.0\linewidth]{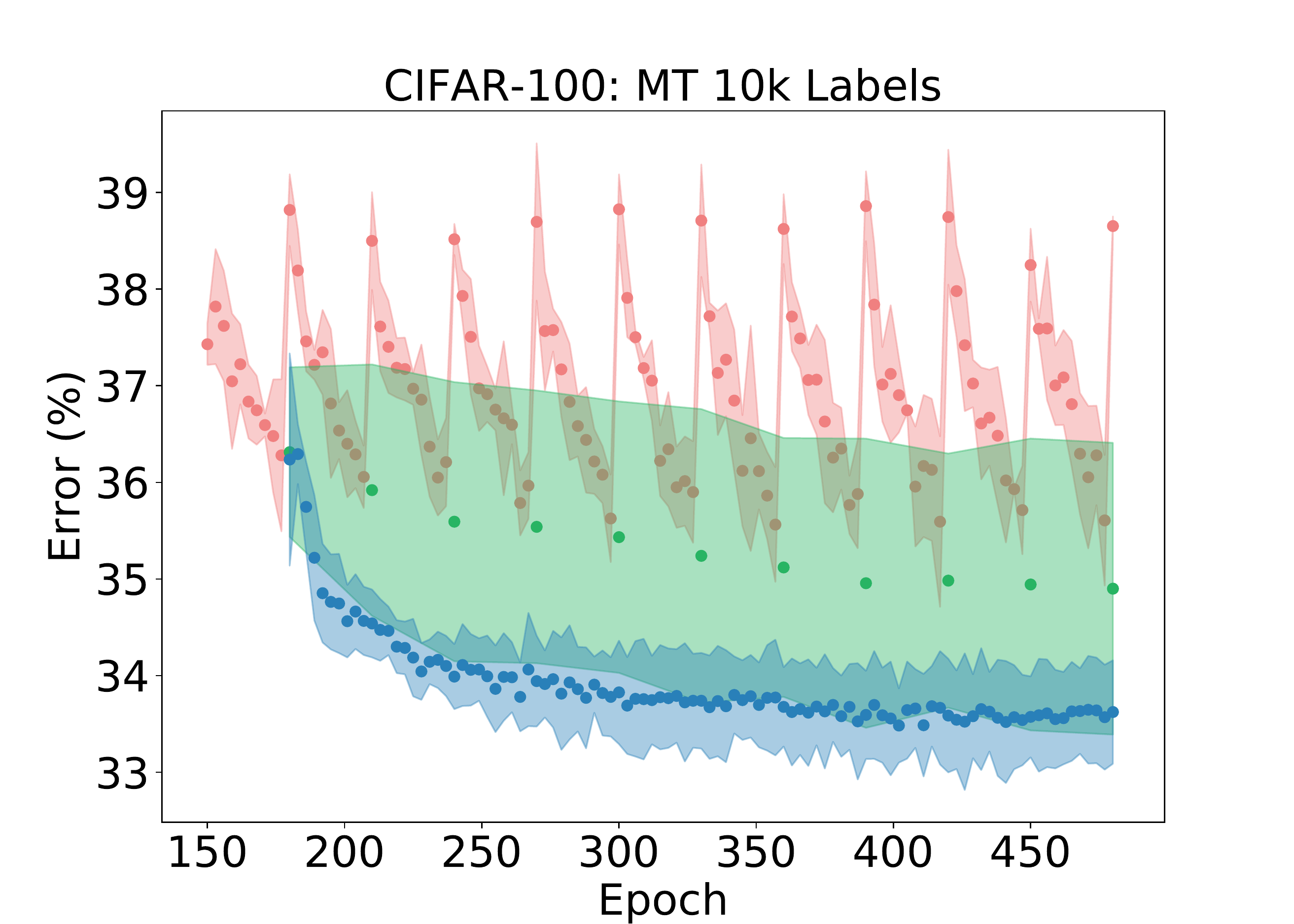}
\endminipage\hfill
\minipage{0.32\textwidth}
       \centering
        \includegraphics[clip, trim={15 0 50 60},width=1.0\linewidth]{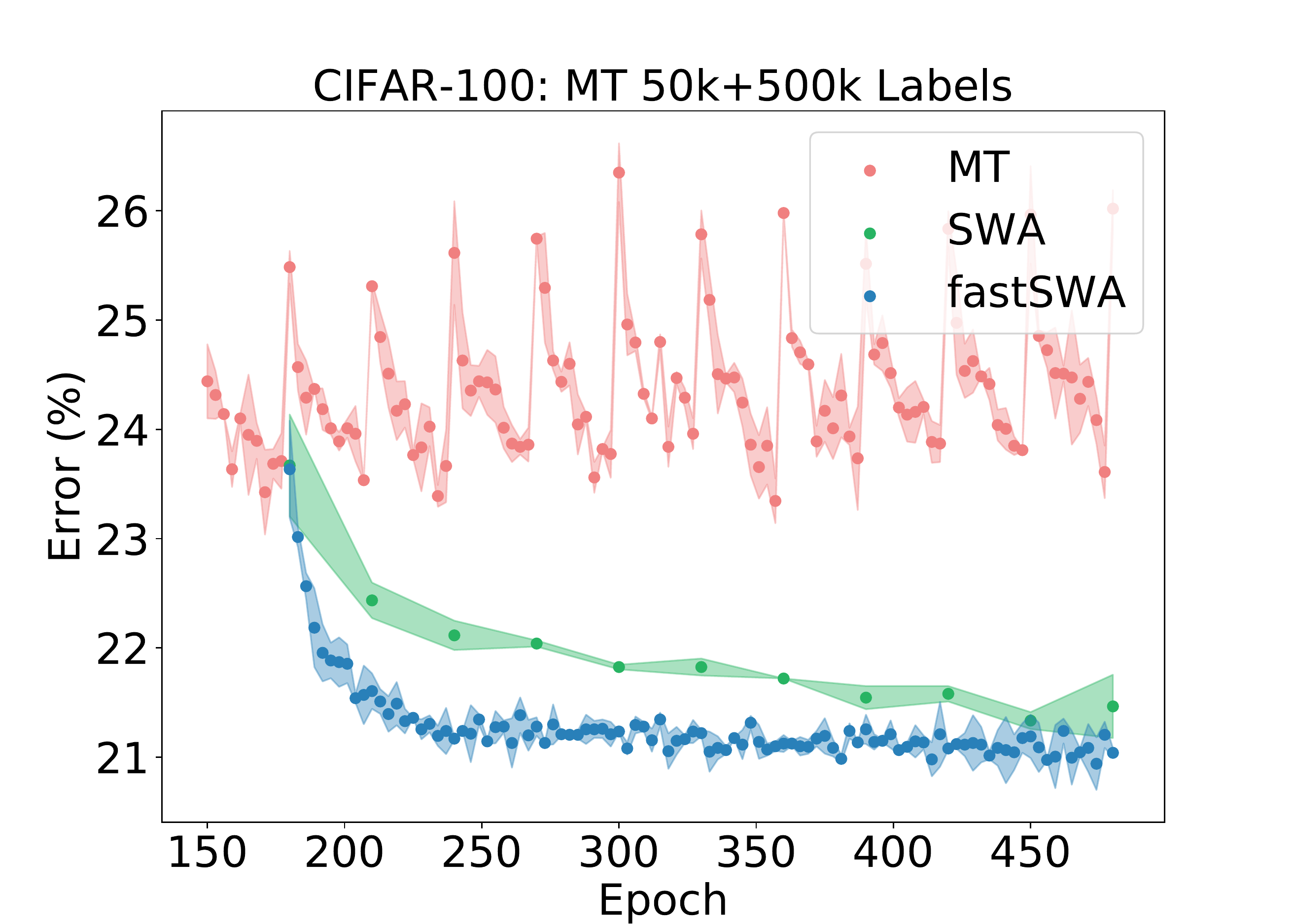}
\endminipage\hfill
  \caption{Prediction errors of base models and their weight averages (\fastswa and \swa) for CNN 
  on \textbf{(left)} CIFAR-10 with $4k$ labels,
  \textbf{(middle)} CIFAR-100 with $10k$ labels,
  and \textbf{(right)} CIFAR-100 $50k$ labels and extra $500k$ unlabeled data from Tiny Images \citep{tiny_images}. 
  }
  \label{fig:cnn_c10_c100_trend}
\end{figure}

In Figure \ref{fig:cnn_c10_c100_trend} (left), we visualize the test
error as a function of iteration using the CNN.
We observe that when the cyclical learning rate starts after epoch $\ell=180$, 
the base models drop in performance due to the sudden increase in learning rate (see Figure \ref{fig:swa_model} left).
However, \fastswa continues to improve while collecting the weights corresponding to high learning rates for averaging.
In general, we also find that the cyclical learning rate improves the base models beyond the usual cosine annealing 
schedule and increases the performance of \fastswa as training progresses. 
Compared to \swa, we also observe that \fastswa converges substantially faster, 
for instance, reducing the error to $10.5\%$  at epoch $200$ while \swa attains 
similar error at epoch $350$ for CIFAR-10 $4k$ labels (Figure 
\ref{fig:cnn_c10_c100_trend} left). We provide additional plots in Section \ref{sec:add_results} 
showing the convergence of the \pi and MT models in all label 
settings, where we observe similar trends that \fastswa results in faster error reduction. 

We also find that the performance gains of \fastswa over base models are higher for the \pi model compared to the MT model, which is consistent with the convexity observation in Section \ref{sec:empirical_convexity} 
and Figure \ref{fig:convexity}.
In the previous evaluations \citep[see e.g.][]{ssl-eval, mt}, the \pi model was shown to be inferior to the MT model.
However, with weight averaging, \fastswa reduces the gap between \pi and MT performance. 
Surprisingly, we find that the \pi model can outperform MT after applying \fastswa  
with moderate to large numbers of labeled points. 
In particular, the $\Pi$+fast-SWA model outperforms MT+\fastswa on CIFAR-10 with $4k$, $10k$, and $50k$ labeled
data points for the Shake-Shake  architecture.

\vspace{-0.8\baselineskip}

\subsection{CIFAR-100 and Extra Unlabeled Data} \label{sec:cifar100_cnn} \label{sec:cifar100}

We evaluate the \pi and MT models with \fastswa on 
CIFAR-100.
We train our models using $50000$ images with $10k$ and $50k$ labels using the $13$-layer CNN. 
We also analyze the effect of using the Tiny Images dataset \citep{tiny_images} as an additional source of unlabeled data.

The Tiny Images dataset consists of $80$ million images, mostly unlabeled, and contains CIFAR-100 as a subset.       
Following \citet{te}, we use two settings of unlabeled data, $50k$+$500k$ and $50k$+$237k^*$ 
where the $50k$ images corresponds to CIFAR-100 images from the training set and the $+500k$ 
or $+237k^*$ images corresponds to additional $500k$ or $237k$ images from the Tiny Images dataset. 
For the $237k^*$ setting, we select only the images that belong to the classes in CIFAR-100, 
corresponding to $237203$ images. 
For the $500k$ setting, we use a random set of $500k$ images whose classes can be different from CIFAR-100. 
We visualize the results in Figure \ref{fig:c100_cnn}, where we again observe that
\fastswa substantially improves performance for every configuration of the number of labeled and unlabeled data.
In Figure \ref{fig:cnn_c10_c100_trend} (middle, right) we show the errors of MT, \swa and \fastswa as a
function of iteration on CIFAR-100 for the $10k$ and $50k$+$500k$ label settings. 
Similar to the CIFAR-10 experiments, we observe that \fastswa reduces the errors substantially faster than \swa.
We provide detailed experimental results in Table \ref{table:cifar100} of the
Appendix and include preliminary results using the Shake-Shake architecture in Table \ref{table:cifar100_shakeshake}, Section \ref{sec:add_results}.

\subsection{Advancing State-of-the-Art} \label{sec:soa}

We have shown that \fastswa can significantly improve the performance of both the \pi and 
MT models. We provide a summary comparing our results with the previous best results in the literature in Table  \ref{table:soa}, using the $13$-layer CNN and
the Shake-Shake architecture that had been applied previously. We also provide detailed results the Appendix \ref{sec:add_results}.

\begin{table}[h]
  \caption{Test errors against current state-of-the-art semi-supervised results. 
  The previous best numbers are obtained from \citep{mt}$^1$,  \citep{vad}$^2$, \citep{te}$^3$ and \citep{sntg}$^4$.
  CNN denotes performance on the benchmark $13$-layer CNN (see \ref{sec:sup_architectures}). 
Rows marked $^{\dagger}$ use the Shake-Shake architecture.
The result marked $^\ddagger$ are from \pi + \fastswa, where the rest are based on MT + \fastswa. 
The settings $50k$+$500k$ and $50k$+$237k^*$ use additional $500k$ and $237k$ unlabeled data from the Tiny Images dataset \citep{tiny_images} where $^*$ denotes that we use only the images that correspond to CIFAR-100 classes.
  } \label{table:soa}
  \centering
  \begin{tabular}{c  c c c | c c c c}
  \toprule
Dataset 			& 	\multicolumn{3}{c}{CIFAR-10}  & \multicolumn{3}{c}{CIFAR-100} \\
 	\cmidrule(r){2-7}	
No. of Images   	&	50k & 50k & 50k 	
				& 	50k 
				&	50k+500k	&	50k+237k$^*$		\\
No. of Labels   		&	1k	& 	2k	&	4k
				& 	10k 	
				&	50k	
				&	50k		\\
\midrule
Previous Best CNN
				& 	18.41$^4$  
				&	13.64$^4$  
				&	9.22$^2$ 		
				&	38.65$^3$ 	
				&	23.62$^3$ 	
				&	23.79$^3$ 	
				\\
Ours CNN
				& 	15.58 	 	
				& 	11.02       	 	
				&	9.05     		
				&	33.62		
				&	21.04 		
				&	20.98 		
				\\
				\cmidrule(r){2-7}
Previous Best$^\dagger$
				& 				
				& 				
				&	6.28$^1$			
				&				
				&				
				&				
				\\

Ours$^\dagger$
				& 	6.6  			
				& 	5.7  			
				&	5.0$^\ddagger$ 	
				&	28.0 			
				&	19.3 			
				& 	17.7 			
				\\
 \bottomrule
  \end{tabular}
\end{table}

\subsection{Preliminary Results on Domain Adaptation}
\label{sec:domain_adaptation}
Domain adaptation problems involve learning using a source domain $X_s$ 
equipped with labels $Y_s$ and performing classification on the target 
domain $X_t$ while having no access to the target labels at training time. 
A recent model by \citet{da_mt} applies the consistency enforcing principle for 
domain adaptation and achieves state-of-the-art results on many datasets. 
Applying \fastswa to this model on domain adaptation from CIFAR-10 to STL we 
were able to improve the best results reported in the literature from
$19.9\%$ to $16.8\%$. 
See Section \ref{sec:exp_da} for more details on the domain adaptation experiments.

\section{Discussion} \label{sec:conclusion}

Semi-supervised learning is crucial for reducing the dependency of deep learning on 
large labeled datasets.  Recently, there have been great advances in semi-supervised 
learning, with consistency regularization models achieving the best known results. 
By analyzing solutions along the training trajectories for two of the most successful 
models in this class, the \pi and Mean Teacher models, 
we have seen that rather than converging to a single solution SGD continues to 
explore a diverse set of plausible solutions late into training. As a result, we can expect that
averaging predictions or weights will lead to much larger gains in performance than for supervised training. 
Indeed, applying a variant of the recently proposed stochastic weight averaging (SWA)
we advance the best known semi-supervised results 
on classification benchmarks.  

While not the focus of our paper, we have also shown that weight averaging
has great promise in domain adaptation \citep{da_mt}. 
We believe that application-specific
analysis of the geometric properties of the training objective and optimization
trajectories will further improve results over a wide range of application specific
areas, including reinforcement learning with sparse rewards, generative adversarial 
networks \citep{gan_averaging}, or semi-supervised natural language processing.

\normalsize{
\normalsize{
\bibliographystyle{abbrvnat}
\bibliography{semisup}
}
}

\clearpage
\appendix 
\section{Appendix} \label{sec:supplementary}

\subsection{Additional Plots} \label{sec:sup_rayplots}

\begin{figure}[h]
  \centering
  \begin{tabular}[c]{c c c c}
  \hspace*{-0.4cm}
\minipage{0.32\textwidth}
       \centering
      \hspace*{-0.3cm}
      \includegraphics[width=0.96\textwidth,trim={0 0 0 0}, clip]{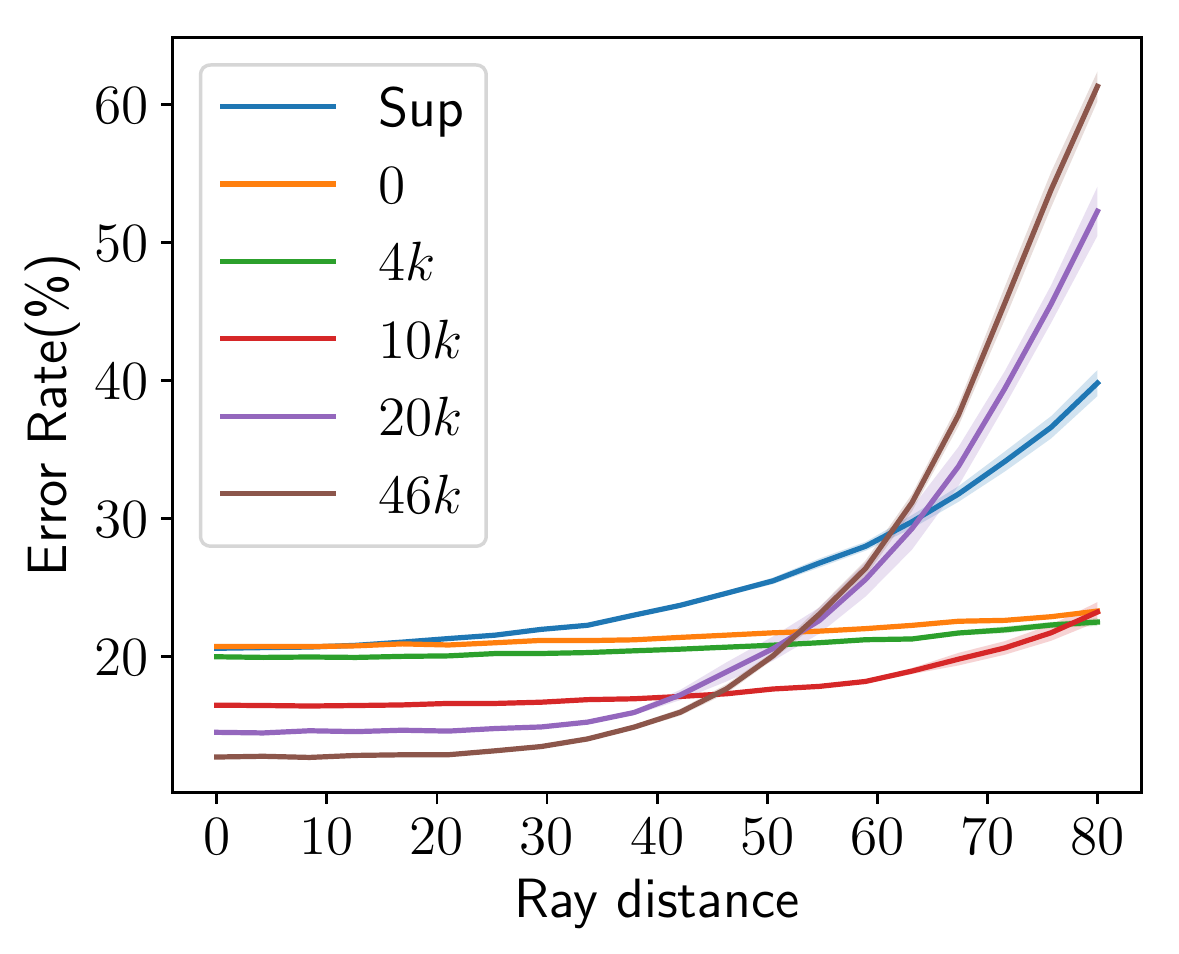}
\endminipage
& 
\hspace*{-0.4cm}
\minipage{0.32\textwidth}
       \centering
      \hspace*{-0.3cm}
      \includegraphics[width=0.96\textwidth,trim={0 0 0 0}, clip]{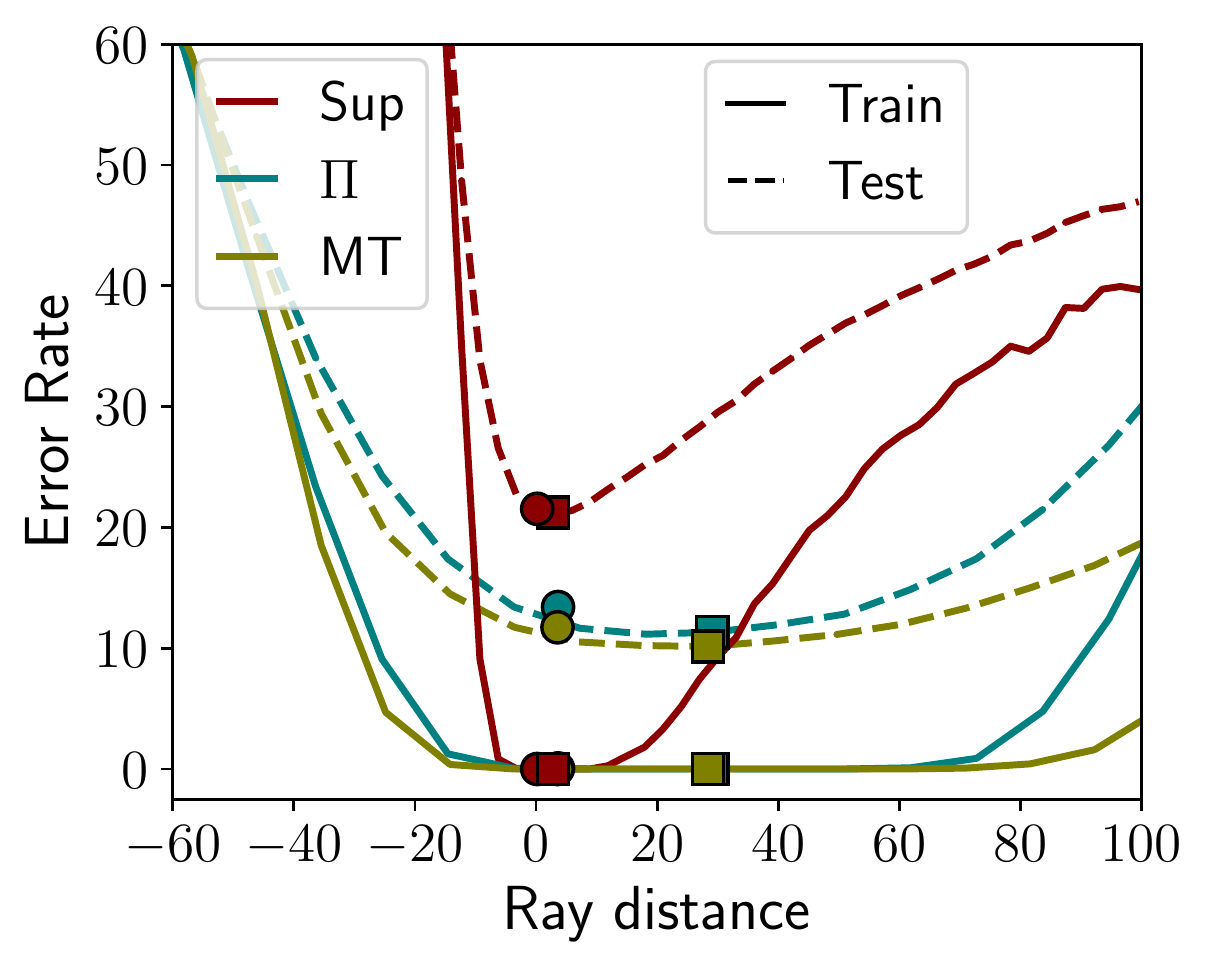}
\endminipage 
&
\hspace*{-0.4cm}
\minipage{0.32\textwidth}
       \centering
      \hspace*{-0.3cm}
      \includegraphics[width=0.96\textwidth,trim={0 0 0 0}, clip]{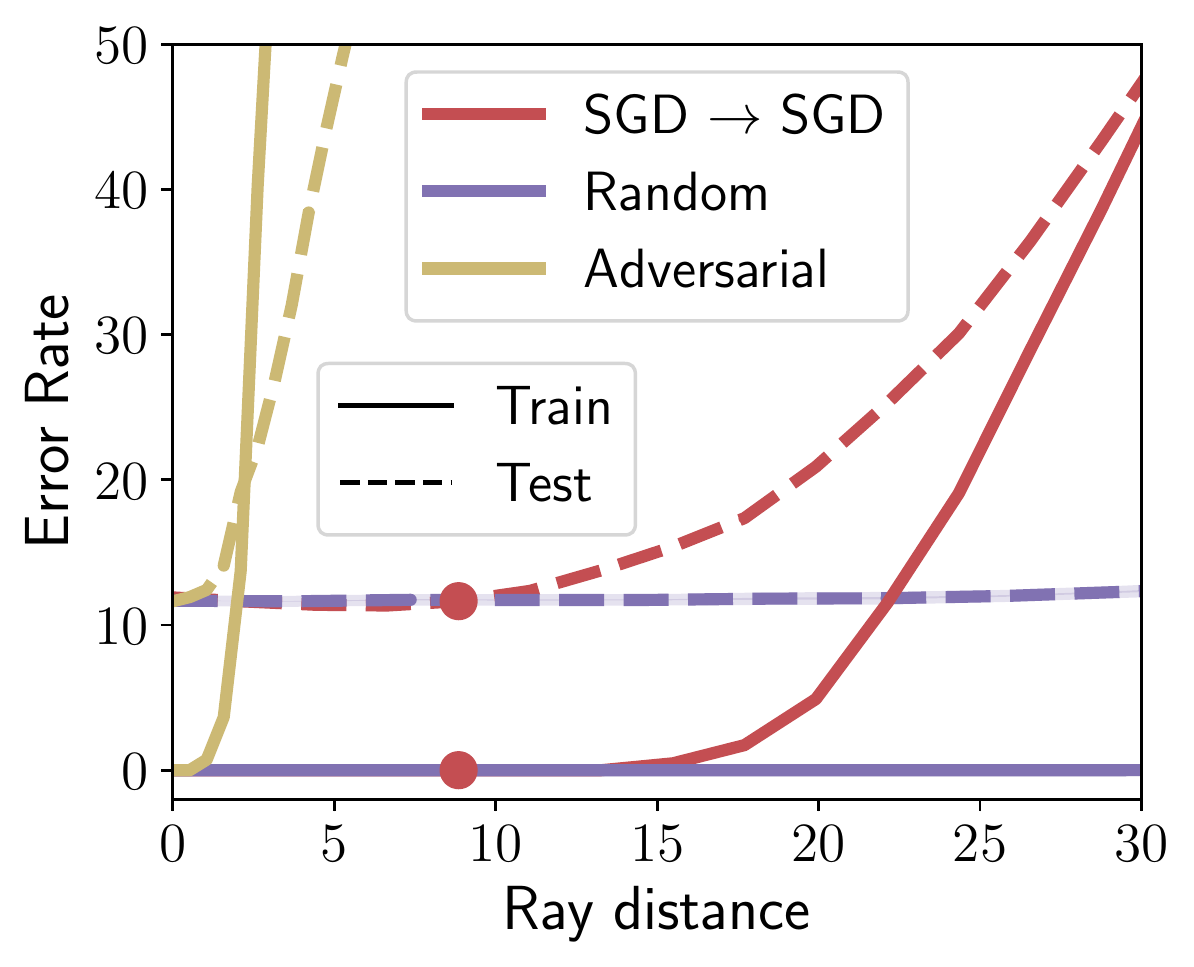}
\endminipage
\end{tabular}
  \caption{
    All plots are a obtained using the $13$-layer CNN on CIFAR-$10$ with 
    $4k$ labeled and $46k$ unlabeled data points unless specified otherwise.
    \tb{Left}: Test error as a function of distance along random rays for the \pi model
    with $0$, $4k$, $10k$, $20k$ or $46k$ unlabeled data points, and standard fully supervised training
    which uses only the cross entropy loss. All methods use $4k$ labeled examples.
    \tb{Middle}: Train and test errors along rays connecting SGD solutions 
    (showed with circles) to SWA solutions (showed with squares) for each respective model. 
    \tb{Right}: Comparison of train and test errors along rays connecting two 
    SGD solutions, random rays, and adversarial rays for the Mean Teacher model. 
  }
  \label{fig:sup_ray_plots}
\end{figure}

In this section we provide several additional plots visualizing the train and
test error along different types of rays in the weight space. The left panel of 
Figure \ref{fig:sup_ray_plots} shows how the behavior of test error
changes as we add more unlabeled data points for the \pi model. We observe that
the test accuracy improves monotonically, but also the solutions become narrower
along random rays. 

The middle panel of Figure \ref{fig:sup_ray_plots} visualizes the train and test error behavior
along the directions connecting the \fastswa solution (shown with squares) to 
one of the SGD iterates used to compute the average (shown with circles) for 
\pi, MT and supervised training.  
Similarly to \citet{swa} we observe that for all three methods \fastswa finds
a centered solution, while the SGD solution lies near the boundary of a wide
flat region. Agreeing with our results in section \ref{sec:trajectories} we 
observe that for \pi and Mean Teacher models the train and test error surfaces 
are much wider along the directions connecting the \fastswa and SGD solutions 
than for supervised training.

In the right panel of Figure \ref{fig:sup_ray_plots} we show the behavior of
train and test error surfaces along random rays, adversarial rays and 
directions connecting the SGD solutions from epochs $170$ and $180$ for the
Mean Teacher model (see section \ref{sec:trajectories}).

\begin{figure}[h]
  \centering
  \begin{tabular}[c]{c c c c}
  \hspace*{-0.1cm}
\hspace*{-0.5cm}
\minipage{0.25\textwidth}
       \centering
      \hspace*{-0.3cm}
       \includegraphics[height=3.5cm,trim={0 0 0 0}, clip]{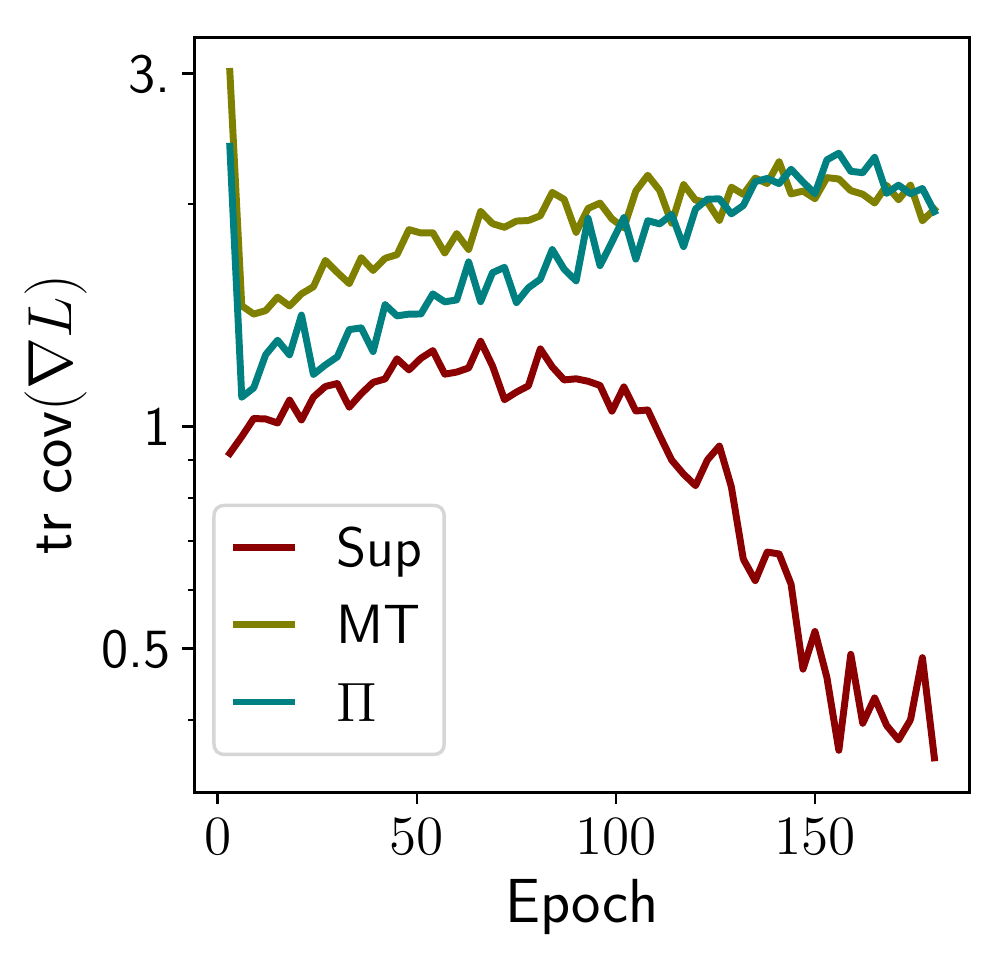}
      \vspace*{-1.cm}
\endminipage 
&
\minipage{0.3\textwidth}
       \centering
      \hspace*{-0.3cm}
      \includegraphics[height=4.cm,trim={0 0 0 0}, clip]{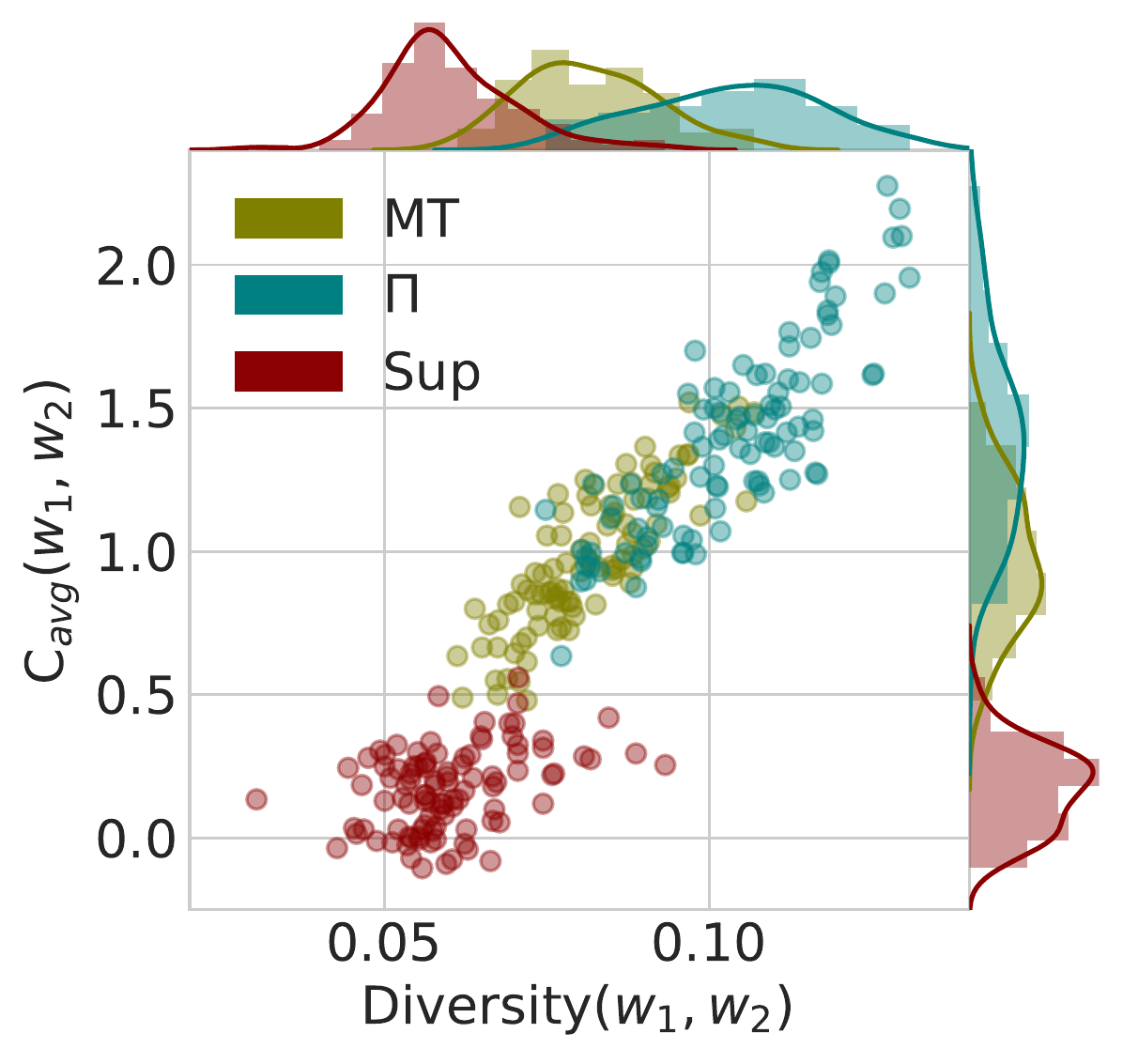}
\endminipage
&
\hspace*{-0.5cm}
\minipage{0.3\textwidth}
       \centering
      \hspace*{-0.3cm}
      \includegraphics[height=4.cm,trim={0 0 0 0}, clip]{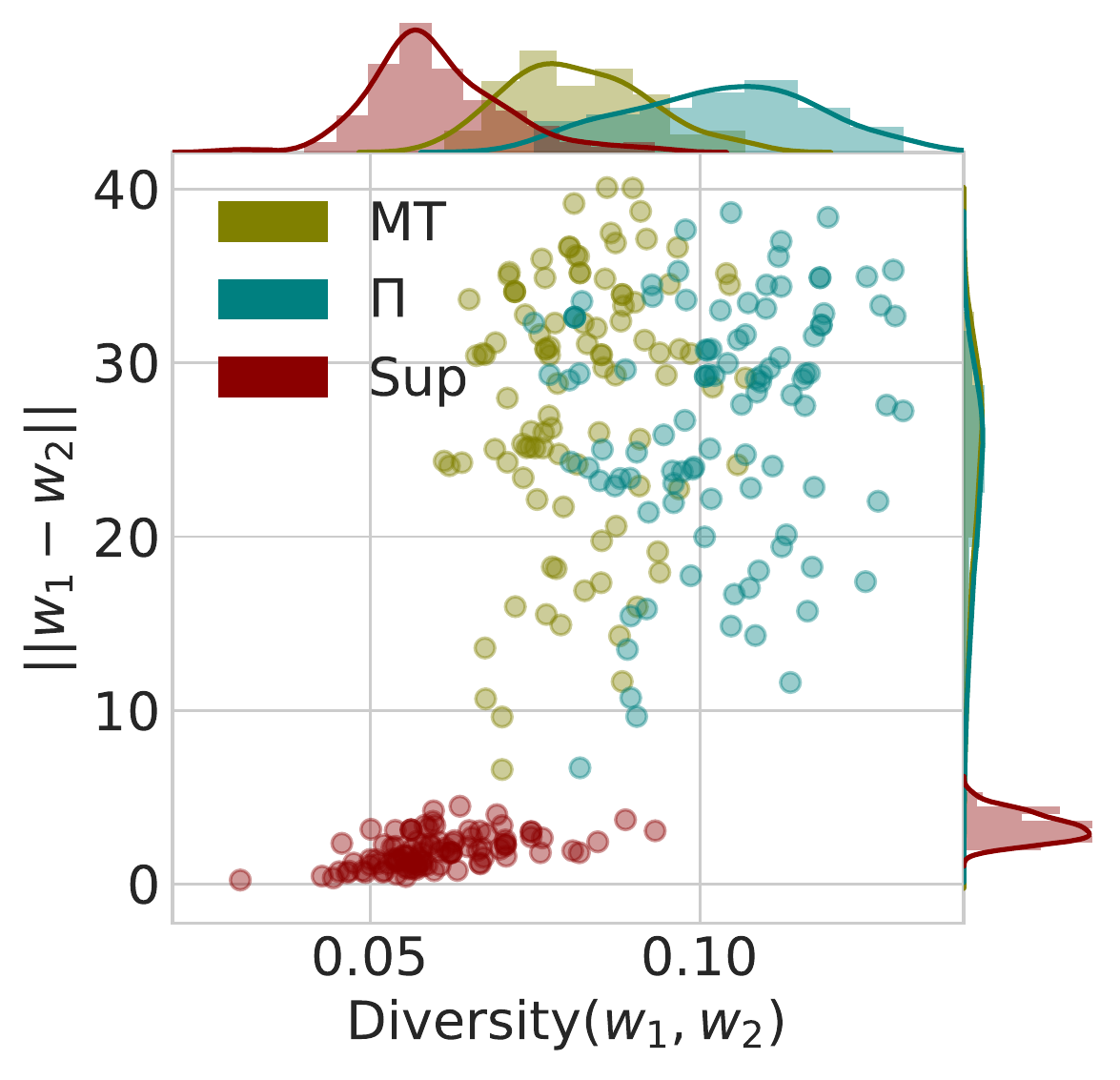}
\endminipage
&   \hspace*{-0.5cm}
\end{tabular}
  \caption{
    \tb{(Left)}: The evolution of the gradient covariance trace in the \pi, MT, 
    and supervised models during training.
    \tb{(Middle)}: Scatter plot of the decrease in error $C_{avg}$ for weight averaging versus diversity. 
    \tb{(Right)}: Scatter plot of the distance between pairs of weights
    versus diversity in their predictions. 
  }
  \label{fig:sup_hists}
\end{figure}

In the left panel of Figure \ref{fig:sup_hists} we show the evolution of the 
trace of the gradient of the covariance of the loss
\[
    \mbox{tr } \mbox{cov} (\nabla_w L(w)) =
    \E \|\nabla_w L(w) - \E \nabla_w L(w) \|^2
\]
for the \pi, MT and supevised training. We observe that the variance of the 
gradient is much larger for the \pi and Mean Teacher models compared to 
supervised training.

In the middle and right panels of figure \ref{fig:sup_hists} we provide
scatter plots of the improvement $C$ obtained from averaging weights against
diversity and diversity against distance. We observe that diversity is 
highly correlated with the improvement $C$ coming from weight averaging.
The correlation between distance and diversity is less prominent.

\subsection{Detailed Results} \label{sec:add_results}

In this section we report detailed results for the \pi and Mean
Teacher models and various baselines on CIFAR-10 and CIFAR-100 using 
the $13$-layer CNN and Shake-Shake.

The results using the 13-layer CNN are summarized in Tables 
\ref{table:cifar10_cnn13_additional} and \ref{table:cifar100} 
for CIFAR-10 and CIFAR-100 respectively. 
Tables \ref{table:cifar10_shakeshake} and \ref{table:cifar100_shakeshake}
summarize the results using Shake-Shake on CIFAR-10 and CIFAR-100. 
In the tables \pi EMA is the same method as \pi, where instead
of \swa we apply Exponential Moving Averaging (EMA) for the student
weights. We show that simply performing EMA for the student network
in the \pi model without using it as a teacher (as in MT)
typically results in a small improvement in the test error.

\begin{figure}[h!]
\centering
\begin{subfigure}[t]{0.19\linewidth}
	\includegraphics[trim={42 10 50 22},width=\linewidth,valign=t]{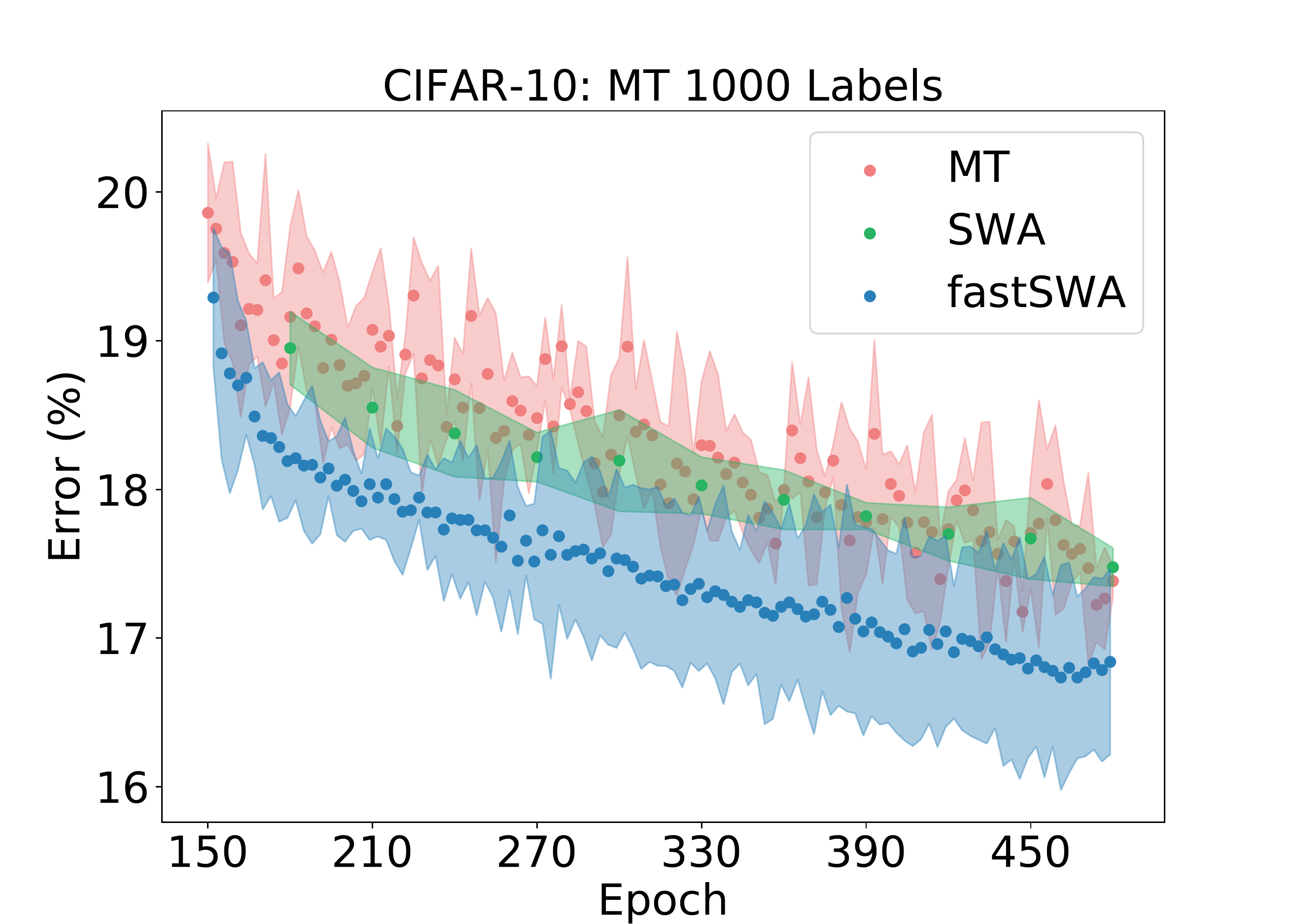}
\end{subfigure}
\hfill
\begin{subfigure}[t]{0.19\linewidth}
	\includegraphics[trim={42 10 50 22},width=\linewidth,valign=t]{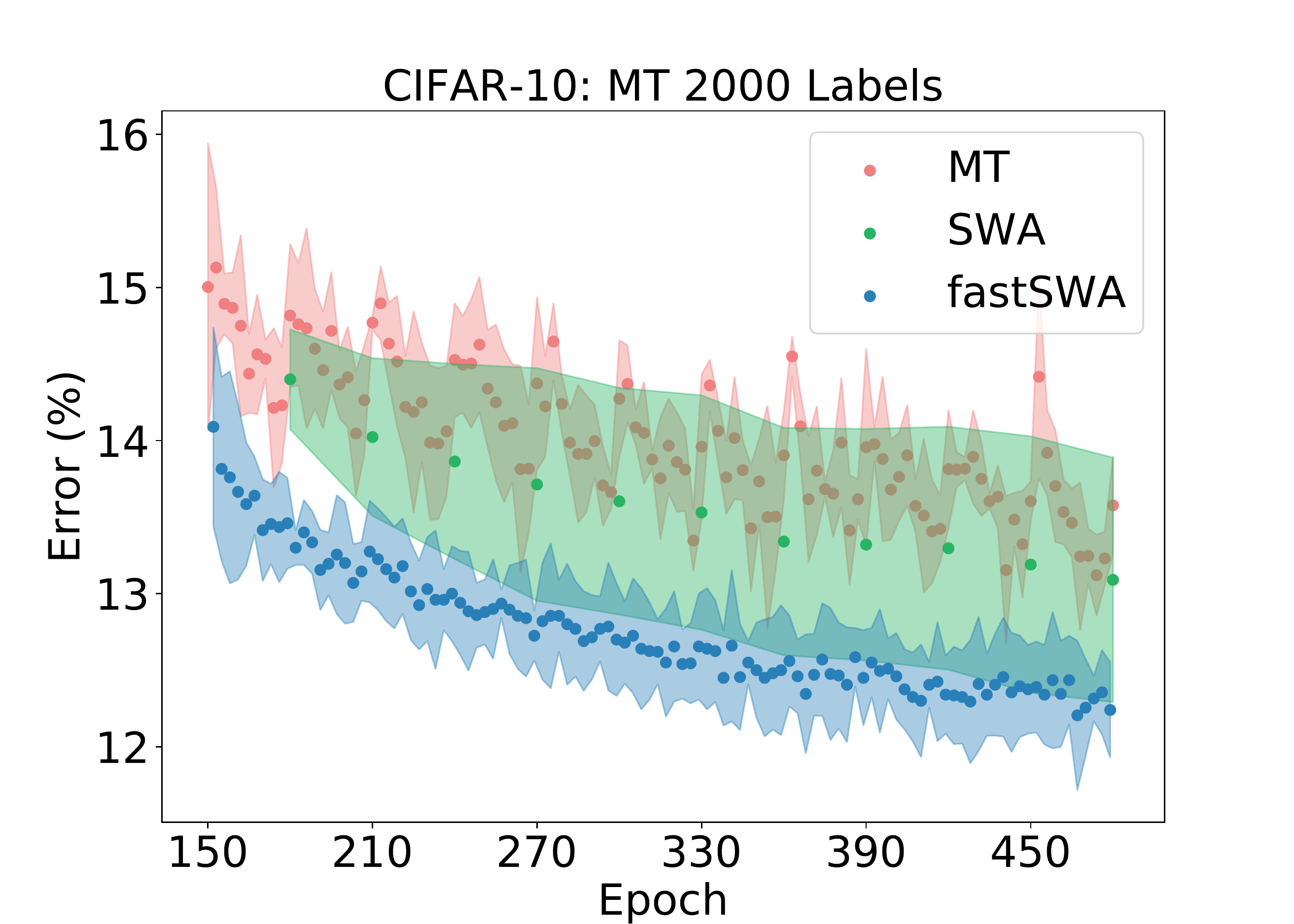}
\end{subfigure}
\hfill
\begin{subfigure}[t]{0.19\linewidth}
	\includegraphics[trim={42 10 50 22},width=\linewidth,valign=t]{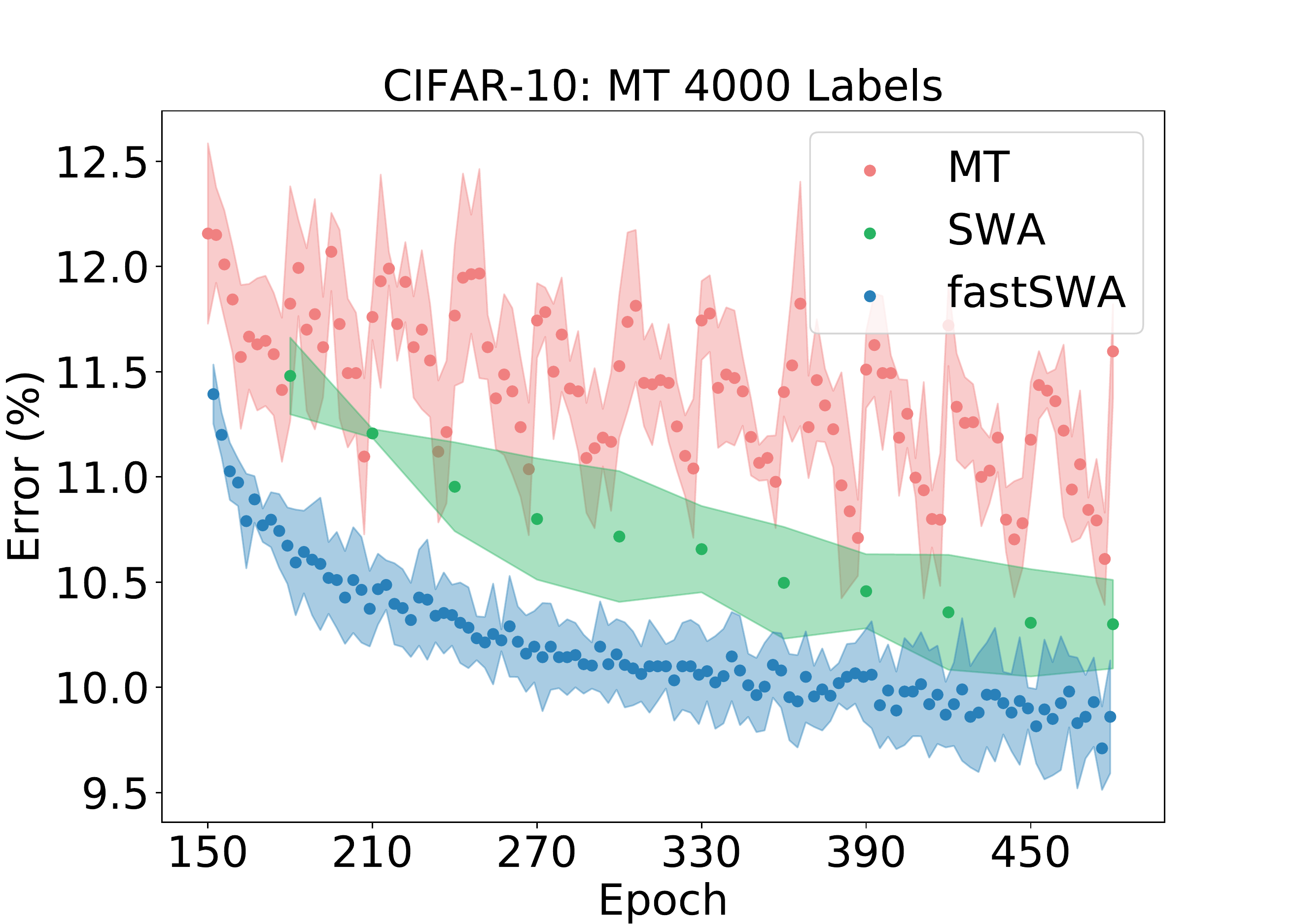}
\end{subfigure}
\hfill
\begin{subfigure}[t]{0.19\linewidth}
	\includegraphics[trim={42 10 50 22},width=\linewidth,valign=t]{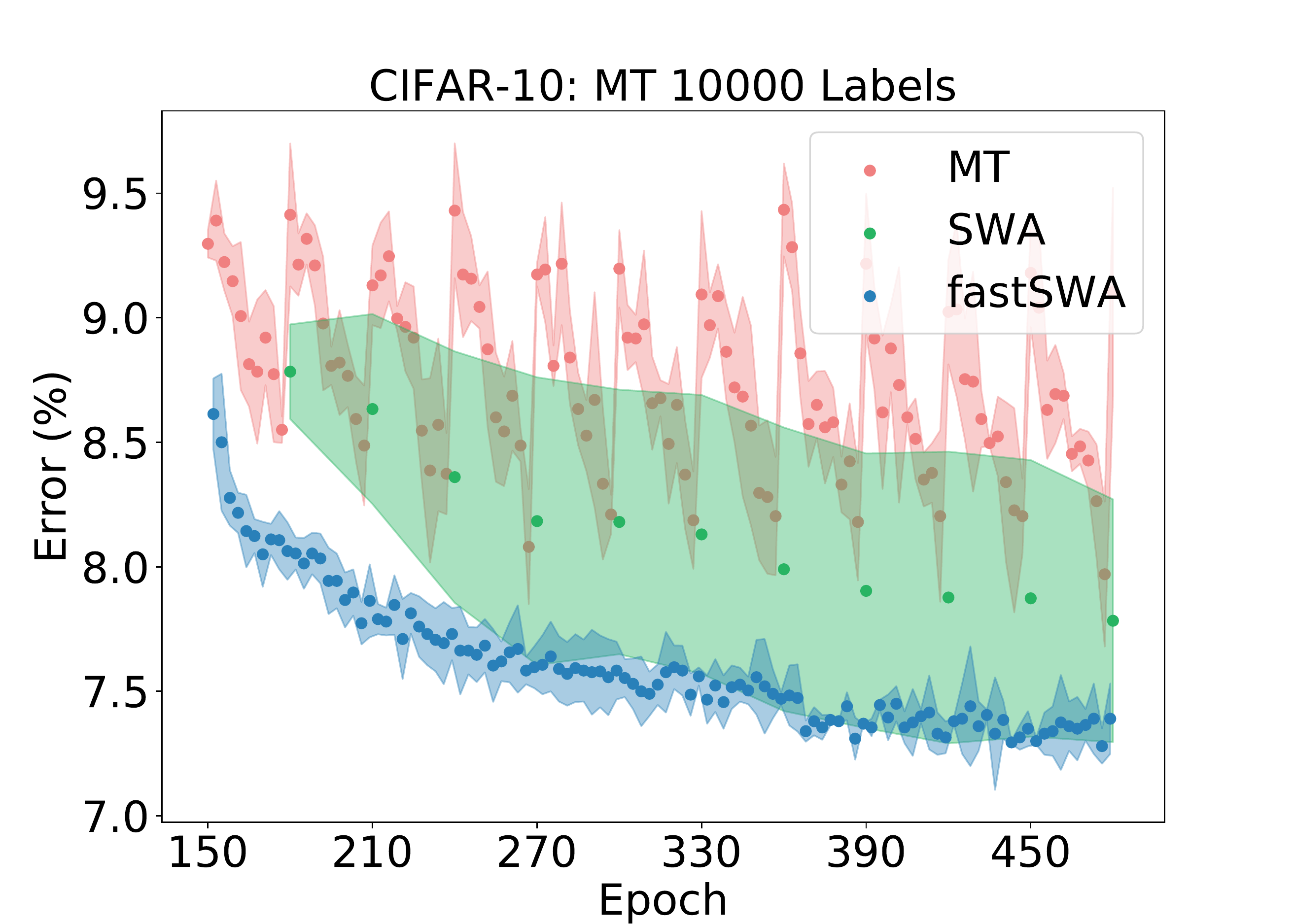}
\end{subfigure}
\begin{subfigure}[t]{0.19\linewidth}
	\includegraphics[trim={42 10 50 22},width=\linewidth,valign=t]{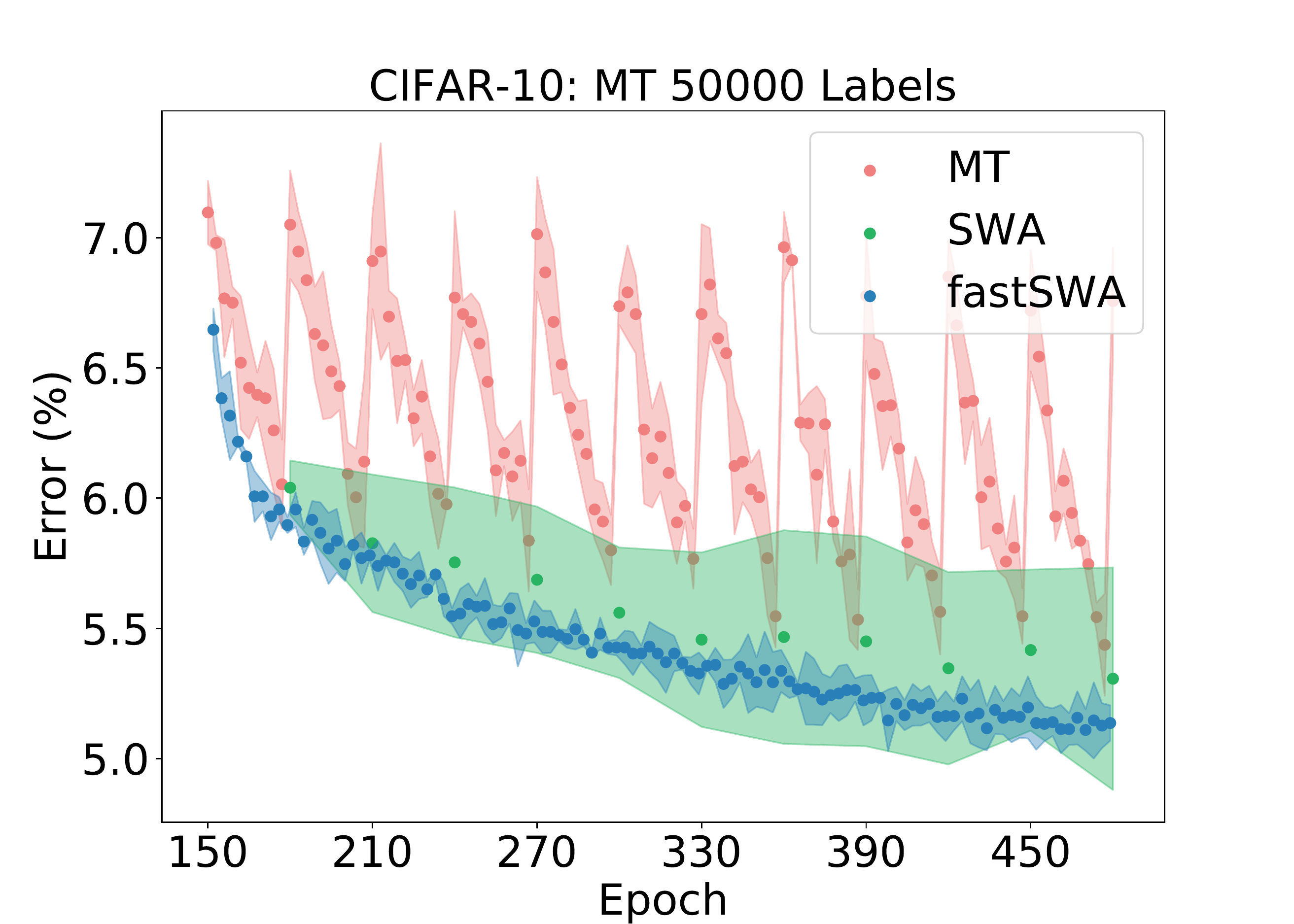}
\end{subfigure}

\begin{subfigure}[t]{0.19\linewidth}
	\includegraphics[trim={42 10 50 22},width=\linewidth,valign=t]{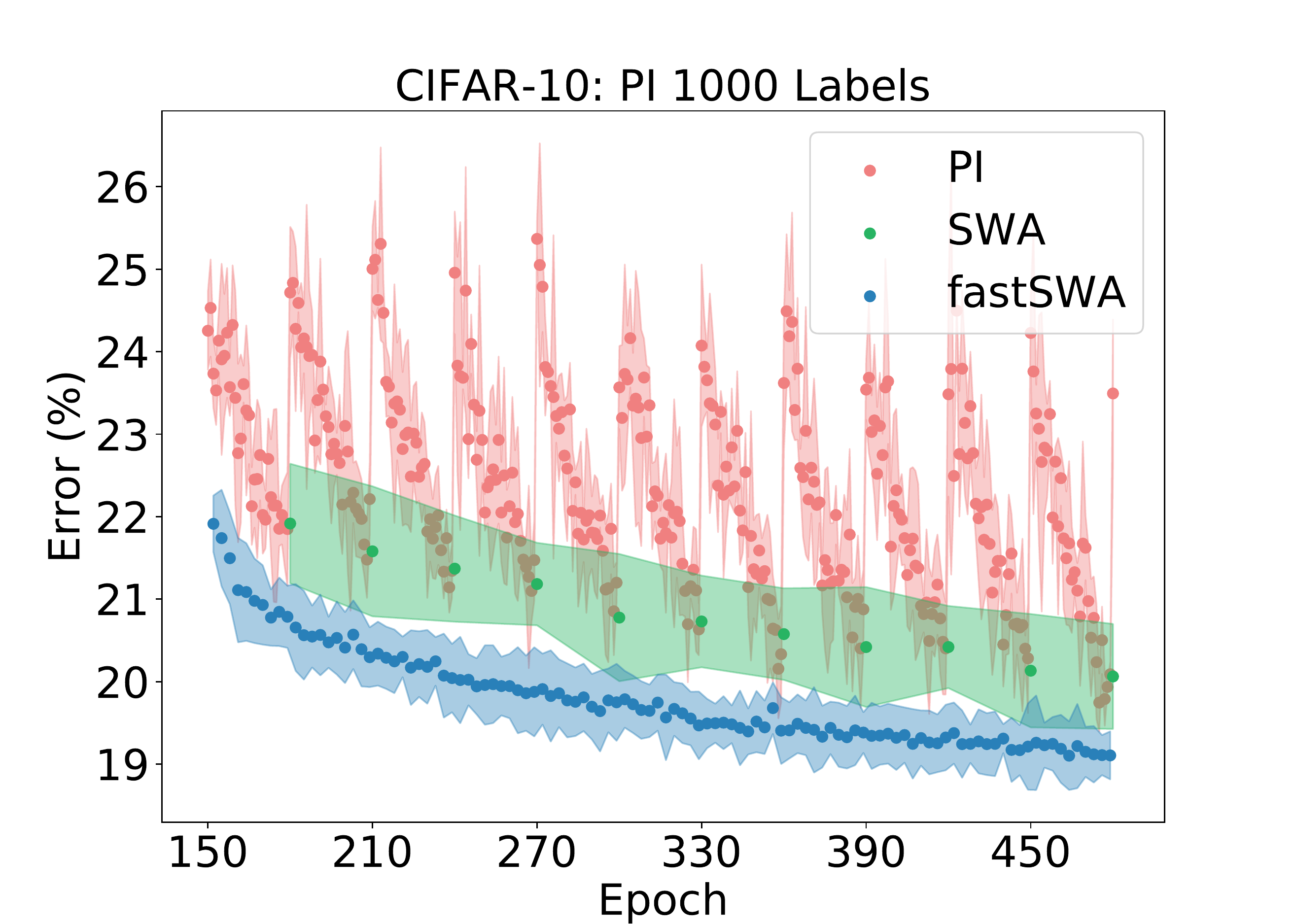}
\end{subfigure}
\hfill
\begin{subfigure}[t]{0.19\linewidth}
	\includegraphics[trim={42 10 50 22},width=\linewidth,valign=t]{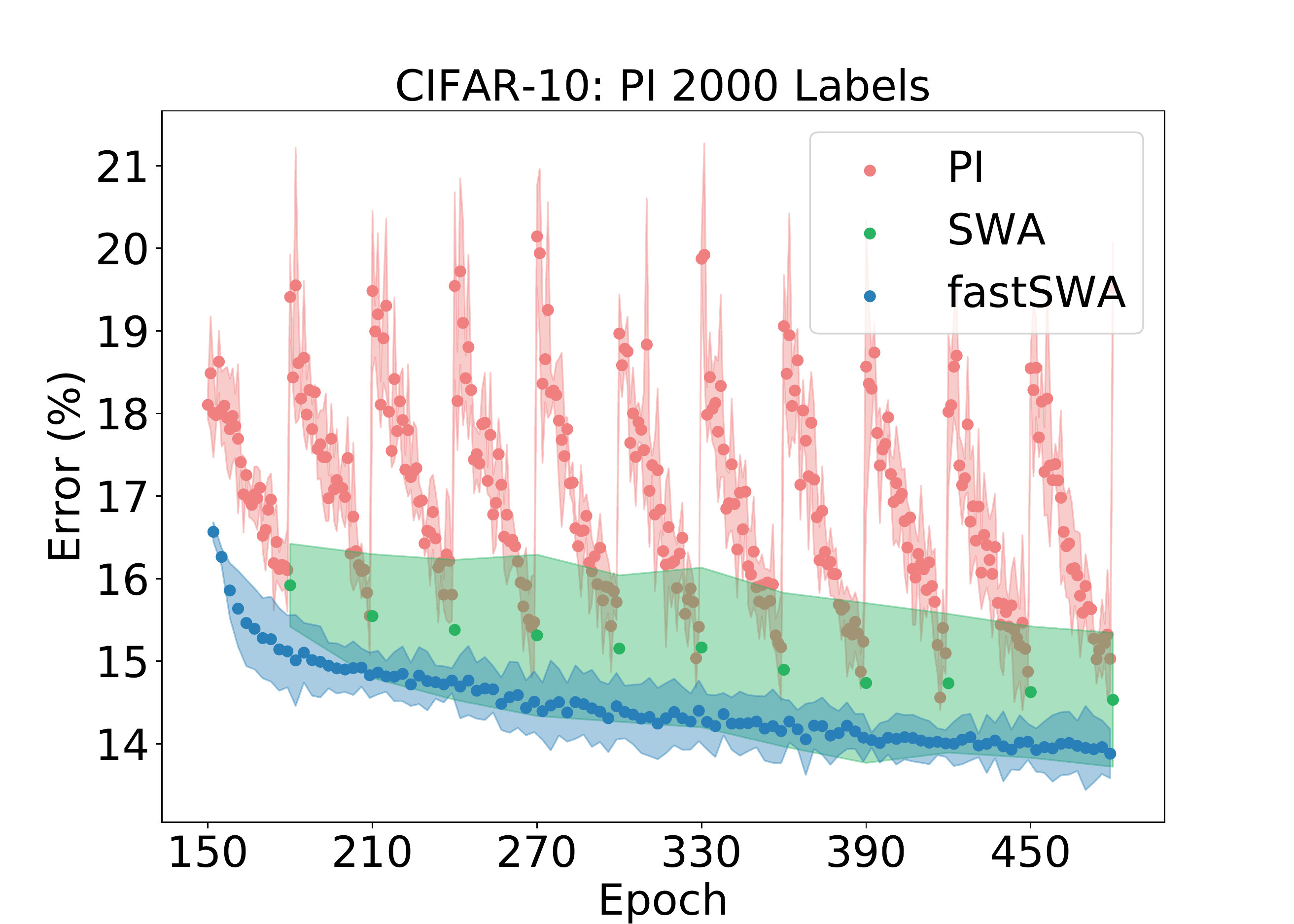}
\end{subfigure}
\hfill
\begin{subfigure}[t]{0.19\linewidth}
	\includegraphics[trim={42 10 50 22},width=\linewidth,valign=t]{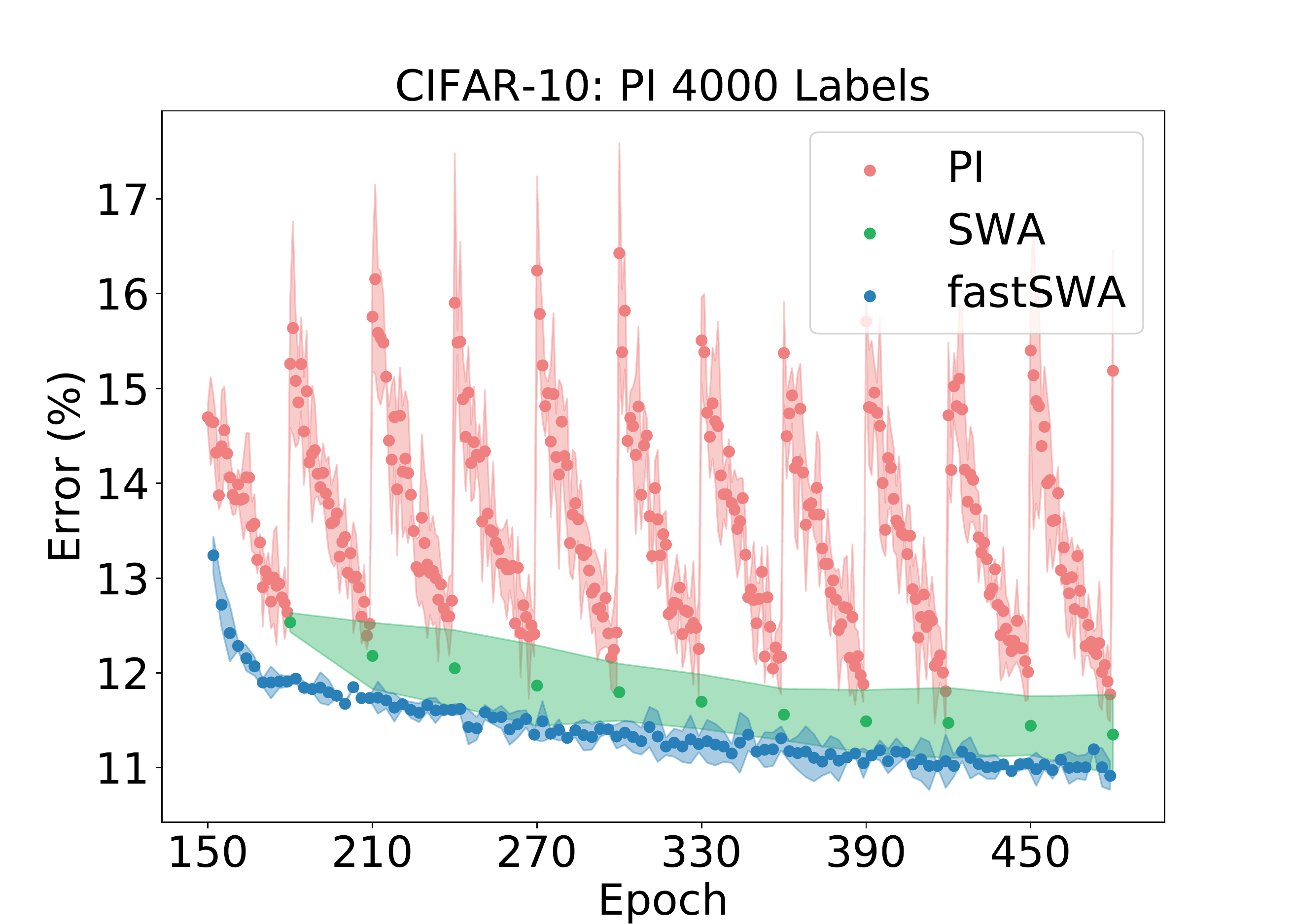}
\end{subfigure}
\hfill
\begin{subfigure}[t]{0.19\linewidth}
	\includegraphics[trim={42 10 50 22},width=\linewidth,valign=t]{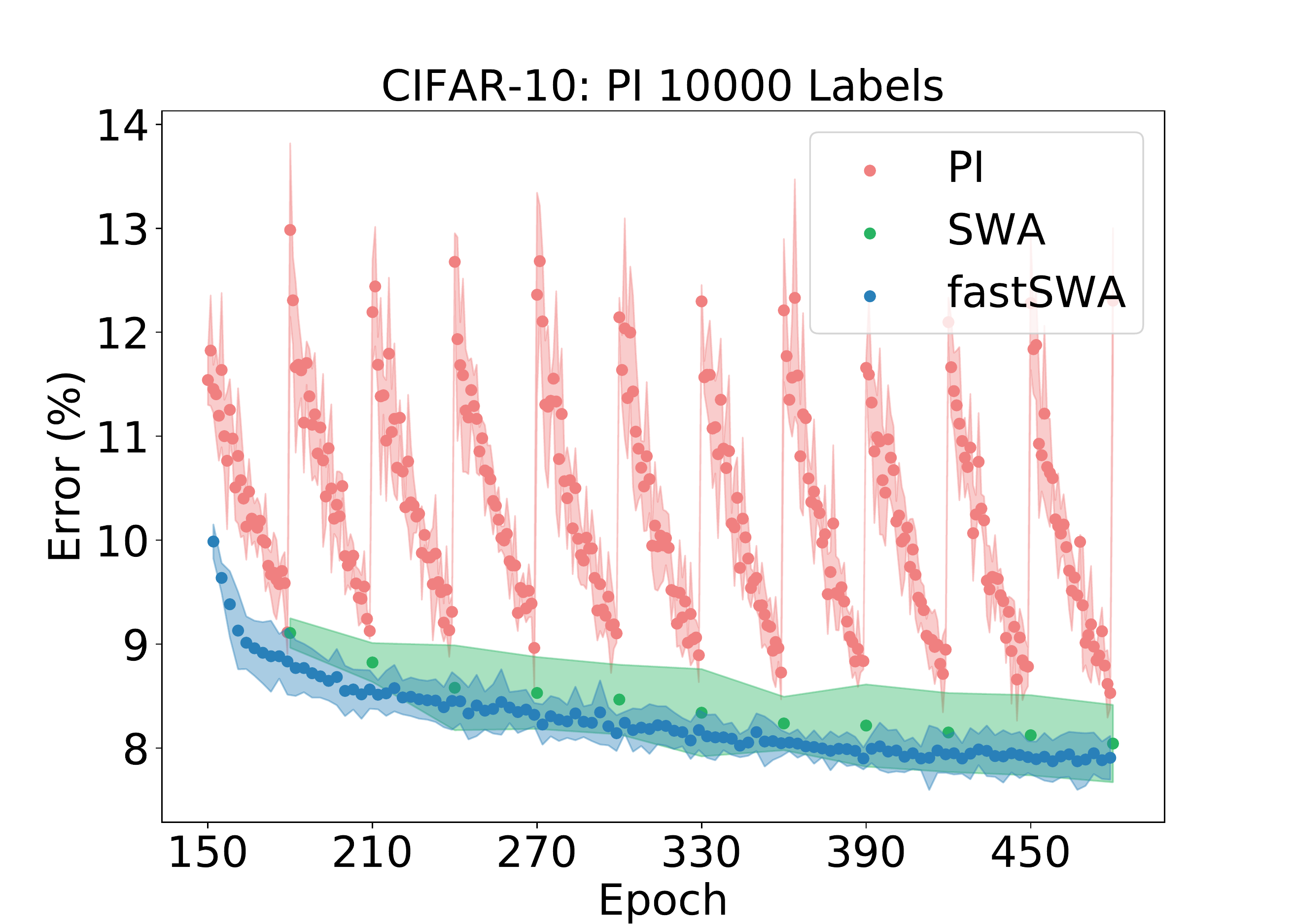}
\end{subfigure}
\hfill
\begin{subfigure}[t]{0.19\linewidth}
	\includegraphics[trim={42 10 50 22},width=\linewidth,valign=t]{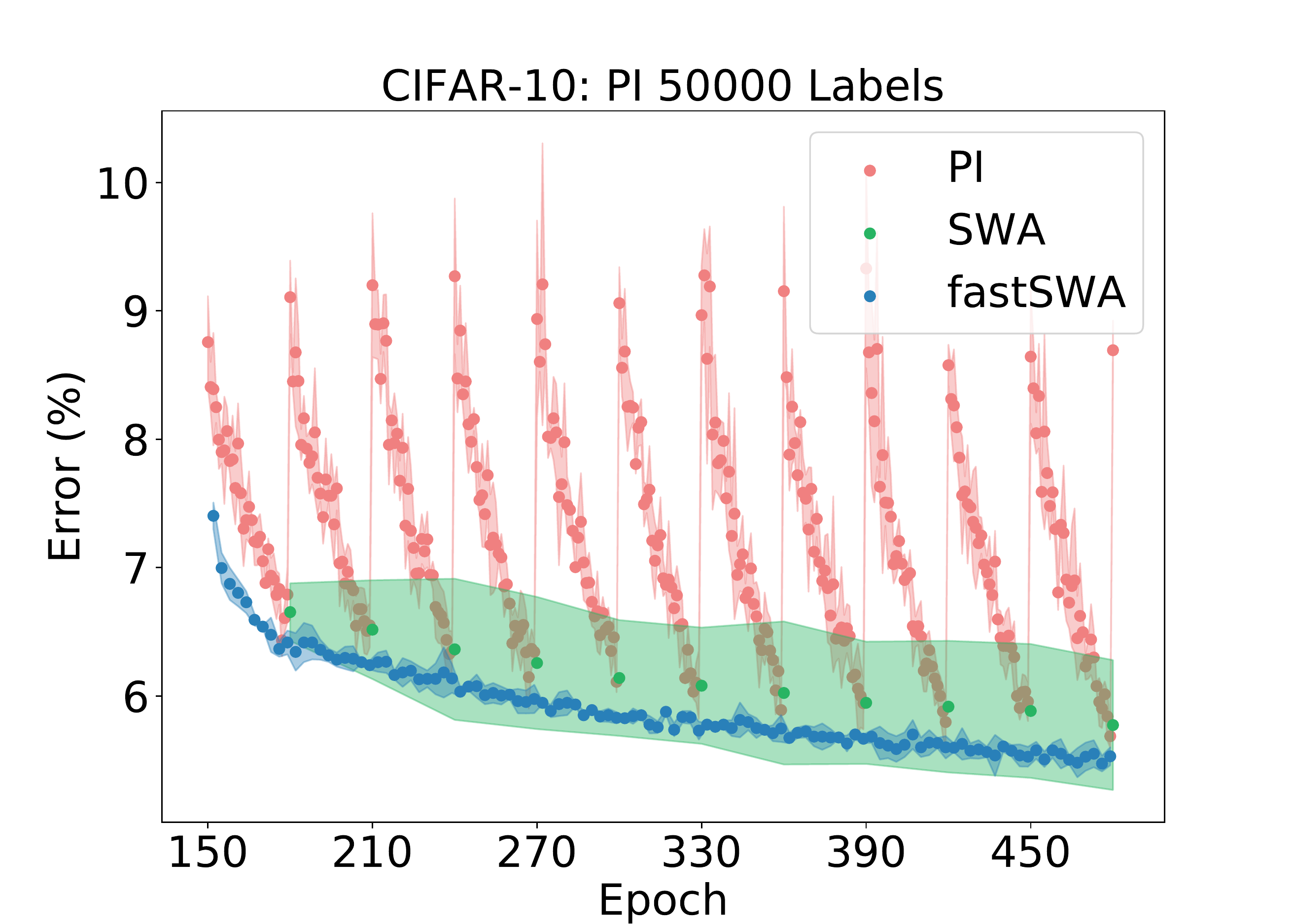}
\end{subfigure}
  \centering
  \caption{
  Test errors as a function of training epoch for baseline models,
  \swa and \fastswa on CIFAR-10 trained using $1k$, $2k$, $4k$, and $10k$ labels for 
  \textbf{(top)} the MT model
  \textbf{(bottom)} the \pi model.
  All models are trained using the 13-layer CNN. 
  }
  \label{fig:c10_trend_additional}
\end{figure}

\begin{figure}[h!]
\centering
\begin{subfigure}[t]{0.24\linewidth}
	\includegraphics[trim={42 10 50 22},width=\linewidth,valign=t]{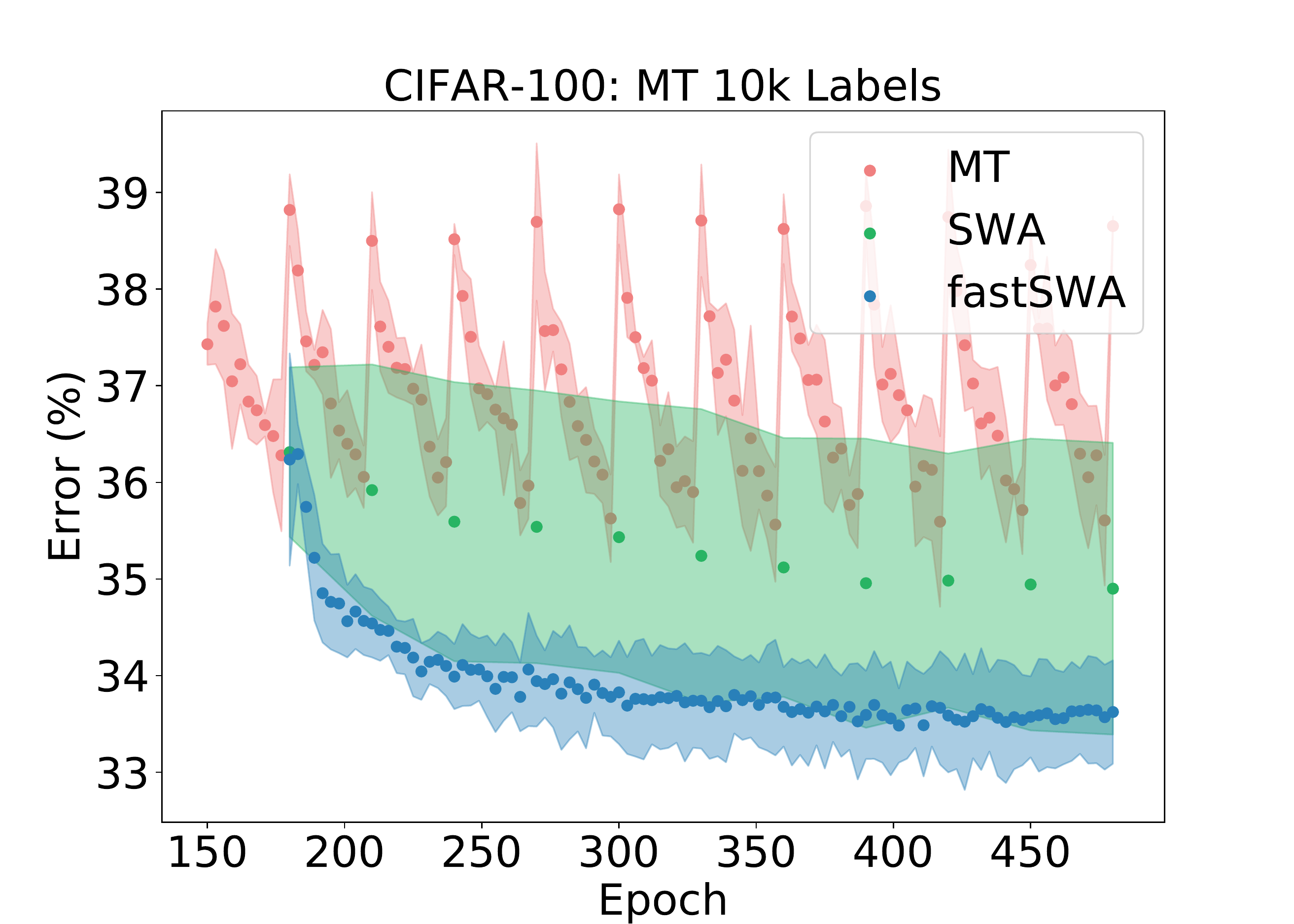}
\end{subfigure}
\hfill
\begin{subfigure}[t]{0.24\linewidth}
	\includegraphics[trim={42 10 50 22},width=\linewidth,valign=t]{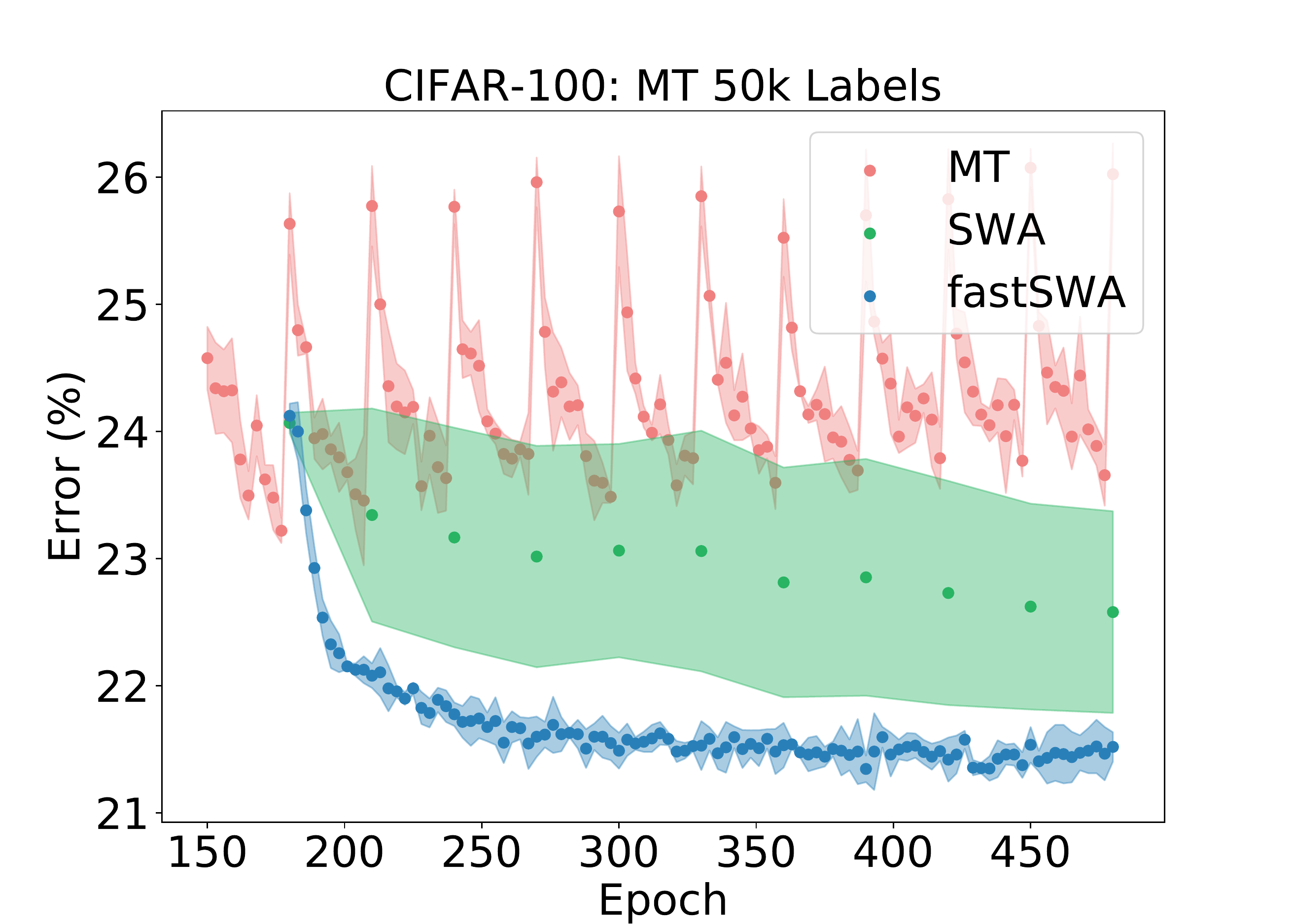}
\end{subfigure}
\hfill
\begin{subfigure}[t]{0.24\linewidth}
	\includegraphics[trim={42 10 50 22},width=\linewidth,valign=t]{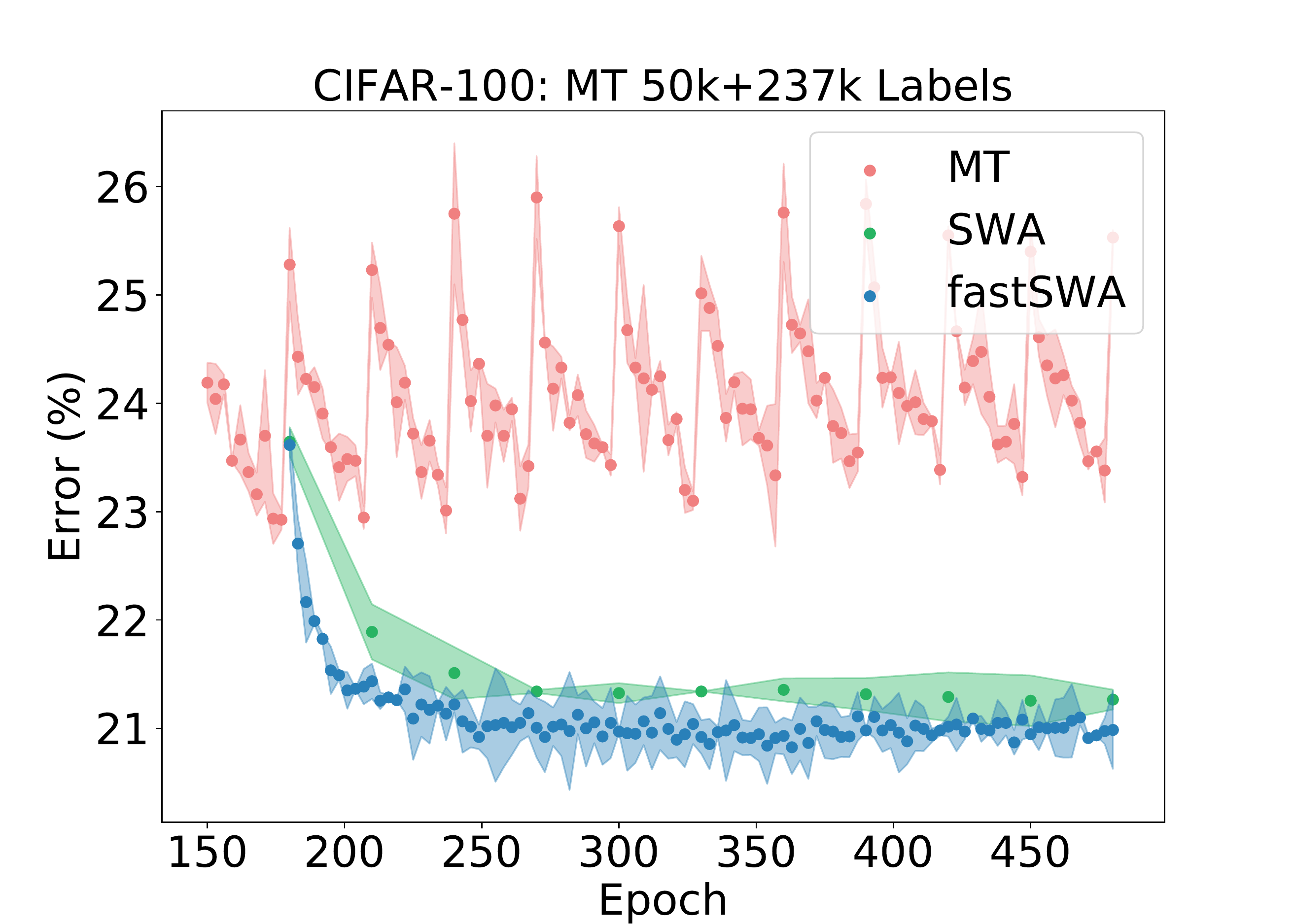}
\end{subfigure}
\hfill
\begin{subfigure}[t]{0.24\linewidth}
	\includegraphics[trim={42 10 50 22},width=\linewidth,valign=t]{figs/exps_results/epochs/c100_cnn_MT_50kExt500_wmodelTrue}
\end{subfigure}

\begin{subfigure}[t]{0.24\linewidth}
	\includegraphics[trim={42 10 50 22},width=\linewidth,valign=t]{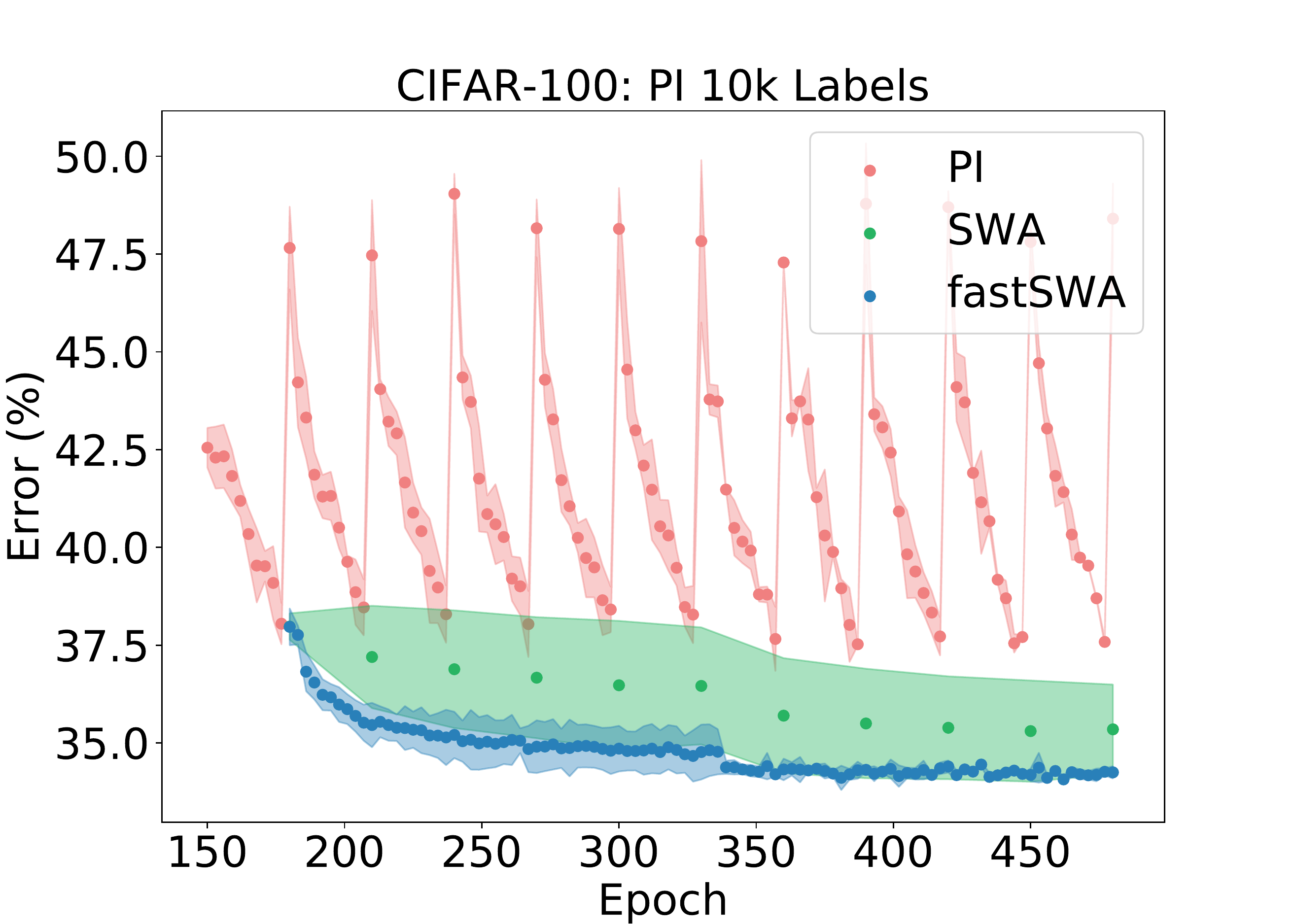}
\end{subfigure}
\hfill
\begin{subfigure}[t]{0.24\linewidth}
	\includegraphics[trim={42 10 50 22},width=\linewidth,valign=t]{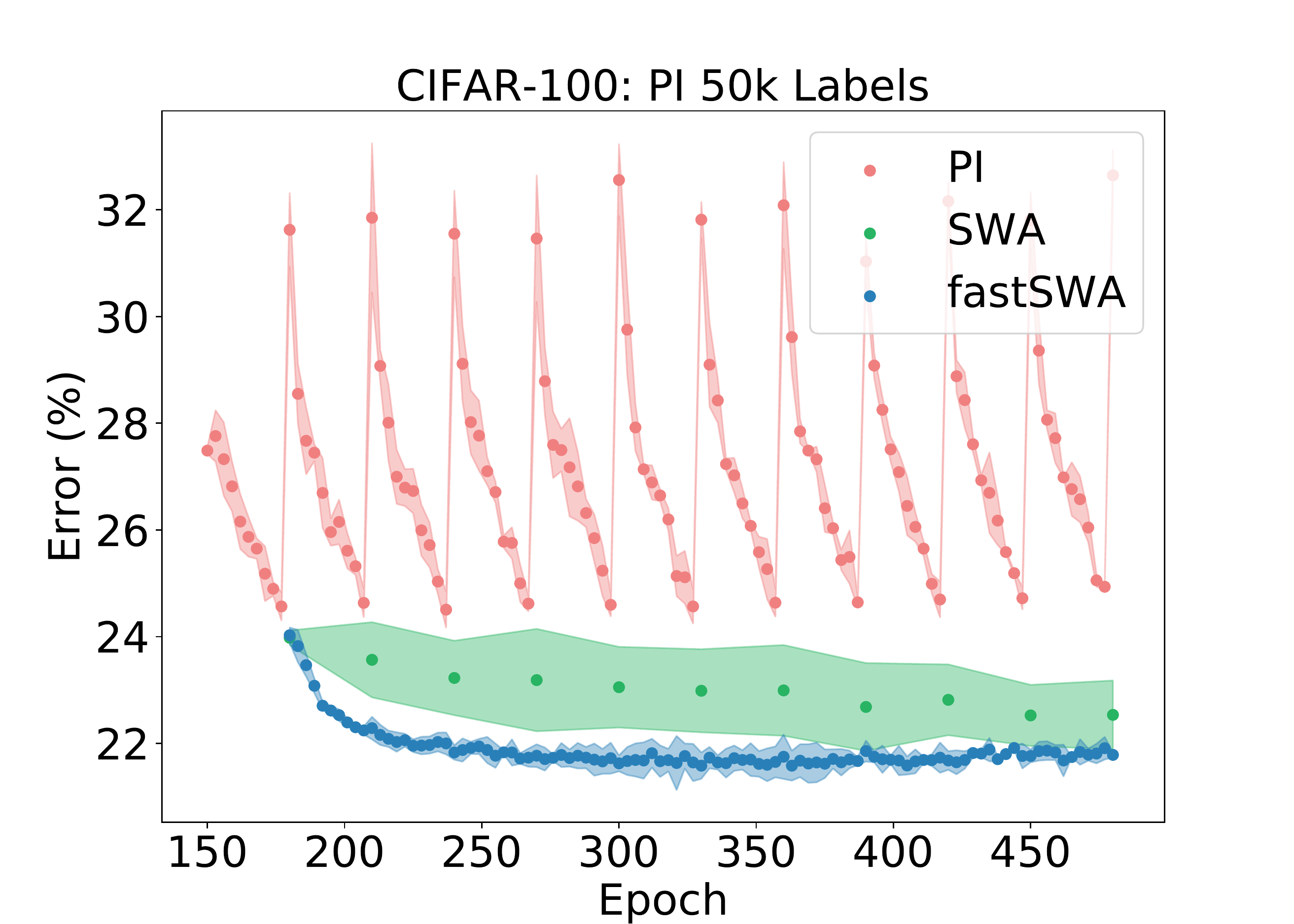}
\end{subfigure}
\hfill
\begin{subfigure}[t]{0.24\linewidth}
	\includegraphics[trim={42 10 50 22},width=\linewidth,valign=t]{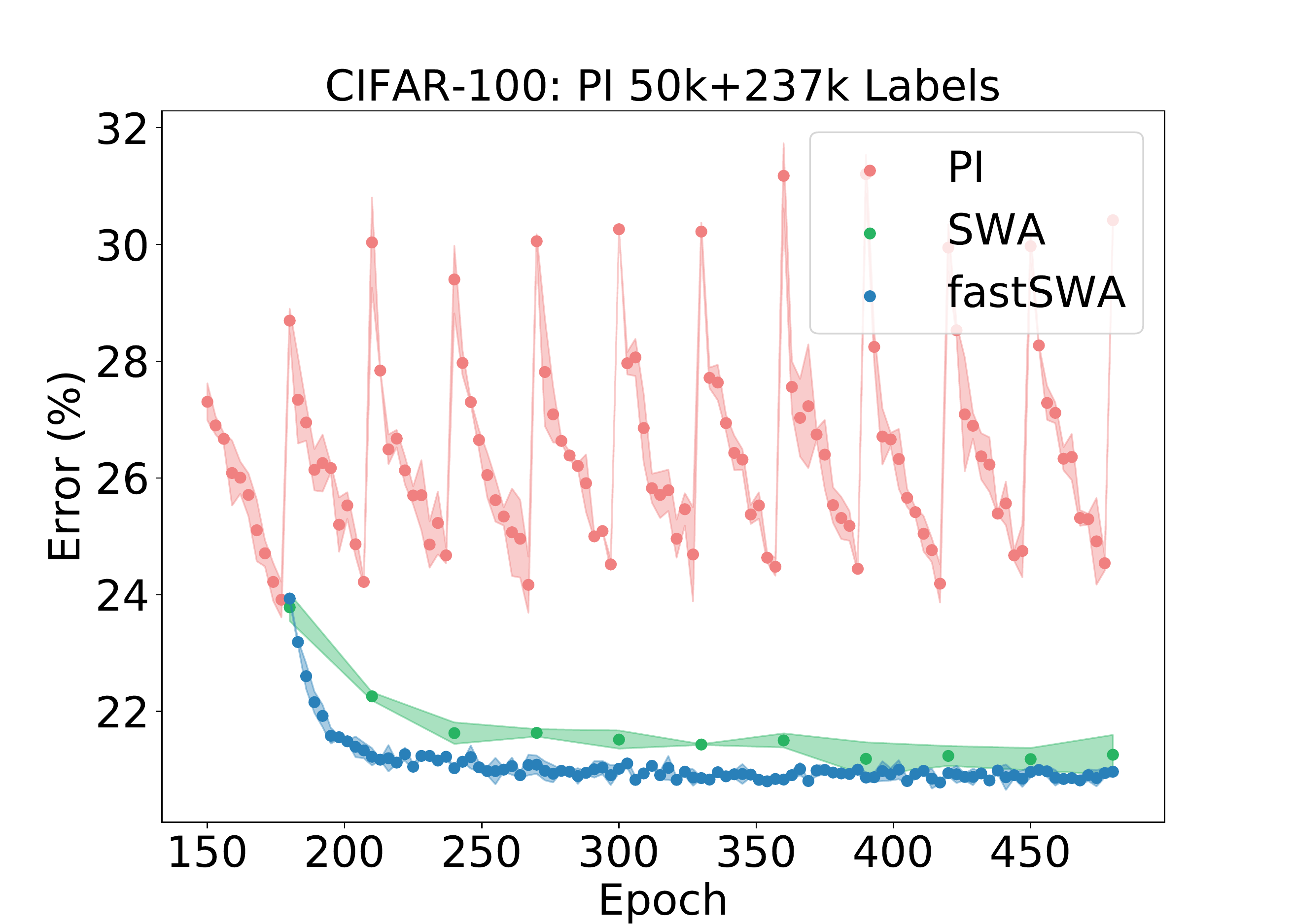}
\end{subfigure}
\hfill
\begin{subfigure}[t]{0.24\linewidth}
	\includegraphics[trim={42 10 50 22},width=\linewidth,valign=t]{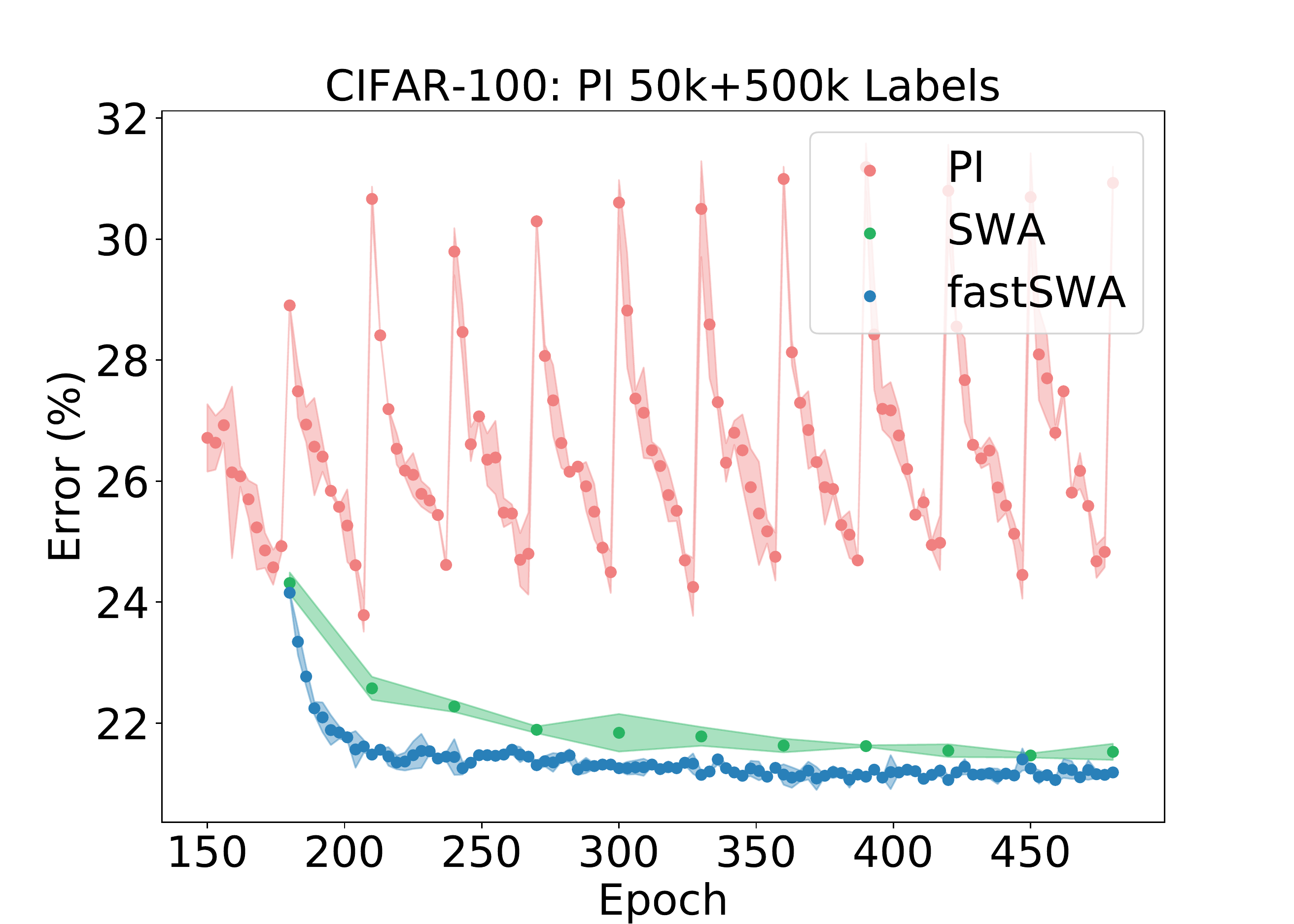}
\end{subfigure}
  \centering
  \caption{
  Test errors versus training epoch for baseline models,
  \swa and \fastswa on CIFAR-100 trained using $10k$, $50k$, $50k$+$500k$, and $50k$+$237k^*$ labels for 
  \textbf{(top)} the MT model
  \textbf{(bottom)} the \pi model.
  All models are trained using the 13-layer CNN. 
  }
  \label{fig:c100_trend_additional} 
\end{figure}

Figures \ref{fig:c10_trend_additional} and \ref{fig:c100_trend_additional}  show 
the performance 
of the \pi and Mean Teacher models as a function of the training epoch for 
CIFAR-10 and CIFAR-100 respectively for SWA and \fastswa.

\begin{table}[h!]
  \caption{CIFAR-10 semi-supervised errors on test set with a 13-layer CNN. 
  The epoch numbers are reported in parenthesis. 
  The previous results shown in the first section of the table are obtained 
  from \citet{mt}$^1$,  \citet{vad}$^2$, \citet{te}$^3$, \citet{vat2}$^4$. 
  }
  \label{table:cifar10_cnn13_additional}
  \centering
  \begin{tabular}{l l l l l l l l }
Number of labels	& 	1000  		& 	2000 		& 	4000  		& 10000 	& 	50000 	\\
\midrule
TE$^3$	
        &	-		       	& 	-		&	12.16 \tpm 0.31	&	& 5.60 \tpm 0.15	\\
Supervised-only$^1$
        & 46.43 \tpm 1.21	& 33.94 \tpm 0.73	&	20.66 \tpm 0.57	&	& 5.82 \tpm 0.15  	\\
\pi$^1$
        & 27.36 \tpm 1.20 	& 18.02 \tpm 0.60	&	13.20 \tpm 0.27	&	& 6.06 \tpm 0.15 	\\
MT$^1$
        & 21.55 \tpm 1.48     	& 15.73 \tpm 0.31	&	12.31 \tpm 0.28	&	& 5.94 \tpm 0.15 	\\
VAdD$^3$
        &     	& 	& 9.22 \tpm 0.10	&  & \bf{4.40 \tpm 0.12}		\\
VAT	+ EntMin$^4$
        &     	& 	& 10.55	&		\\
\midrule
     MT 			&   18.78 \tpm 0.31 &  14.43 \tpm 0.20 &  11.41 \tpm 0.27 &  8.74 \tpm 0.30 &  5.98 \tpm 0.21 \\
MT + \fastswa (180)	&   18.19 \tpm 0.38 &  13.46 \tpm 0.30 &  10.67 \tpm 0.18 &  8.06 \tpm 0.12 &  5.90 \tpm 0.03 \\
MT + \fastswa (240)	&   17.81 \tpm 0.37 &  13.00 \tpm 0.31 &  10.34 \tpm 0.14 &  7.73 \tpm 0.10 &  5.55 \tpm 0.03 \\
MT + \swa (240)	&   18.38 \tpm 0.29 &  13.86 \tpm 0.64 &  10.95 \tpm 0.21 &  8.36 \tpm 0.50 &  5.75 \tpm 0.29 \\
MT + \fastswa (480)	&   16.84 \tpm 0.62 &  12.24 \tpm 0.31 &   9.86 \tpm 0.27 &  7.39 \tpm 0.14 &  5.14 \tpm 0.07 \\
MT + \swa (480)	&   17.48 \tpm 0.13 &  13.09 \tpm 0.80 &  10.30 \tpm 0.21 &  7.78 \tpm 0.49 &  5.31 \tpm 0.43 \\
MT + \fastswa (1200)&   \bf{15.58  \tpm 0.12} 	&	\bf{11.02         \tpm 0.23}   	&	\bf{9.05     \tpm 0.21} 	& \bf{6.92     \tpm 0.07} 	& 	\bf{4.73    \tpm 0.18} \\
MT + \swa (1200)&  15.59      \tpm 0.77   	& 11.42        \tpm 0.33	& 9.38 \tpm 0.28		& 7.04      \tpm 0.11	&  5.11 \tpm 0.35 \\
\midrule 
\pi 				&  21.85 \tpm 0.69 &  16.10 \tpm 0.51 &  12.64 \tpm 0.11 &  9.11 \tpm 0.21 &  6.79 \tpm 0.22 \\
\pi EMA 			&  21.70 \tpm 0.57 &  15.83 \tpm 0.55 &  12.52 \tpm 0.16 &  9.06 \tpm 0.15 &  6.66 \tpm 0.20 \\
\pi + \fastswa (180) 	&  20.79 \tpm 0.38 &  15.12 \tpm 0.44 &  11.91 \tpm 0.06 &  8.83 \tpm 0.32 &  6.42 \tpm 0.09 \\
\pi + \fastswa (240) 	&  20.04 \tpm 0.41 &  14.77 \tpm 0.15 &  11.61 \tpm 0.06 &  8.45 \tpm 0.28 &  6.14 \tpm 0.11 \\
\pi + \swa (240) 	&  21.37 \tpm 0.64 &  15.38 \tpm 0.85 &  12.05 \tpm 0.40 &  8.58 \tpm 0.41 &  6.36 \tpm 0.55 \\
\pi + \fastswa (480) 	&  19.11 \tpm 0.29 &  13.88 \tpm 0.30 &  10.91 \tpm 0.15 &  7.91 \tpm 0.21 &  5.53 \tpm 0.07 \\
\pi + \swa (480) 	&  20.06 \tpm 0.64 &  14.53 \tpm 0.81 &  11.35 \tpm 0.42 &  8.04 \tpm 0.37 &  5.77 \tpm 0.51 \\
\pi + \fastswa (1200) 	&  \bf{17.23 \tpm 0.34}	& \bf{12.61 \tpm 0.18}	& 	\bf{10.07    \tpm 0.27} 	&  \bf{7.28      \tpm 0.23}   		& \bf{4.72    \tpm 0.04} \\
\pi + \swa (1200) 	&  17.70      \tpm 0.25  & 12.59  \tpm 0.29 & 	10.73 \tpm 0.39 & 7.13      \tpm 0.23 & 4.99 \tpm 0.41
\\
\midrule 
VAT		&			       	& 				&	11.99			&			\\
VAT + SWA		&			       	& 				&	11.16 &		\\
VAT+ EntMin		&			       	& 				& 11.26				&			\\
VAT + EntMin + SWA     		&			       	& 				&	\bf{10.97}	\\
\end{tabular}
\end{table}

\begin{table}[h!]
 \caption{CIFAR-100 semi-supervised errors on test set. 
 All models are trained on a 13-layer CNN. The epoch numbers are reported in parenthesis.   
  The previous results shown in the first section of the table are obtained 
  from \citep{te}$^3$. 
  }
  \label{table:cifar100} \label{table:table:c100_trend_full}
  \centering
  \begin{tabular}{l l l l l l l l }
Number of labels		& 	10k 		& 	50k 		& 	50k + 500k 		& 50k +  237k$^*$  	\\
\midrule
Supervised-only$^3$ 
        & 	44.56 \tpm 0.30		&	26.42 \tpm 0.17		&	 	&	  		\\
$\Pi$ model$^3$ 
        & 	39.19 \tpm 0.54		&	26.32 \tpm 0.04		&	25.79 \tpm 0.17  	&	25.43 \tpm 0.17   		\\
Temporal Ensembling$^3$ 
        & 	38.65 \tpm 0.51		&	26.30 \tpm 0.15		&	23.62 \tpm 0.17  	&	 23.79 \tpm 0.17  		\\
\midrule
MT (180)    			& 	35.96 \tpm 0.77 &  23.37 \tpm 0.16 &  23.18 \tpm 0.06 &  23.18 \tpm 0.24 \\
MT + \fastswa (180)	&	34.54 \tpm 0.48 &  21.93 \tpm 0.16 &  21.04 \tpm 0.16 &  21.09 \tpm 0.12 \\
MT  + \swa (240)	&	35.59 \tpm 1.45 &  23.17 \tpm 0.86 &  22.00 \tpm 0.23 &  21.59 \tpm 0.22 \\
MT  + \fastswa (240)	&	34.10 \tpm 0.31 &  21.84 \tpm 0.12 &  21.16 \tpm 0.21 &  21.07 \tpm 0.21 \\
MT  + \swa (1200)		&	34.90 \tpm 1.51 &  22.58 \tpm 0.79 &  21.47 \tpm 0.29 &  21.27 \tpm 0.09 \\
MT  + \fastswa (1200) 	&	\tb{33.62 \tpm 0.54} &  \tb{21.52 \tpm 0.12} &  \tb{21.04 \tpm 0.04} &  \tb{20.98 \tpm 0.36} \\
\midrule
\pi (180)     			&      38.13 \tpm 0.52 &  24.13 \tpm 0.20 &  24.26 \tpm 0.15 &  24.10 \tpm 0.07 \\
\pi  + \fastswa (180) 	& 	35.59 \tpm 0.62 &  22.08 \tpm 0.21 &  21.40 \tpm 0.19 &  21.28 \tpm 0.20 \\ 
\pi + \swa (240)		&	36.89 \tpm 1.51 &  23.23 \tpm 0.70 &  22.17 \tpm 0.19 &  21.65 \tpm 0.13 \\
\pi  + \fastswa (240)	&	35.14 \tpm 0.71 &  22.00 \tpm 0.21 &  21.29 \tpm 0.27 &  21.22 \tpm 0.04 \\
\pi + \swa 	(1200)	&  	35.35 \tpm 1.15 &  22.53 \tpm 0.64 &  21.53 \tpm 0.13 &  21.26 \tpm 0.34 \\
\pi  + \fastswa (1200)	&      \tb{34.25 \tpm 0.16} &  \tb{21.78 \tpm 0.05} &  \tb{21.19 \tpm 0.05} &  \tb{20.97 \tpm 0.08} \\
  \end{tabular}
\end{table}

\begin{table}[H]
  \caption{CIFAR-10 semi-supervised errors on test set. All models use Shake-Shake Regularization \citep{shake} + ResNet-26 \citep{resnet}. 
  }
  \label{table:cifar10_shakeshake}
  \centering
  \begin{tabular}{l l  l l l l l l }
Number of labels			& 	1000  		& 	2000 		& 	4000  	& 10000 		& 	50000 	\\
\midrule
\multicolumn{5}{l}{Short Schedule $(\ell=180$) } \smallskip \\
MT$^\dagger$ \citep{mt}		&			&		&	6.28 &			&			&
\\ \midrule 
MT  (180)   			&	10.2  &   8.0  &   7.1  &    5.8  &    3.9  \\
MT + \swa (240)	&   9.7  &   7.7  &   6.2  &    4.9  &    3.4  \\
MT + \fastswa (240)    	 &   9.6  &   7.4  &   6.2  &    4.9  &    3.2  \\
MT + \swa   (1200)  		&	7.6	&	6.4	&  5.8	& 4.6		&	\tbf{3.1} \\
MT + \fastswa	(1200)	& 	\tbf{7.5}	& 	\tbf{6.3}	&	5.8		& 4.5			& 	\tbf{3.1} 		\\
\midrule 
\pi (180)    			& 	12.3  &   9.1  &   7.5  &    6.4  &    3.8  \\
\pi + \swa (240)	& 	11.0  &   8.3  &   6.7  &    5.5  &    3.3  \\
\pi + \fastswa (240) 	&	11.2  &   8.2  &   6.7  &    5.5  &    3.3  \\
\pi + \swa	(1200)		&	8.2 		& 	6.7		&	5.7		&	4.2		& 	\tbf{3.1}\\
\pi + \fastswa (1200)   		& 	8.0		&      	6.5		&	\tbf{5.5}	& \tbf{4.0} 		& 	\tbf{3.1} 		\\
\midrule
\multicolumn{5}{l}{Long Schedule ($\ell=1500$) } \smallskip \\
Supervised-only \citep{shake}
				&			&			&			&			& 	\tb{2.86} 	\\
MT (1500)    				&	7.5		& 	6.5		&	6.0		& 	5.0	 	&	3.5 \\
MT + \fastswa (1700)	&	6.4		&	5.8	&	5.2	& 	3.8 		& 3.4		\\
MT + \swa (1700)		&	6.9		&	5.9 & 	5.5	& 	4.2		&	3.2	\\
MT + \fastswa 	(3500)    	&	\tbf{6.6} 	& 	\tbf{5.7}	&   	5.1		& 	3.9	& 3.1 \\
MT + \swa (3500)	    	&	6.7	 	& 	5.8		&	5.2		& 	3.9	 	&	3.1 \\ 
\midrule
\pi (1500)   			& 	8.5		&      	7.0		&	6.3		& 	5.0	 	&	3.4 \\
\pi + \fastswa (1700)		& 	7.5		&	6.2		&	5.2		&	4.0		&	3.1	\\
\pi + \swa (1700)     	& 	7.8		&	6.4		&	5.6		&	4.4		&	3.2	\\
\pi + \fastswa (3500)		&	7.4		&	6.0		&	\tb{5.0}		&	\tb{3.8}	&	3.0	\\
\pi + \swa 	(3500)	   	& 	7.9		&	6.2 		&	5.1		& 	4.0	 	&	3.0 \\	
\end{tabular}
\end{table}

\clearpage

\begin{table}[H]
  \caption{CIFAR-100 semi-supervised errors on test set. Our models use Shake-Shake Regularization \citep{shake} + ResNet-26 \citep{resnet}. 
  }
  \label{table:cifar100_shakeshake}
  \centering
  \begin{tabular}{l c c c c c l l l l l }
Number of labels			& 	10k 		& 	50k 		& 	50k + 500k 		& 50k +   237k$^*$ 	\\ 
\midrule
TE (CNN) \citep{te}  & 	38.65 \tpm 0.51		&	26.30 \tpm 0.15		&	23.62 \tpm 0.17  	&	 23.79 \tpm 0.17  	\\ 
\midrule
\multicolumn{5}{l}{Short Schedule $(\ell=180$) } \smallskip \\
MT (180)   		& 29.4     	&  19.5     	&  21.9     		&  19.0    			\\ 
MT + \fastswa(180)	& 28.9     	&  19.3     	&  19.7     		&  18.3     			\\
MT + \swa (240)	& 28.4      	&  18.8    	&  19.9     		&  17.9     			\\
MT + \fastswa (240)  & 28.1     	&  18.8     	&  19.5     		&  17.9     			\\
MT + \swa	 (300)	& 28.1     	&  18.5     	&  \tbf{18.9}     	&  \tbf{17.5}     		\\
MT + \fastswa (300) 	&  \tbf{28.0}     & \tbf{18.4}     &  19.3     &  17.7     		\\
\end{tabular}
\end{table}

\subsection{Effect of Learning Rate Schedules} \label{sec:lr_schedule}
\begin{figure}[h!]
  \centering
\minipage{0.5\textwidth}
       \centering
       \centering
       \includegraphics[clip, trim={0 0 0 0},width=1.0\linewidth]{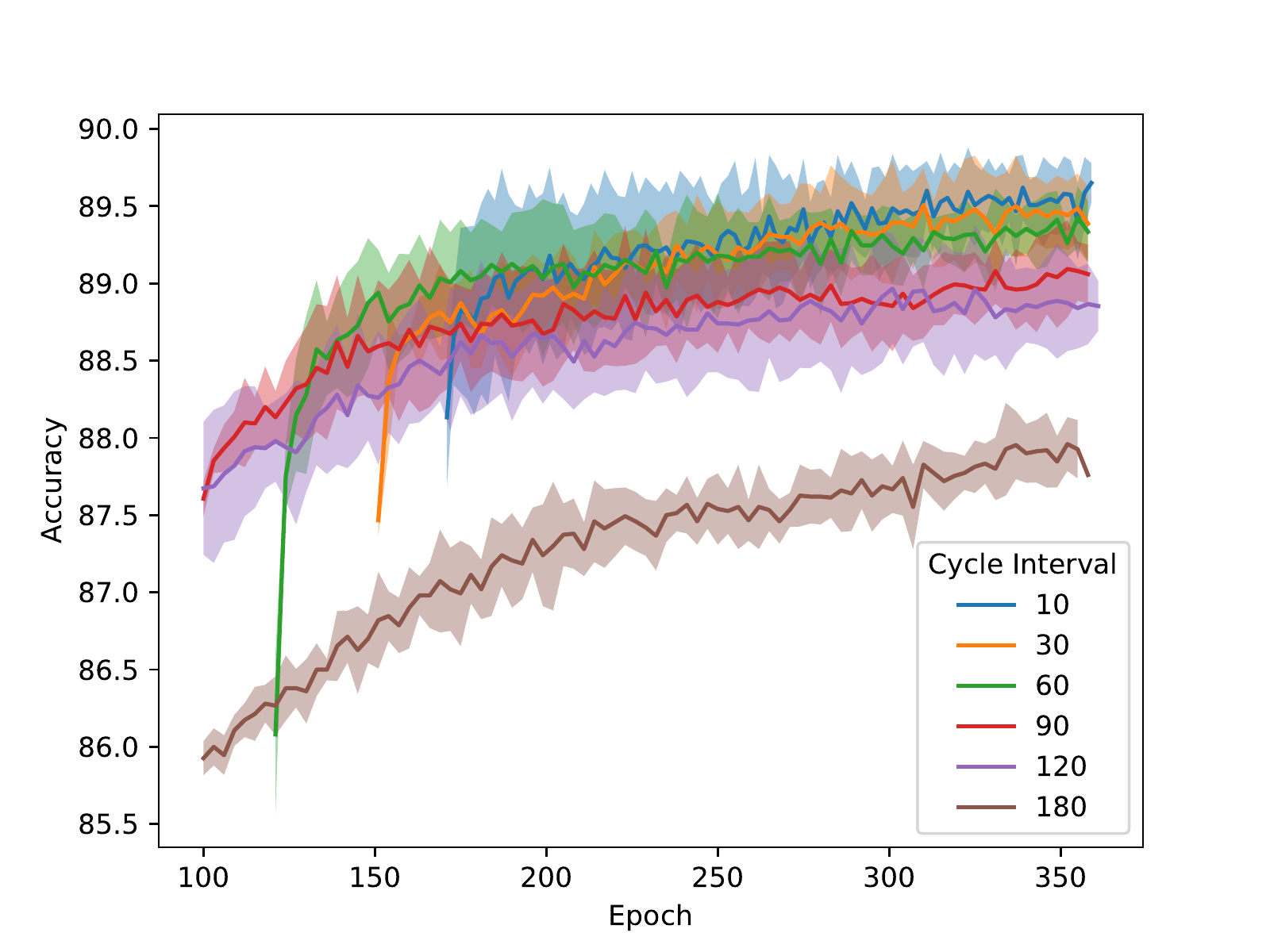}
   \vspace{-1\baselineskip}
  \subcaption{\label{fig:cycle_lr}}
  \endminipage\hfill
  \minipage{0.5\textwidth}
\centering
    \includegraphics[clip, trim={0 0 0 0}, width=1.0\linewidth]{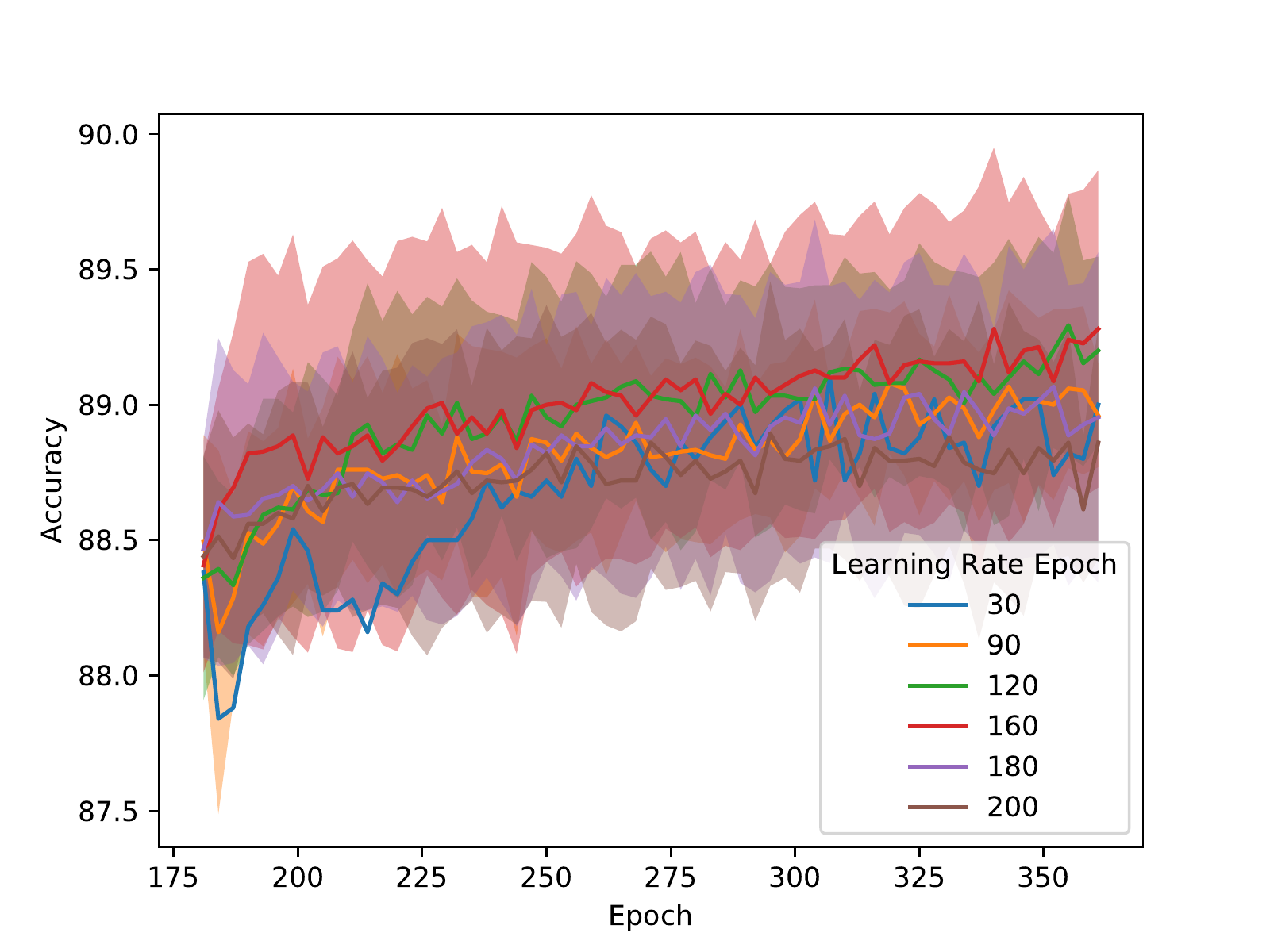}
    \vspace{-1\baselineskip}
 \subcaption{\label{fig:const_lr}}
\endminipage\hfill
  \caption{
  The plots are generated using the MT model with CNN trained on CIFAR-10. We randomly select
  $5k$ of the $50k$ train images as a validation set. The remaining $45k$
  images are splitted into $4k$ labeled and $41k$ unlabeled data points.
  \tbf{(a)} Validation accuracy as a function of training epoch for
  different cycle lengths $c$ 
  \tbf{(b)} \fastswa with constant learning rate. The ``learning rate epoch'' 
  corresponds to the epoch in the unmodified cosine annealing schedule (Figure \ref{fig:swa_model}, left)
  at which the learning rate is evaluated. We use this fixed learning rate 
  for all epochs $i \ge \ell$.
  }
  \label{fig:cycle_length}
\end{figure}

The only hyperparameter for the \fastswa setting is the cycle length $c$.  
We demonstrate in Figure \ref{fig:cycle_lr} that \fastswa's performance is not 
sensitive to $c$ over a wide range of $c$ values. 
We also demonstrate the performance for constant learning schedule. \fastswa with
cyclical learning rates generally converges faster due to higher variety in the
collected weights. 

\subsection{EMA versus SWA as a Teacher}
\label{sec:swa_as_teacher}

\begin{figure}[h!]
\centering
\hfill
\vspace{-0.4\baselineskip}
\minipage{0.48\textwidth}
       \centering
       \includegraphics[clip, trim={0 0 0 0},width=0.8\linewidth]{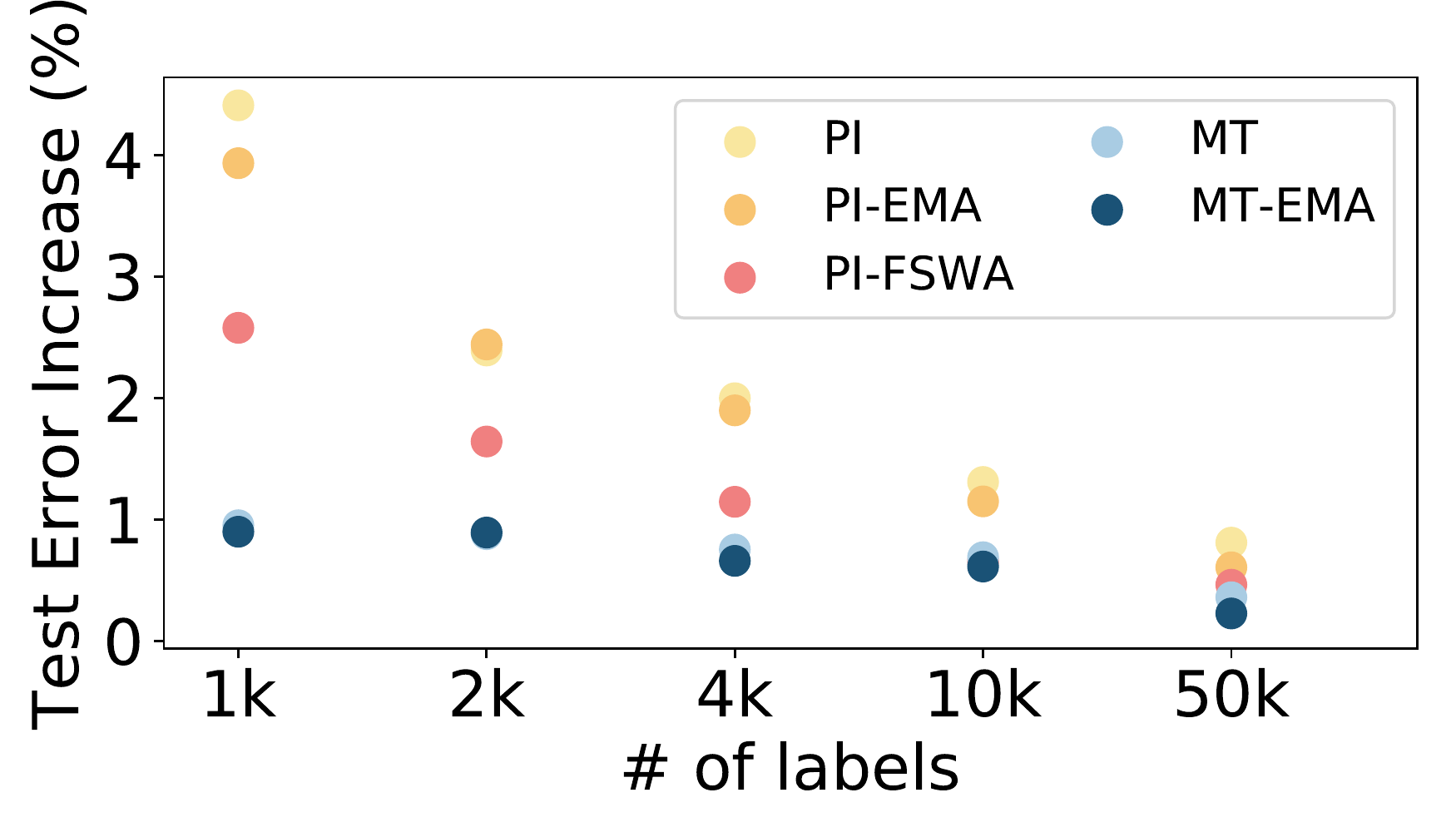}
\endminipage \hfill
\minipage{0.48\textwidth}
       \centering
       \includegraphics[clip, trim={0 0 0 0},width=0.6\linewidth]{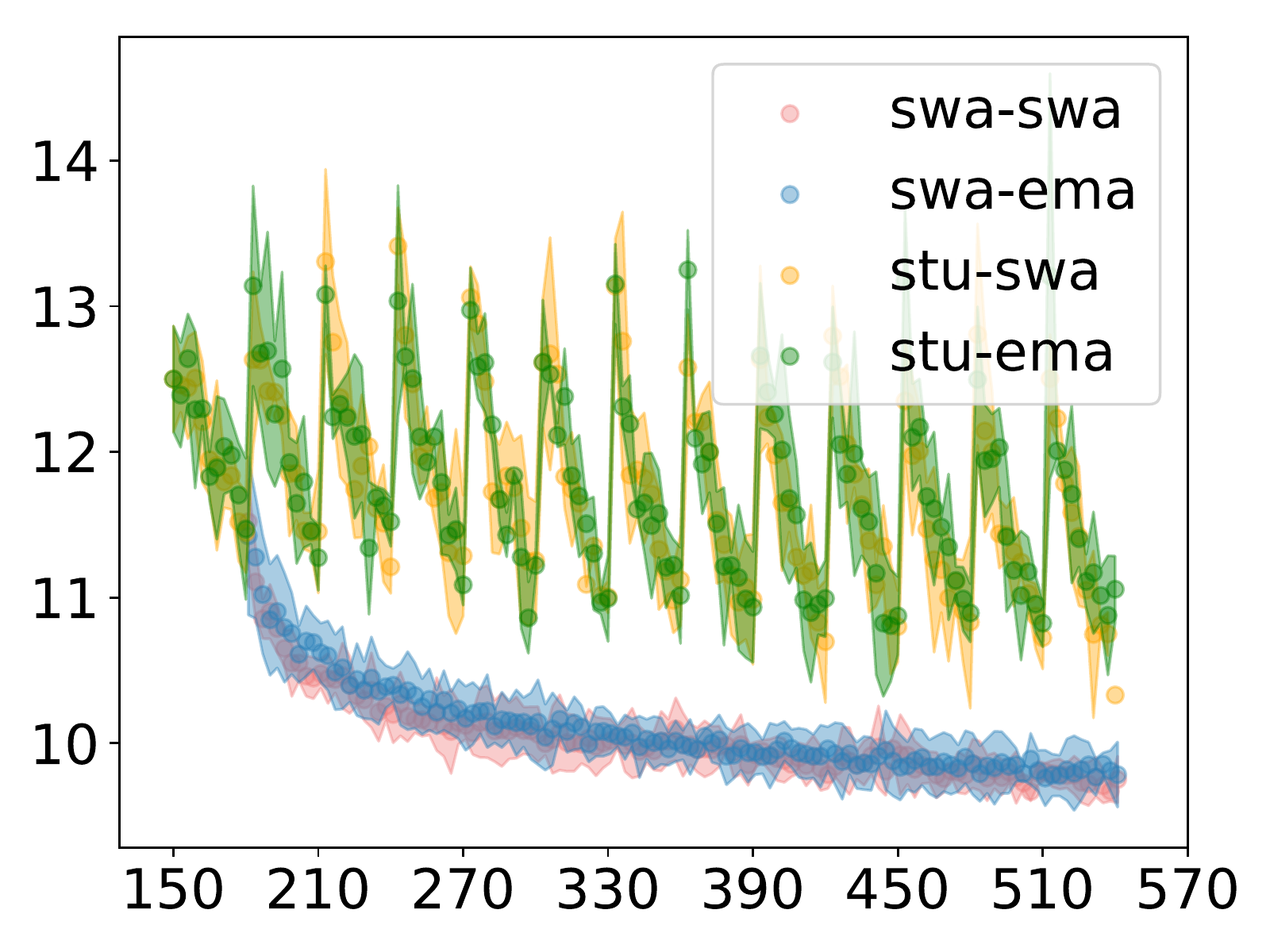}
\endminipage \hfill
  \caption{
  \textbf{(left)} Comparison of different averaging methods. The y axis 
    corresponds to the increased error with respect to the MT model with 
    \fastswa solution (which has $y=0$). All numbers are taken from epoch $180$.
  \textbf{(right)} The effects of using SWA as a teacher. {\sc w-t} model 
    corresponds to the performance of a model with weight {\sc w} using a model 
    with a teacher being {\sc t}. 
  }
  \label{fig:ema_swa} 
\end{figure}

The MT model uses an exponential moving average (EMA) of the student weights as 
a teacher in the consistency regularization term. 
We consider two potential effects of using EMA as a teacher: first,
averaging weights improves performance of the teacher for the reasons discussed
in Sections \ref{sec:trajectories}, \ref{sec:empirical_convexity}; second,
having a better teacher model leads to better student performance which in turn
further improves the teacher. In this section we try to separate these two
effects. We apply EMA to the \pi model in the same way in which we apply \fastswa
instead of using EMA as a teacher and compare the resulting performance to
the Mean Teacher. Figure \ref{fig:ema_swa} shows the improvement in error-rate 
obtained by applying EMA to the \pi model in different label settings. 
As we can see while EMA improves the results over the baseline \pi model, the 
performance of \pi-EMA is still inferior to that of the Mean Teacher method, especially
when the labeled data is scarce.
This observation suggests that the improvement of the Mean Teacher over the 
\pi model can not be simply attributed to EMA improving the student performance
and we should take the second effect discussed above into account.

Like SWA, EMA is a way to average weights of the networks, but it puts more 
emphasis on very recent models compared to SWA. 
Early in training when the student model  
changes rapidly EMA significantly improves performance and helps a lot
when used as a teacher.
However once the student model converges to the vicinity of the optimum, 
EMA offers little gain. In this regime SWA is 
a much better way to average weights. We show the performance of SWA 
applied to \pi model in Figure \ref{fig:ema_swa} (left).

Since SWA performs better than EMA, we also experiment with using SWA as a 
teacher instead of EMA. We start with the usual MT model pretrained until 
epoch $150$. Then we switch to using SWA as a teacher at epoch $150$. In Figure 
\ref{fig:ema_swa} (right), our results suggest that using SWA as a teacher
performs on par with using EMA as a teacher. We conjecture that once we are at 
a convex region of test error close to the optimum (epoch $150$), 
having a better teacher doesn't lead to substantially improved performance. It is 
possible to start using SWA as a teacher earlier in training; however, during early epochs 
where the model undergoes rapid improvement EMA is more sensible than SWA as
we discussed above.

\subsection{Consistency Loss Approximates Jacobian Norm} \label{sec:pi_model} 

\paragraph{Estimator mean and variance:}
In the simplified \pi model with small additive data perturbations that are normally distributed, $z\sim\N(0,I)$,
\[
\hat{Q} = \lim_{\epsilon \to 0}\tfrac{1}{\epsilon^2} 
\tfrac{1}{m}\sum_{i=1}^m\ell_{cons}(w, x_i,\epsilon) = \tfrac{1}{m}\sum_{i=1}^m \lim_{\epsilon \to 0}\tfrac{1}{\epsilon^2}\|f(w, x_i + \epsilon z_i) - f(w, x_i)\|^2 
\]

Taylor expanding $\ell_{cons}$ in $\epsilon$, we obtain $\ell_{cons} (w, x,\epsilon)  = 
  \epsilon^2 z^T J_x^T J_x z + O(\epsilon^4) $, where $J_x$ is the Jacobian of the network outputs $f$ with respect to the input at a particular value of $x$.
Therefore, 
\[
\hat{Q}_i = \lim_{\epsilon \to 0}\tfrac{1}{\epsilon^2}  \ell_{cons} (w, x_i,\epsilon) =  z_i^T J(x_i)^T J(x_i) z_i .
\]
We can now recognize this term as a one sample stochastic trace estimator for $\tr(J(x_i)^T J(x_i))$
 with a Gaussian probe variable $z_i$; see \citet{trace} for derivations and guarantees on stochastic trace estimators.
$$\mathbb{E}_z[\hat{Q}_i] = \mbox{tr }(J(x_i)^T J(x_i)\mathbb{E}[z_iz_i^T]) = \|J(x_i)\|^2_F.$$ Taking an expectation over the $m$ samples of $x$, we get $\mathbb{E} [ \hat{Q}] =  \mathbb{E}_x[\|J_x\|^2_F$. 

In general if we have $m$ samples of $x$ and $n$ sampled perturbations for each $x$, then for a symmetric matrix $A$ with $z_{ik} \stackrel{iid}{\sim} N(0,I)$ and independent $x_i \stackrel{iid}{\sim} p(x)$, 

the estimator $$\hat{Q} = \frac{1}{m} \sum_i^m \frac{1}{n} \sum_k^n z_{ik}^T A(x_i)z_{ik} \quad \textrm{      has variance}$$  
$$\textrm{Var}[\hat{Q}] = \frac{1}{m}\bigg(\textrm{Var}[\tr(A)] + \frac{2}{n} \mathbb{E}[\tr(A^2)]\bigg). $$

Proof: 
 Let $q_{ik} \equiv z_{ik}^T A(x_i)z_{ik}$.
It is easy to show that for fixed $x$, $\mathbb{E}_z[q_{11}|x_1] = 2 \tr(A(x_1)^2) + \tr(A(x_1))^2$, \citep[see e.g.][]{trace}. 
Note that $\mathbb{E}_{z_1,z_2}[q_{i1} q_{i2}|x_i] = \tr(A(x_i))^2$. Since $\big\{\frac{1}{n}\sum_k^n q_{ik}\big\}_{i=1}^{m}$ are i.i.d random variables,
\[ 
 \textrm{Var} \big[\frac{1}{m}\sum_{i}^m \frac{1}{n}\sum_{k}^n q_{ik}\big] = \frac{1}{m}\textrm{Var} \big[ \frac{1}{n}\sum_k^n q_{1k}\big],
\]
whereas this does not hold for the opposite ordering of the sum.
\begin{align*} 
\mathbb{E}\big[\big(\frac{1}{n}\sum_k^n q_{1k}\big)^2\big] &=
\mathbb{E}_{x_1} \mathbb{E}_{\{z\}}\bigg[\frac{1}{n^2}\sum_l^n\sum_k^n q_{il}q_{ik}\big| \{x\}\bigg] \\
&= \mathbb{E}_{x_1} \mathbb{E}_{\{z\}}\bigg[\frac{n}{n^2} q_{11}^2 + \frac{n(n-1)}{n^2} q_{11}q_{12}\big| \{x\}\bigg]\\
&=  \mathbb{E}_x \bigg[\frac{1}{n} \big(2\tr(A^2) + \tr(A)^2 \big) + \big(1 - \frac{1}{n} \big)\tr(A)^2 \bigg]
=  \bigg(\mathbb{E}_x[\tr(A)^2] + \frac{2}{n}\mathbb{E}_x[\tr(A^2)] \bigg)
\end{align*} 

Plugging in $A = J^T J$ and $n=1$, we get
\[
    \textrm{Var}[\hat{Q}] =  \frac{1}{m}\bigg(\textrm{Var}[\|J_x\|^2_F] + 2 \mathbb{E}[\|J_x^TJ_x\|^2_F]\bigg).
\]

\paragraph{Non-isotropic perturbations along data manifold}
Consistency regularization with natural perturbations such as image translation
can also be understood as penalizing a Jacobian norm as in Section \ref{sec:flatness_theory}. For example, consider perturbations sampled from a normal distribution on the tangent space, $z \sim P(x)\N(0,I)$
where $P(x) = P(x)^2$ is the orthogonal projection matrix that projects down from $\mathbb{R}^d$ to $T_x(\mathcal{M})$, the tangent space of the
 image manifold at $x$. Then the consistency regularization penalizes the Laplacian norm of the network on the manifold (with the inherited metric 
 from $\R^d$). $\E[z] = 0$ and $\E[zz^T] = PP^T (=) P^2 = P$ which follows if $P$ is an orthogonal projection matrix. Then,
 \[\E[z^T J^T J z] = \tr(J^T J PP^T) = \tr(P^TJ^TJP) = \tr(J_\mathcal{M}^TJ_\mathcal{M}) = \|J_\mathcal{M}\|^2_F.\] 
We view the standard data augmentations such as random translation
(that are applied in the \pi and MT models) as approximating samples of nearby elements of the data manifold and therefore differences $x'-x$ approximate elements of
its tangent space.

\subsection{Relationship Between $\mathbb{E}_x[\|J_w\|^2_F]$ and Random Ray Sharpness}
\label{sec:broad}
In the following analysis we review an argument for why smaller $\mathbb{E}_x[\|J_w\|^2_F]$, 
implies broader optima. To keep things simple, we focus on the MSE loss, but in principle
a similar argument should apply for the Cross Entropy and the Error rate. For a single
data point $x$ and one hot vector $y$ with $k$ classes, the hessian of $\ell_{MSE}(w) = \| f(x,w) - y \|^2$ can be decomposed into two terms, the Gauss-Newton matrix $G =  J_w^TJ_w$ and a term which depends on the labels.
\[H(w,x,y) = \nabla^2\ell_{MSE}(w) = J_w^TJ_w + \sum_{i=1}^{k} (\nabla^2 f_i) (f_i(x) - y_i),
\]

\[\tr(H) =  \|J_w\|^2_F + \underbrace{\sum_{i=1}^{k} \tr(\nabla^2 f_i) (f_i(x) - y_i)}_{\alpha(x,y)}
\]

Thus $\tr(H)$ is also the sum of two terms, $\|J_w\|^2_F$ and $\alpha$. As the solution improves, the relative size of $\alpha$ goes down.
In terms of random ray sharpness, consider the expected $\textrm{MSE}$ loss, or risk, $R_\textrm{MSE}(w)  =  \mathbb{E}_{(x,y)} \| f(x,w) - y \|^2$ along random rays. 
Let $d$ be a random vector sampled
from the unit sphere and $s$ is the distance along the random ray. Evaluating the risk on a random ray,
and Taylor expanding in $s$ we have
\[
  R_\textrm{MSE} (w+sd) 
 = 
  R_\textrm{MSE} (w) + sd^T \mathbb{E}_{(x,y)} [J_w^T (f-y)] + (1/2)s^2 d^T  \mathbb{E}_{(x,y)} [H]d + O(s^3)
\]

Since $d$ is from the unit sphere, $\mathbb{E}[d] = 0$ and $\mathbb{E}[dd^T] = I/p$ where $p$ is the dimension.
Averaging over the rays, $d \sim \textrm{Unif}(S^{p-1})$, we have
\[
   \mathbb{E}_d[R_\textrm{MSE} (w+sd)] - R_\textrm{MSE}(w)
 = 
  \frac{s^2}{2p} \mathbb{E}_x[\tr(H)]+O(s^4) = \frac{s^2}{2p} \mathbb{E}_x[\|J_w\|^2_F] + \frac{s^2}{2p} \mathbb{E}_{(x,y)}[\alpha(x,y)]+ O(s^4)
\]
All of the odd terms vanish because of the reflection symmetry of the unit sphere. This means that locally,
the sharpness of the optima (as measured by random rays) can be lowered by decreasing $\mathbb{E}_x[\|J_w\|^2_F]$. 

\subsection{Including High Learning Rate Iterates Into \swa} \label{sec:theory_fastswa}
As discussed in \citet{mandt2017}, under certain assumptions SGD samples from a Gaussian distribution centered at the optimum of the loss $w_0$
with covariance proportional to the learning rate. Suppose then that we have $n$ weights sampled at learning rate $\eta_1$, $w_i^{(1)}  \stackrel{iid}{\sim} \N(w_0, \eta_1\Sigma)$ and $m$ weights sampled
with the higher learning rate $\eta_2$,
$w_j^{(2)} \stackrel{iid}{\sim} \N(w_0, \eta_2\Sigma)$. 
For the SWA estimator $\hat{w}_{\textrm{SWA}} = \frac{1}{n} \sum_i w_i^{(1)}$,
$\mathbb{E}[\|\hat{w}_\textrm{SWA} - w_0\|^2] = \tr(\textrm{Cov}(\hat{w}_{\textrm{SWA}})) = \frac{\eta_1}{n} \tr(\Sigma)$.
But if we include the high variance points in the average, as in \fastswa,  $\hat{w}_{\textrm{fSWA}} = \frac{1}{n+m} \big(\sum_i w_i^{(1)} + \sum_j w_j^{(2)}\big)$, then
$\mathbb{E}[\|\hat{w}_\textrm{fSWA} - w_0\|^2] = \frac{n\eta_1 + m\eta_2}{(n+m)^2} \tr(\Sigma)$. If $\frac{n\eta_1 + m\eta_2}{(n+m)^2} < \frac{\eta_1}{n}$ then including the high learning rate points decreases the MSE of the estimator
for $m > n \big(\frac{\eta_2}{\eta_1} - 2 \big)$. If we include enough points, we will still improve the estimate.


\subsection{Network Architectures} \label{sec:sup_architectures}

In the experiments we use two DNN architectures -- $13$ layer CNN and
Shake-Shake. The architecture of $13$-layer CNN is described in Table 
\ref{tbl:architecture}. It closely follows the architecture used in 
\citep{pi, vat2, mt}. We re-implement it in PyTorch and removed the Gaussian
input noise, since we found having no such noise improves generalization.
For Shake-Shake we use 26-2x96d Shake-Shake regularized architecture of 
\citet{shake} with $12$ residual blocks.

\begin{table}[H]
\centering
\begin{threeparttable}[t]
\caption{
\label{tbl:architecture}
A 13-layer convolutional neural networks for the CNN experiments (CIFAR-10 and CIFAR-100) in Section \ref{sec:cifar10_cnn} and \ref{sec:cifar100_cnn}. Note that the difference from the architecture used in \citet{mt} is that we removed a Gaussian noise layer after the horizontal flip. 
}
\vspace*{\baselineskip}
\begin{tabular}{ l l }
\noalign{\medskip}
\bf{Layer} & \bf{Hyperparameters} \\
\Xhline{1pt}\noalign{\smallskip}
Input  & $32\times32$ RGB image \\
Translation & Randomly \{$\Delta x, \Delta y\} \sim [-4, 4]$ \\
Horizontal flip & Randomly $p=0.5$ \\
Convolutional & $128$ filters, $3\times3$, \textit{same} padding \\
Convolutional & $128$ filters, $3\times3$, \textit{same} padding \\
Convolutional & $128$ filters, $3\times3$, \textit{same} padding \\
Pooling   & Maxpool $2\times2$ \\
Dropout   & $p=0.5$ \\
Convolutional & $256$ filters, $3\times3$, \textit{same} padding \\
Convolutional & $256$ filters, $3\times3$, \textit{same} padding \\
Convolutional & $256$ filters, $3\times3$, \textit{same} padding \\
Pooling & Maxpool $2\times2$ \\
Dropout & $p=0.5$ \\
Convolutional & $512$ filters, $3\times3$, \textit{valid} padding \\
Convolutional & $256$ filters, $1\times1$, \textit{same} padding \\
Convolutional & $128$ filters, $1\times1$, \textit{same} padding \\
Pooling & Average pool ($6\times6 \to 1\times$1 pixels) \\
Softmax & Fully connected $128 \to 10$ \\
\Xhline{1pt}\noalign{\smallskip}
\end{tabular}
\end{threeparttable}
\end{table}

\subsection{Hyperparameters} \label{sec:sup_hypers}

We consider
two different schedules. In the \emph{short schedule} we set the 
cosine half-period $\ell_0 = 210$ and training length $\ell = 180$, following
the schedule used in \citet{mt} in Shake-Shake experiments. 
For our Shake-Shake experiments we also report results with  
\emph{long schedule} where we set $\ell = 1800, \ell_0 = 1500$ following
\citet{shake}. To determine the initial learning rate $\eta_0$ and the cycle length $c$
we used a separate validation set of size $5000$ taken from the unlabeled data.
After determining these values, we added the validation set to the unlabeled data and
trained again. We reuse the same values of $\eta_0$ and
$c$ for all experiments with different numbers of labeled data for both
\pi model and Mean Teacher for a fixed architecture ($13$-layer CNN or Shake-Shake).
For the short schedule we use cycle length $c=30$ and average models once every
$k=3$ epochs. For long schedule we use $c=200$, $k=20$.

In all experiments we use stochastic gradient descent optimizer with Nesterov momentum 
\citep{sgdr_cosine_restart}. 
In \fastswa we average every the weights of the models corresponding to every
third epoch.
In the \pi model, we back-propagate the gradients through the student side only 
(as opposed to both sides in \citep{te}).
For Mean Teacher we use $\alpha=0.97$ decay rate 
in the Exponential Moving Average (EMA) of the student's weights.
For all other hyper-parameters we reuse the values from \citet{mt} 
unless mentioned otherwise.

Like in \citet{mt}, we use $\| \cdot \|^2$ for divergence in the consistency loss. Similarly, we ramp up the consistency cost $\lambda$ over the first $5$ epochs from $0$ up to it's maximum value of $100$
as done in \cite{mt}. We use cosine annealing learning rates with no learning rate ramp up, unlike in the original MT implementation \citep{mt}.
Note that this is similar to the same hyperparameter settings as in \citet{mt} for ResNet\footnote{We use the public Pytorch code \url{https://github.com/CuriousAI/mean-teacher} as our base model for the MT model and modified it for the \pi model.}. We note that we use the exact same hyperparameters for the \pi and MT models in each experiment setting. 
In contrast to the original implementation in \citet{mt} of CNN experiments, we use SGD instead of Adam. 

\paragraph{Understanding Experiments in Sections \ref{sec:trajectories}, 
    \ref{sec:empirical_convexity}}
We use the $13$-layer CNN with the short learning
rate schedule. 
We use a total batch size of $100$ for CNN experiments with a labeled batch size of $50$
for the \pi and Mean Teacher models.
We use the maximum learning rate  $\eta_0 = 0.1$.
For Section \ref{sec:trajectories} we run SGD only for $180$ 
epochs, so $0$ learning rate cycles are done. For Section \ref{sec:empirical_convexity}
we additionally run $5$ learning rate cycles and sample pairs of SGD iterates
from epochs $180$-$330$ corresponding to these cycles.
 
\paragraph{CIFAR-10 CNN Experiments}
We use a total batch size of $100$ for CNN experiments with a labeled batch size of $50$. We use the maximum learning rate  $\eta_0 = 0.1$.

\paragraph{CIFAR-10 ResNet + Shake-Shake} 
We use a total batch size of $128$ for ResNet experiments with a labeled batch size of $31$.  We use the maximum learning rate $\eta_0 = 0.05$ for CIFAR-10.
This applies for both the short and long schedules.

\paragraph{CIFAR-100 CNN Experiments}
We use a total batch size of $128$ with a labeled batch size of $31$ for $10k$ and $50k$ label settings. For the settings $50k$+$500k$ and $50k$+$237k^*$, we use a labeled batch size of $64$. We also limit the number of unlabeled images used in each epoch to $100k$ images.  We use the maximum learning rate  $\eta_0 = 0.1$. 

\paragraph{CIFAR-100 ResNet + Shake-Shake} 
We use a total batch size of $128$ for ResNet experiments with a labeled batch size of $31$ in all label settings. For the settings $50k$+$500k$ and $50k$+$237k^*$, we also limit the number of unlabeled images used in each epoch to $100k$ images.  We use the maximum learning rate  $\eta_0 = 0.1$.
This applies for both the short and long schedules.

\subsection{Domain Adaptation}  \label{sec:exp_da}

We apply \fastswa to the best experiment setting {\sc MT+CT+TFA} for CIFAR-10 
to STL according to \citet{da_mt}. This setting involves using confidence 
thresholding (CT) and also an augmentation scheme with translation, flipping, 
and affine transformation (TFA). 

We modify the optimizer to use SGD instead of Adam \citep{adam} and use cosine 
annealing schedule with $\ell_0=600, \ell=550, c=50$. We experimented with two 
\fastswa methods: averaging weights once per epoch and averaging once every iteration, which is much more frequent that averaging every epoch as in the semi-supervised case. 
Interestingly, we found that for this task averaging the
weights in the end of every iteration in \fastswa converges significantly
faster than averaging once per epoch and results in better  
performance. 
We report the results in Table \ref{table:da}. 

We observe that averaging every iteration converges much faster 
($600$ epochs instead of $3000$) and results in better test accuracy. 
In our experiments with semi-supervised learning averaging more often than
once per epoch didn't improve convergence or final results. We hypothesize
that the improvement from more frequent averaging is a result of specific 
geometry of the loss surfaces and training trajectories in domain adaptation.
We leave further analysis of applying \fastswa to domain adaptation for future
work.

\begin{table}[h]
  \caption{Domain Adaptation from CIFAR-10 to STL.  VADA results are from \citep{dirt} and the original SE$^*$ is from \citet{da_mt}. SE is the score with our implementation without \fastswa. \fastswa$^1$ performs averaging every epoch and the final result is obtained at epoch $3000$. \fastswa$^2$ performs the averaging every \emph{iteration} and the final result is obtained at epoch $600$. 
  }
  \label{table:da}
  \centering
  \begin{tabular}{c c c c c c}
    Method & VADA  & SE$^*$ & SE & SE + \fastswa$^1$ & SE + \fastswa$^2$  \\
    \midrule
    Test Error & 20.0 & 19.9  & 18.1 & 17.1 & \tbf{16.8}
  \end{tabular}
\end{table}

\paragraph{Implementation Details}
We use the public code\footnote{\url{https://github.com/Britefury/self-ensemble-visual-domain-adapt.git}} of \citet{da_mt} to train the model and apply \fastswa.  While the original implementation uses Adam \citep{adam}, we use stochastic gradient descent with Nesterov momentum and cosine annealing learning rate with $\ell_0 = 600, \ell = 550, c = 100$ and $k=100$. We use the maximum learning rate $\eta_0 = 0.1$ and momentum $0.9$ with weight decay of scale $2 \times 10^{-4}$. We use the data augmentation setting MT+CF+TFA in Table 1 of \citet{da_mt} and apply \fastswa. The result reported is from epoch $4000$.

\end{document}

%% file: math_commands.tex
\usepackage{amsmath,amsfonts,bm}

\def\eqref#1{equation~\ref{#1}}

\def\1{\bm{1}}

\DeclareMathAlphabet{\mathsfit}{\encodingdefault}{\sfdefault}{m}{sl}
\SetMathAlphabet{\mathsfit}{bold}{\encodingdefault}{\sfdefault}{bx}{n}

\newcommand{\E}{\mathbb{E}}

\newcommand{\R}{\mathbb{R}}

